\documentclass{article}

\usepackage{float}
\usepackage[section]{placeins}
\usepackage{todonotes}
\usepackage{xltabular}
\usepackage{amsmath}
\usepackage{amssymb}
\usepackage{longtable}
\usepackage{graphicx}
\usepackage{array}
\usepackage{booktabs}
\usepackage{longtable}
\usepackage{adjustbox}
\usepackage{tabularx}
\usepackage{makecell}
\usepackage[table]{xcolor}
\usepackage{enumitem}
\usepackage[preprint]{neurips_2026}

\usepackage[utf8]{inputenc} 
\usepackage[T1]{fontenc}    
\usepackage{hyperref}       
\usepackage{colortbl}

\definecolor{goodblue}{RGB}{0, 91, 187}
\usepackage{hyperref}
\hypersetup{
  colorlinks=true,
  allcolors=goodblue,
  urlcolor=goodblue,
  citecolor=goodblue,
  pdfborder={0 0 0},
  breaklinks=true,
}

\usepackage{url}            
\usepackage{booktabs}       
\usepackage{amsfonts}       
\usepackage{nicefrac}       
\usepackage{microtype}      
\usepackage{xcolor} 
\title{LLMs are not stochastic parrots: Evidence for meaning-mediated abstraction from conlang-like tasks}

\author{%
Julia Witte Zimmerman\thanks{\texttt{julia.zimmerman@uvm.edu}}$^{*,\dagger,\ddagger}$ \\
Calla G. Beauregard$^{*,\ddagger}$ \\
Tabia Tanzin Prama$^{*}$ \\
Parisa Suchdev$^{\dagger}$ \\
Kathryn Cramer$^{*}$ \\
Elisabeth Kollrack$^{*}$ \\
\\[0.5em]
$^{*}$Computational Story Lab \\
$^{\dagger}$Computational Ethics Lab \\
$^{\ddagger}$Equal contribution (first project members) \\
Vermont Complex Systems Institute \\
University of Vermont \\
Burlington, VT 05405, USA \\
}

\begin{document}

\maketitle

\begin{abstract}
    The strong version of the \textit{stochastic parrot} argument claims that, although large language models (LLMs) may exceed rote regurgitation, they cannot move beyond statistical pattern matching into abstraction or reasoning, remaining ontologically near the lower bound of pattern reuse despite producing alluringly fluent text.
    We test this hypothesis using conlang-like tasks. Several LLMs are given only natural‑language descriptions of fictional languages that subvert prominent superficial patterns in training data by combining statistically uncommon and unattested features. Crucially, no example outputs are given.
    We argue that if the models exhibit rule-following behaviour, they cannot be relying solely on superficial statistical patterns; such patterns often work against the correct output. Instead, successful performance requires representations of the constraints specified in the prompt. Across three complementary task families, models systematically move in the meaning-predicted direction: they distinguish prompt exposure from instructed use, alter semantic relationships in response to novel constraints, and sometimes produce exact matches to complex translation answer keys. Although performance varies across the spectrum of models used, these results provide evidence for meaning-mediated abstraction in LLMs and refute the strong stochastic parrot hypothesis. Our work shows that, under appropriate architectural and contextual constraints, statistical learning can produce meaning-mediated abstractions, although generation remains strongly constrained by superficial plausibility. We discuss implications for model development and for understanding how increasingly abstract representations may emerge from plausible-text-generation objectives.
\end{abstract}

\section{Introduction}
\label{sec:introduction}

This pre-print describes the initial results of an empirical probe of a widespread claim about LLM limitations. It has been created to provide access to our findings to participants at ISC Summer School 2026, where we presented this project as a poster, as well as IC2S2 2026 and SLSA/ 4S 2026, where we presented\footnote{Or will present, in the case of SLSA/ 4S 2026.} related works~\citep{ISC2026SummerSchool,IC2S22026,SLSA4S2026,4S2026}. 

In this work, we test the strong version of the \textit{stochastic parrot} hypothesis using conlang-like tasks\footnote{Constructed languages, or conlangs, are communication systems intentionally designed by someone (or something, or someones). For brevity, we refer to our conlang-like tasks as conlangs, tasks, or rules.} where following the meaning of an instruction conflicts with familiar surface distributions. This project is currently underway with a suite of 14 conlang-like rules (17 sets of prompts, with 2 non-conlang controls and 1 correction), but we present our initial results focused on a subset of these rules and a subset of our planned analyses as part 1 of 2.

A brief summary of the \textbf{key points} in this pre-print:
\begin{itemize}[nosep] 
    \item Statistical plausibility and meaning-mediated plausibility can be experimentally dissociated. Our three conlang families dissociate plausibilities in different ways: in brief, prompt exposure versus instructed use, semantic similarity versus instructed irrelevance, and relational preservation across radically different surface forms.
    \item Across three independent rule families LLMs moved in the meaning-predicted direction, showing that superficial statistical patterns are not a sufficient explanation for their behavior even though such patterns remain powerful constraints. For every rule, at least one model shows clear evidence of rule-following; for most rules, most models do.
    \item The strong stochastic parrot hypothesis is behaviorally inadequate, even for base-model LLMs, because successful performance on these tasks is better explained by meaning-mediated abstraction than by superficial statistical regularities.
    \item We provide an existence proof of meaning-mediated abstraction implicated through rule-following behaviour, not evidence of human-equivalent understanding.\footnote{Much less, ``reasoning''.} 
    \item Plausible output alone does not reveal its mechanism: Benchmark success can come without construct validity. 
    \item Next-token prediction may give models access to meaning while simultaneously anchoring generation disproportionately to its most superficial and fungible forms. One bottleneck may be control over which kind of plausibility governs generation, rather than knowledge acquisition per se.\footnote{In other words, PTGA may over-privilege token-scale plausibility, and some successful interventions may work by changing which relationships/ patterns are salient (and local). Note that we do not suggest that this is the \textit{only} bottleneck! Other kinds of reasoning -- beyond deciding what is plausible -- may or may not be something that can be taught through PTGA (for example, what might be called metacognition and deliberative reasoning). The relationship between the meanings which can be encoded and shared via language and the kinds of thinking that can be done \textit{with} those meanings is not at all clear.}
\end{itemize}

\subsection{Why this mental model matters}
\label{sec:whythismentalmodelmatters}

As with the introduction of prior information technologies -- photography, printing, even writing -- the reaction to Generative AI (Gen AI) is divergent and contentious due to its destabilizing effects amongst a complex milieu of motivations, beliefs, and constraints~\citep{yates_memory,zimmerman2025locality}. To contend with these disruptions, people develop mental models (including specific phrases and associations) which they draw on when answering questions as to Gen AI's behavior, origin, and capacity, in order to meaningfully interpret it~\citep{RappDiLodovicoTorrielliDicar2025,g_wayofwords,Agha2003sociallife,eckert2008variation,silverstein2003indexical,Peirce1931,gibson1979affordances}.\footnote{Although not the subject of this paper, we have two in-progress works related to this topic~\citep{VALIproject,indexicalityproject}.} A similarly complex milieu shapes how the components of those mental models develop and endure~\citep{IrvineGal2000}. We believe good mental models are a necessary precursor to effectively navigating this fraught landscape (independent of stance, but see Sec.~\ref{appendix:disclosures}).\footnote{Bad mental models have multiple harms. They can lead to incorrect predictions, act as thought-terminating cliches, undermine an otherwise compelling argument in the eyes of the intended audience, and cause disagreement where none was necessitated.}

Regardless of the original intent behind it, the term \textit{stochastic parrot} has been widely and variously adopted since its introduction in \citet{10.1145/3442188.3445922}. It is now a popular mental model for Gen AI, and is frequently applied even to models that differ substantially from those discussed in the original paper.

In this work, we are not concerned with the term itself, but with empirically evaluating a claim it is often taken to imply: that LLMs trained on text are fundamentally limited to relatively superficial statistical regularities and therefore incapable of forming the kinds of abstractions associated with meaningful (or thoughtful) behavior.\footnote{This mental model is prevalent. Beyond popular descriptions such as ``spicy autocorrect,'' we have encountered similar claims in academic settings, e.g. a comparison of deep learning to photographs that assemble meaningful patterns via photons without interacting with that meaning beyond its most superficial manifestation.} For our purposes, the strong stochastic parrot hypothesis (SSP) is the view that model behaviour can ultimately be explained in terms of token co-occurrence statistics, surface-level syntactic patterns, and similar symbolic patterns. By contrast, we argue that, given the nature of the correct outputs in our tasks, explanations invoking less superficial aspects of meaning are substantially more compelling.

Rather than treating statistical learning and abstraction as mutually exclusive possibilities, we ask under what conditions statistical learning can give rise to increasingly abstract behavior.\footnote{And potentially increasingly thought-like, connecting meaning construction across architectures and scales, drawing on the Distributional Hypothesis (DH) as a partial explanation for similar processes in people. Although LLM architecture is extremely different from ours, the DH is explicitly the inspiration for the distributional-semantics-related aspects of LLM architecture as well~\citep{sahlgren2008distributional,harris1954structure,harris1968mathematical}. In this work, our aim is not to demonstrate equivalence between human and machine cognition -- or to argue that machines can ``think'' -- but to evaluate a more specific claim about the limits of statistical learning.}
We argue that understanding how statistical learning scaffolds transitions along the concrete-abstract pattern spectrum is important for understanding AI systems as well as human cognition.\footnote{Language operates across multiple levels of abstraction, from highly fungible surface forms to deeply contextual and socially grounded interpretations. See discussion of the pattern-to-thought transition and the fungibility-groundedness spectrum in \citet{zimmerman2025locality}. See also upcoming works with Grayson Wycliffe Storer, Alice Patania, and Juniper L. Lovato.} Dismissing statistical learning as inherently incapable risks obscuring mechanisms through which meaning can emerge.

We contend that if models systematically move in the direction predicted by our tasks, then explanations invoking only superficial symbolic regularities become increasingly difficult to maintain.

\subsection{What is plausibility?}
\label{sec:plausibility}

Despite significant variation, LLMs are fundamentally built around plausible text generating architectures, comprising plausible text generating objectives (such as masked language models like RoBERTa~\citep{liu2019robertarobustlyoptimizedbert} and autoregressive transformers like ChatGPT~\citep{openai2024gpt4technicalreport}) and token-centric design choices. Specifically, next-token-prediction (NTP) is the popular architectural choice underlying the models we utilize here. For concision, we're referring back to the family of neural architectures united by this fundamental learning strategy as plausible-text-generating-architectures (PTGA).

These models optimize for plausibility by repeatedly guessing what token comes next and being rewarded for good guesses, based on a large corpus of mostly human-generated text. In that operationalization of plausibility, everything that goes into making someone likely to continue the utterance in that particular way is collapsed into a single score.\footnote{Determinants such as world knowledge, mood, past experience, history with the speaker, situation, pragmatics, prosody, semantics, tone, location, physical ability (for example, ability to hear in a noisy room), propositional content, implications, consequences, belief, truth, commonsense, motivation, personality, attentiveness, familiarity, memory, and so on. On a fungibility--groundedness spectrum, the most personal of these factors are unique and only indirectly observable through the stigmergic trace of the communication. The more fungible a factor is, the more likely it is to be shared across individuals through common embodiment, culture, or experience, and therefore detectable by a statistical learner.}

Speaking approximately, many determinants of plausibility fall under what we might call the \textit{meaning} of the utterance. Some are more directly perceptible via statistical learning than others, with the most accessible to an LLM being superficial statistical regularities such as token co-occurrence. Other determinants may require increasingly abstract representations, which may themselves be assembled from lower-level patterns. Humans appear capable of learning across this spectrum, both from concrete statistical regularities and from contextually-mediated abstractions.\footnote{Although both are relevant to meaning and language, the abstract--concrete spectrum does not necessarily map cleanly onto the fungibility--groundedness spectrum. Some highly abstract concepts are widely communicable through language, and some comparatively concrete aspects of experience remain difficult to transmit except through direct grounding~\citep{Nagel1974-NAGWII}. Our results therefore speak primarily to abstractions recoverable from linguistic evidence, not to every form of meaning, cognition, or reasoning. The potential of LLMs to derive indirect grounding in human experience is intriguing but out of scope for this paper. More discussion of this topic is available in \citet{zimmerman2025locality}. For an example of how medium and message are inherently intertwined and both decoding and encoding are architecture-dependent, see \citet{pera2026billionssketchesrevealhidden}.}

Plausibility alone is lossy. When an LLM generates plausible text, it is unclear whether that is a result of the model's understanding of the relevant concepts -- any part of what might be called the \textit{meaning} of the prompt (beyond the most explicitly statistical components) -- or a result of successful pattern recognition and regurgitation. For convenience, we'll call these meaning-mediated plausibility or meaning plausible and statistically-mediated plausibility or statistically plausible, respectively.\footnote{However, keep in mind these are conceptually along the same concrete-abstract, fungibility-groundedness spectrum of meaning, with the superficial statistical patterns (symbol co-occurrence, some morphology and syntax) on the more concrete, immediately apparent side and semantic, pragmatic, register, and discourse-level content higher on the abstract side. There is no bright line between them! We want to probe the abstract end and discredit the most concrete end (including some low-level abstractions) as sufficient explanation for behaviour.}

The fact that superficial statistical likelihood often\footnote{And perhaps specifically, often but not always. We will return to this topic in Section~\ref{sec:modulatingptga}.} aligns with the meaning of an utterance is what makes plausible text generation so powerful. Under the right conditions, statistical learning may access traces of meaning, not just the most superficial patterns visible in the training data. Meaning-mediated abstractions therefore need not be seen as alternatives to statistical learning. They may emerge because they are useful compressions for solving statistical prediction problems, both \textit{in silico} and \textit{in vivo}~\citep{Greco2024PredictiveLearning,Stockl2024PredictionLearning}. 

The challenge is that plausible outputs do not reveal which process produced them. Both superficial statistical regularities and meaning-mediated abstractions can generate plausible text, obscuring inference of the underlying mechanism from the response alone. This creates a construct-validity problem for many benchmark-style evaluations (see Section~\ref{sec:performanceonexistingbenchmarks}). We design our experiment to address this challenge, by forcing the relatively concrete plausibilities of token statistics to contradict the relatively abstract plausibilities of meaning.

\begin{figure*}[t]
  \centering
  \includegraphics[width=0.49\columnwidth]{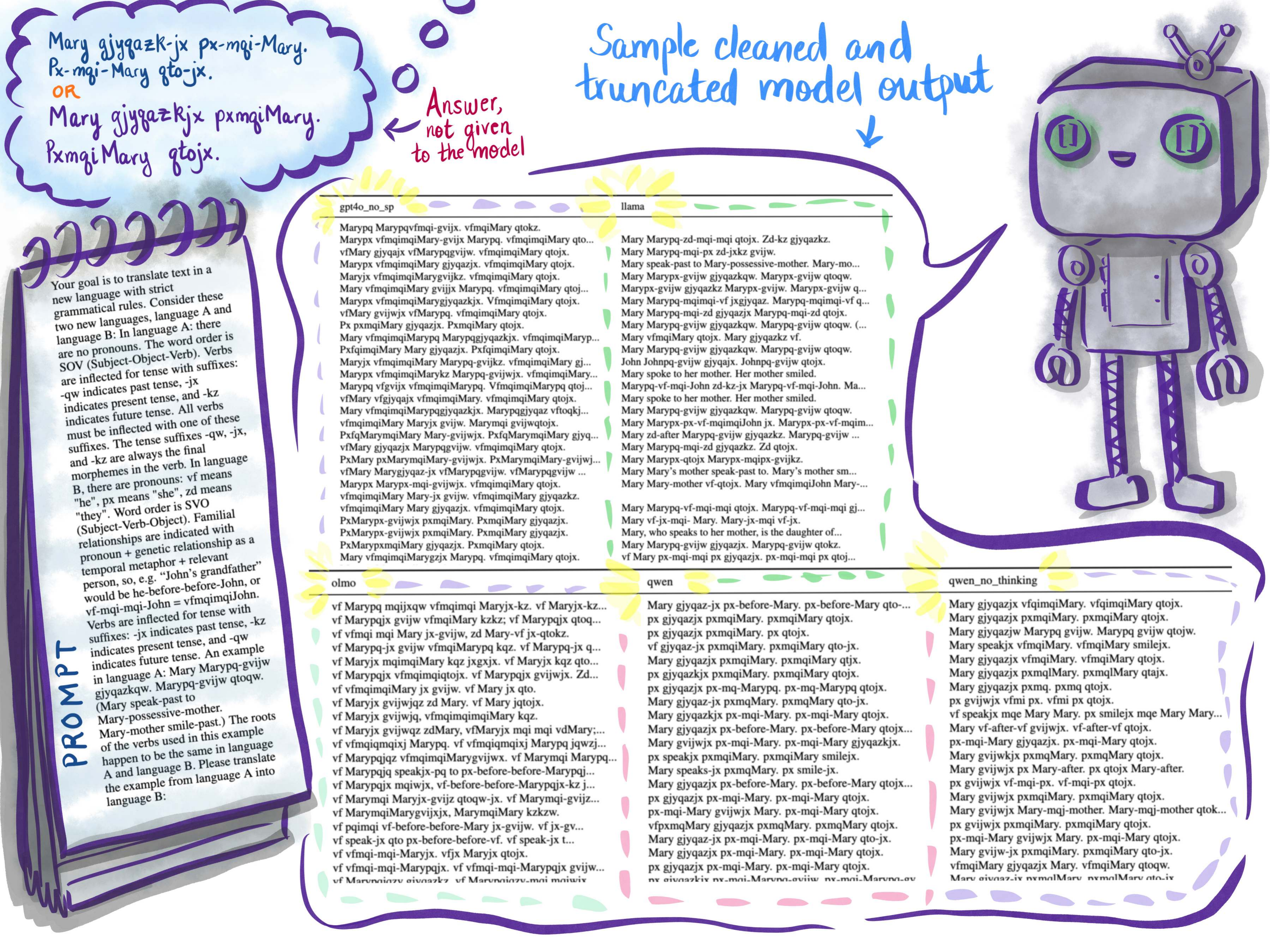}
  \includegraphics[width=0.49\columnwidth]{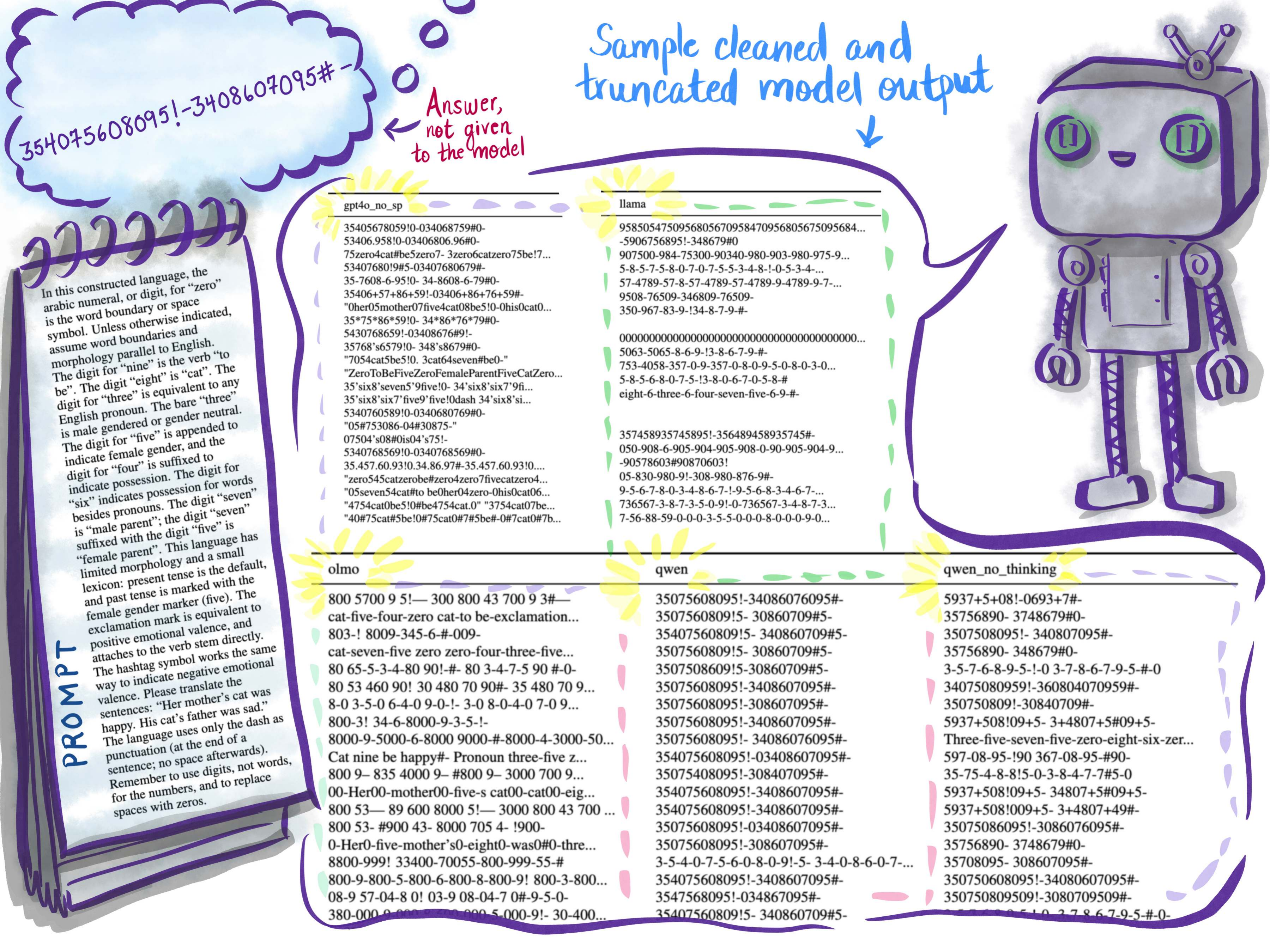}
  \caption{This figure shows prompts describing the translation family of rules (translate-1 and dan-1), their answer keys, and sample outputs from each model, to illustrate what we mean by separating meaning-mediated plausibility from statistical plausibility (although we do that in different ways across the rule families). It is apparent from these responses that these tasks are hard, that some models do much better than others, and that dan-1 is more difficult than translate-1. However, it is also apparent that models can produce outputs exactly matching or near to the answer keys, behaviour which is better explained by meaning plausibility than statistical plausibility.}
  \label{fig:translate1dan1sampleoutput}
\end{figure*}

\subsection{Prior work and superficial patterns} 
\label{sec:priorwork} 

Our argument is not that LLMs don't rely on patterns. Quite the opposite: the existence of learned patterns is what makes these systems interesting. If we allow patterns at any level of abstraction, essentially everything discussed in this paper can be described as pattern matching.\footnote{Even as statistical pattern matching! And described in terms of the plausible next thing.} The question is therefore not whether models learn and use patterns, but what kinds of patterns they learn (and when they learn them, when they use them, and what they can use them for, although all of those questions are out of scope for this project; but see Section~\ref{sec:discussion}). The strong stochastic parrot hypothesis (SSP) says that LLM behaviour can be explained by relatively concrete regularities: this symbol follows this symbol, these symbols tend to occur near those symbols, and so on.\footnote{By contrast, people appear capable of learning across a wide spectrum, from concrete regularities to more abstract principles that can be flexibly expressed through different symbolic systems, without a discrete boundary where pattern matching becomes abstraction.} See Fig.~\ref{fig:patternmatchinginstructionfollowing.png}.

Several prior works suggest that LLMs are not restricted solely to superficial patterns. \citet{digutsch2023overlap} find that semantic similarity predicts GPT-3 activations better than word co-occurrence. \citet{hewitt2025neologismlearningcontrollabilityselfverbalization} show that models can learn and self-verbalize meanings associated with newly introduced embeddings, while \citet{hewitt2024instructionfollowinginstructiontuning} find evidence of instruction-following behaviour prior to explicit instruction tuning. \citet{Cui_2024} identify the possibility of a transition from positional to semantic pattern matching.\footnote{Further discussion of the pattern-to-thought transition in machines in \citet{zimmerman2025locality} and \citet{zimmerman2024tokensoftoverlookedappetizerlarge}, although our views continue to evolve. The current work is also inspired by \citet{diamond2023genlangszipfslawlanguages,huang2023lexinvariantlanguagemodels,kallini2024mission,mahowald2024dissociatinglanguagethoughtlarge,zimmerman2024blind}.}

Taken together, these observations establish antecedent plausibility for meaning-mediated structure in LLMs, but they do not cleanly dissociate statistical and meaning-based explanations. Our contribution is to probe that distinction explicitly through conlang-like tasks designed to decouple meaning-mediated plausibility from familiar surface distributions.

The crux of our argument therefore depends on constructing tasks whose correct responses are statistically implausible based on natural language corpora, yet which are learnable from natural-language instruction. We use ``meaning-mediated plausibility'' to refer to outputs better explained by the content of an instruction than by its component surface distributions.\footnote{And ``statistical plausibility'' to refer to outputs that \text{can} be explained by their component surface distributions, whether they can be explained by other kinds of plausibility or not.}

\section{Methodology}
\label{sec:methodology}

Our research question is, \textbf{can a model generate an answer that is licensed by a natural-language instruction (plausible via meaning) but disfavored by superficial\footnote{At or near surface-level.} distributions (statistically implausible)}?

Our argument relies on three steps. First, we construct conlang-like tasks in which statistical plausibility and meaning-mediated plausibility point in different directions. See Figure~\ref{fig:translate1dan1sampleoutput} for examples. Second, we formulate explicit hypotheses about what successful rule-following should look like and measure movement in those directions. Third, we ask whether the resulting behaviour is better explained by the content of the instructions than by familiar surface distributions alone.\footnote{We do not view ``instruction following'' and ``abstraction'' as competing explanations. Successfully following a sufficiently novel instruction requires some representation of the instruction's content. Our argument is not that the representation must be human-like, but that explanations based solely on superficial symbol co-occurrence are inadequate. See Figure~\ref{fig:patternmatchinginstructionfollowing.png}.}

We are not trying to show that LLMs are general reasoners or that they learn, create, or use abstractions in a human-like way. Our argument posits the existence of some non-trivial meaning-mediated abstractions, accessible under some conditions, but we do not identify their specific nature.

\subsection{Model selection}
\label{sec:modelselection}

Because our rules vary in difficulty and operate near the frontier of model failure, we balance architectural simplicity against performance. Our strongest argument would be made with a purely autoregressive LLM, so the more ``pure'' a model is, the more imperfect task performance we are willing to tolerate. Based on preliminary testing, we selected models that span next-token-prediction purity, instruction tuning, reasoning-oriented training, openness and training-data transparency, interpretability, and sufficient performance to produce detectable output-level evidence. See Appendix~\ref{appendix:constructvalidity} and Table~\ref{tab:model_characteristics} for details. 

In short, GPT-4o and Qwen were selected for expected performance, OLMo for openness, and Llama as a base model. We treat Qwen with and without thinking enabled separately because we expect CoT-style (chain-of-thought style) fine-tuning to modulate the influence of lower-level statistical plausibility (Section~\ref{sec:modulatingptga}). As a base model, Llama's weaker directional shifts remain theoretically interesting (à la \citet{hewitt2024instructionfollowinginstructiontuning}), especially when considered as part of this spectrum of models. These models were chosen to support interpretation of the present experiment rather than to enable a controlled architecture or scaling comparison, so although our results motivate exploratory hypotheses, they are not probative on those questions.

\subsection{Experimental design}
\label{sec:experimentdesign}

We use five models: GPT-4o, Qwen (thinking, with CoT-style-prompting enabled), Qwen (no thinking, with CoT-style-prompting disabled), OLMo, and Llama.\footnote{There are some minor variations in how we refer to these models, but they all refer to the same five: GPT-4o, Qwen (with thinking enabled), Qwen (without thinking enabled), OLMo, and Llama. GPT-4o is sometimes described as ``no system prompt'', because it has a slightly different expected prompt format than the others. We treated all models as identically as possible, so we did not provide any additional system prompt to GPT-4o, even though that is a common application.} More detailed descriptions of each model are provided in Table~\ref{tab:model_characteristics}.

For each model and rule, we ran 50 independent outputs. Each rule -- except sidequest-1\footnote{Due to a copy-paste error, the sidequest-1 runs analyzed in part 1 use 2/3 identical prompts and 1/3 a second variant. Corrected data (sidequest-1-fix) will be analyzed in part 2.} -- has 3 prompt variations with slight differences in style, formatting, and wording, but interchangeable propositional content, assigned randomly across runs. Since small variations in wording can cause notable changes in model behaviour~\citep{sclar2024quantifying}, we included these variants to help control for sensitivity to surface phrasing without venturing into prompt engineering or optimization. Across all models, we used fixed parameters (temperature of 0.7, a token budget of 4,096, and approximately 20 sentences of requested output where possible). Since our goal is an existence argument rather than performance optimization, we did not perform parameter sweeps. Outputs were generated independently, without in-context accumulation (but see Appendix~\ref{appendix:limitationsandfuturework}). For additional details, see Appendix~\ref{appendix:constructvalidity}.

We frame our argument around existence because that is the logical structure required to refute the SSP while sidestepping common interpretive pitfalls: existence lets us ask a specific and tractable question without placing undue weight on complex unknowns such as consistency across contexts (see Appendix~\ref{appendix:constructvalidity}).\footnote{However, the degree of alignment between our hypotheses and results is stronger than that framing alone suggests. Obviously, that is a topic for Results~\ref{sec:results} and Discussion~\ref{sec:discussion}, but we felt earlier signposting might be helpful.}

\subsection{Identifying and measuring rule-following behaviour}
\label{sec:measuringrulefollowing}

Outputs were cleaned through a combination of manual effort and automation. To enhance intercleaner reliability, we discussed cleaning criteria as a team, reviewed examples together for consistency, and spot-checked 3 randomly chosen computationally-cleaned runs per rule per model to refine the process. Where possible, we applied metrics to raw outputs as well as cleaned outputs, to sanity-check that cleaning choices were not driving results.

During the preliminary testing phase, we developed automatable tests for specific aspects of rule-following -- essentially unit tests for the output -- building measurability into the rules themselves. Each test captures only some aspects of compliance, making them a lower bound on rule-following; underestimating is preferable to overestimating, since overestimating could lead to a false positive. Tests range from open-ended (story rules, double verb rules) to exact (translation rules, which can be checked against an answer key), and applied to both cleaned and raw outputs where possible. We conducted both model-level and pooled analyses, with permutation testing and corrections where warranted. See Appendix~\ref{appendix:constructvalidity} for details.

Some behavior may be visibly rule-directed without being fully captured by automated checks, but for the most part, that analysis will be done in part 2. See Appendix~\ref{sec:weirdoutputmodes} for some preliminary observations.

\subsection{Rule design}
\label{sec:ruledesign}

The rules must satisfy two requirements. First, correct responses should be difficult to explain using superficial statistical regularities. Second, the tasks should remain close to linguistic competence rather than depend primarily on unrelated capacities such as mathematics, counting, very long-distance memory, or sensory experience. 

We therefore aim for a theoretical ``sweet spot'': tasks reliant on structures that models could arguably have learned given their architecture and training data, yet whose correct outputs diverge from familiar natural-language distributions. Neural network architectures can represent data structures underlying approximately context-free and slightly context-sensitive languages, formal classes that align in theory with natural language~\citep{wilcox-etal-2019-hierarchical}. Since LLMs aren't people, we don't exactly know what they could have learned while acquiring the level of linguistic competence they exhibit, but we can consider people as a kind of outer bound: tasks that no human acquires through language exposure alone are unlikely to have been learned by PTGA models. Our conlangs therefore remain within the approximate domain of language learning while being unusual enough that explanations based solely on superficial pattern matching are not merely less parsimonious, but increasingly hard to sustain.\footnote{A caveat: As children, speakers learn tasks that look like aspects of the conlangs we use. But they do so with varying degrees of conscious accessibility. Our tasks foreground linguistic skills as conscious reasoning, which is a non-trivial difference. Furthermore, first language acquisition and second language acquisition are different for people, and are not uniform across language-related skills~\citep{malaia2020age}.}

To undermine statistical plausibility at concrete levels, we introduce linguistic features that are rare to unattested in natural language, including unusual word orders, rare English bigrams, word-order parity rules, inserted verbs, and a translation system in which prompts and correct answers share no character-level overlap. These manipulations make correct outputs atypical under familiar surface distributions while preserving the overall structure of a language task. At the same time, we avoid completely disrupting information locality, requiring mathematical reasoning, or relying on embodiment~\citep{zimmerman2025locality,lovato2024foregrounding,pb_bodyshapes,hahn2020universals,mayzner1965tables,norvig2012mayzner,Liu2024SentenceLength,kallini2024mission,futrell2020dependency,futrell2024linguistic,gibson2019how,harris1954structure,harris1968mathematical,sahlgren2008distributional,zipf1965,futrell2020lossy}. For example, we do not test counting, because linguistic competence does not require counting ability. 

In addition to designing rules that are theoretically unlike the training data, we perform contamination checks for potential training-data exposure (Section~\ref{sec:trainingdataexposure}; Appendix~\ref{appendix:trainingdataexposure}). Because this is an existence argument, success should be understood as systematic movement in the predicted direction rather than perfect rule execution. These are intentionally difficult tasks, and imperfect performance does not imply that the rule was not represented at all. We will discuss the exploratory testing in which we developed the conlangs more in part 2.

\subsection{Training data exposure}
\label{sec:trainingdataexposure}

We performed complementary checks for training data exposure. While these checks cannot prove the absence of exposure of these models to these rules, they can dramatically reduce how much likelihood we allocate to that possibility.

The results of these checks (as well as the number of models and runs per rule, the design and diversity of the rules, and the inclusion of prompt variations) make it implausible that any model we tested could have succeeded by direct recall of training data. We searched for 19 representative strings contained in our prompt and answer pairs in 8 major training corpora including OLMo, DCLM-baseline, and Dolma-v1.7; only one string, matching ``zero is the word boundary OR zero is the space symbol'', appeared in some of the training corpora (ranging from 1-4 instances). These appear to refer to wholly dissimilar contexts (see Table~\ref{tab:training-search-examples}). We used the InfiniGram N-gram LM to test whether the same training corpora assigned the highest probability to a token that was part of the correct continuation for dan-1 and translate-1 prompt variants; of 12 variations, only one condensed form of the dan-1 prompt yielded a correct prediction by any training corpus, and none when we took (fuzzy) order into account~\citep{Liu2024InfiniGram}. See Table~\ref{tab:training-search-counts}, Table~\ref{tab:top-token-correct-continuation}, and Table~\ref{tab:training-search-examples}. For additional details on these checks and the additional checks we performed, see Appendix~\ref{appendix:trainingdataexposure}.

\subsection{Conlang families and hypotheses}
\label{sec:conlangsinpart1}

In part 1, we focus on three rule families that provide complementary evidence for our core argument. In part 2, we extend the discussion to the full set of rules, including additional evidence regarding rule-following behaviour, potential cognitive tradeoffs and a more detailed exploration of output modes.\footnote{As the paper is already somewhat lengthy, we split the minimal core argument from the broader experimental results and exploratory analysis.} 
\begin{itemize}[nosep] 
    \item The story family (comparison-1, sidequest-1, comparison-2)\footnote{And sidequest-1-fix in part 2.} probes whether unusual output can be explained by prompt exposure alone, using sustained generation of unusual lexical distributions over approximately 20 sentences. 
    \item The double verb family (doublev-1, doublev-2) probes whether models can override syntactic and semantic plausibility priors through the insertion of additional irrelevant verbs.\footnote{Doublev-2 arguably also conflicts with dependency locality, since the irrelevant verb is not near anything it modifies. The rules also require some overriding of lexical and morphological priors, to mark the irrelevant verb with a specific rare bigram.} 
    \item The translation family (translate-1, dan-1)\footnote{And translate-2 in part 2.} probes composed symbolic mappings involving lexical, morphological, semantic, and word-order constraints. Dan-1 mitigates the role of character-level overlap between prompt and answer.\footnote{Additionally, translate-1 requires a metaphor map, and dan-1 uses chiasm (``Her mother's cat... His cat's father...'') to further subvert statistical plausibility.}
\end{itemize} 

Summaries of the key hypotheses for each family of rules are:
\begin{itemize}[nosep]
    \item Story family: comparison-1 $<$ sidequest-1 $<$ comparison-2 in rates of rare bigram usage\footnote{In terms of tokens, not types, in the type-token distinction. Rare bigram ratio was computed as the proportion of all output bigrams that belonged to the predefined rare bigram inventory.}, and comparison-2 should be distinguishable from both comparison-1 and sidequest-1.\footnote{Additionally, idealized rule-following would see comparison-1 and sidequest-1 be indistinguishable, but PTGA constraints should prevent models from achieving that. In practice, we expect the best rule-following to show up as comparison-1 and sidequest-1 looking much more like each other than either does to comparison-2.}
    \item Double verb family: production of double verbs; high semantic similarity across paired verbs in doublev-1, obviously lower similarity in doublev-2.
    \item Translation family: outputs should be substantially closer to the answer keys than matched random strings, with weaker performance as the representational regime becomes less familiar (moving from translate-1 to dan-1).
\end{itemize}

Together, these families span lexical, semantic, morphological, syntactic, word-order, metaphorical, discourse-level, and arguably pragmatic rule application. While any individual rule could admit rule-specific alternative explanations, the convergence of evidence across multiple qualitatively different dissociations between meaning-mediated and statistical plausibility strengthens the overall argument. Taken together, the results allow us to probe not only whether models move beyond verbatim regurgitation, but whether the patterns they rely on extend substantially beyond superficial distributional patterns. For readability, all rules are available in Appendix~\ref{appendix:allruletext}. 

\section{Results}
\label{sec:results}

Our hypotheses were specific and directional, and the observed outcomes generally match those predictions. All rules elicit rule-following behaviour from some models. All models exhibit some evidence of rule-following on all rules except the most challenging (dan-1), and the predicted ordering holds, with effect sizes frequently near ceilings, and the same qualitative pattern across cleaned and uncleaned outputs. The weakest evidence, where results are not statistically significant, comes from Llama, which is the model we expected \textit{a priori} to show the weakest signal. Table~\ref{tab:hypotheses_results} summarizes our results. Each family of rules is described in more detail below (Sections~\ref{sec:storyrules},~\ref{sec:doubleverbrules},~\ref{sec:translationrules}), and discussed in Section~\ref{sec:discussion}.

We observe that model responses can be multimodal. Anecdotally, we observe some failures that metaphorically evoke attractors in dynamical systems: initial trajectories of varying durations, eventually collapsing into degenerate cycles. Some initial observations as to such output modes are given in Appendix~\ref{sec:weirdoutputmodes}, though we mostly defer to part 2.

\begin{table*}[h] \centering \small
\caption{Tested conlangs grouped by family, with hypotheses, tests, and summarized findings.} 
\label{tab:hypotheses_results} 
\begin{adjustbox}{width=\textwidth}
\begin{tabular}{| p{1.7cm}| p{2.8cm}| p{3.8cm}| p{3.8cm}| >{\columncolor{green!15}}p{1.4cm}|}
\hline \textbf{Rule(s)} & \textbf{Description} & \textbf{Hypothesis} & \textbf{Test} & \textbf{Y/N}\\
\hline \multicolumn{5}{|l|}{\textit{Family: Story rules}} \\
\hline sidequest-1, comparison-1, comparison-2 & comparison-2 should show a preponderance of rare bigrams versus the other two rules, ruling out prompt-driven echoing as a sufficient explanation. & Rare bigram usage should be higher for comparison-2 than sidequest-1 and comparison-1 (\textit{use} versus \textit{mention} of rare bigrams should yield distinguishable output); ordering should be comparison-1 $<$ sidequest-1 $<$ comparison-2; idealized rule-following should see comparison-1 and sidequest-1 rare bigram usage be indistinguishable, but PTGA should prevent this in practice. & Rate of rare bigram usage comparison across conditions; see Figure~\ref{fig:comparison1comparison2sidequest1box_clean_plot}. Additionally:
Fig.~\ref{fig:comparison1comparison2sidequest1box_allmodeloutput_plot},
Tables~\ref{tab:fisher_combined},
~\ref{tab:storyrulesfisher_combined},
~\ref{tab:storyrules_cleaned_mannwhitneyu},
~\ref{tab:storyrules_uncleaned_mannwhitneyu},
~\ref{tab:storyrules_summary_mean_cleaned},
~\ref{tab:storyrules_cleaned_ks},
~\ref{tab:storyrules_uncleaned_ks},
~\ref{tab:pairedwilcoxon},
~\ref{tab:pairedwilcoxon3models},
~\ref{tab:summarystatsstoryrules}.
& Y (for all cleaned outputs: Qwen, GPT-4o, OLMo; Y for some models for uncleaned outputs)\\
\hline \multicolumn{5}{|l|}{\textit{Family: Double verb rules}} \\
\hline doublev-1, doublev-2 & Models must use a novel double-verb structure; doublev-2 additionally emphasizes that the second verb should be semantically unrelated to the first. & (1) Production of double verbs; (2) Ratio of verbs marked with \textit{-qw} to total verbs $\approx 0.5$; (3) verb pairs in doublev-1 have higher cosine similarity than in doublev-2. & (1 and 2) Count; (3) KS test and Wilcoxon Rank Test for doublev-1 $>$ doublev-2 cosine similarity. See Tables~\ref{tab:ks_rules_cosine},
~\ref{tab:wilcoxon_rules_cosine}, Additionally: Figures~\ref{fig:doublev-1-histograms},~\ref{fig:doublev-2-histograms}, Tables~\ref{tab:wilcoxon_doublev1_cosine},~\ref{tab:ks_doublev1_cosine},~\ref{tab:wilcoxon_doublev2_cosine},~\ref{tab:ks_doublev2_cosine}, Figures~\ref{fig:doublevcosine},
~\ref{fig:all-models-doublev1},
~\ref{fig:all-models-doublev2}. & (1) Y for all models; (2) Y for some models; \textbf{(3) Y for all models}\\
\hline \multicolumn{5}{|l|}{\textit{Family: Translation rules}} \\
\hline translate-1 & Complex novel translation task. & (1) String distance measures -- output Levenshtein distances, normalized Levenshtein, Jaccard, and cosine similarity -- should be closer to the answer key than randomized strings (select-character and all-character). (2) Exact matches to answer key. & KS test and Wilcoxon Rank-Sum on cosine string similarity vs.\ random strings; string distance comparisons; regex. See:
Figures~\ref{fig:translate-1-histograms},
~\ref{fig:all-models-translate-1},
Tables~\ref{tab:wilcoxon_cosine_translate-1},
~\ref{tab:ks_cosine_translate-1}.
& \textbf{(1) Y for all models}; (2) Y for some models\\
\hline dan-1 & Complex translation task. Uses no characters from the prompt, controlling for token exposure. & (1) Output closer to correct answer than random strings; (2) Exact matches. & KS test and Wilcoxon Rank-Sum on cosine similarity vs.\ random strings; string distance comparisons; regex. See:
Figures~\ref{fig:dan-1-histograms},
~\ref{fig:all-models-dan-1},
Tables~\ref{tab:wilcoxon_cosine_dan-1},
~\ref{tab:ks_cosine_dan-1}.
& (1) Y for some models; (2) Y for some models\\
\hline translate-1, dan-1 & Supplementary check to support the above results via established machine translation metrics (BLEU, chrF, TER, METEOR). & Similar cosine similarity findings across both rules; translation metric scores approach optimal (BLEU\,=\,100, chrF\,=\,100, TER\,=\,0, METEOR\,=\,99.6). & Cross-rule comparison of cosine similarity; translation benchmarks against reference. See Figures~\ref{fig:all_models_translation_performance},~\ref{fig:all_models_source_length_truncated_translation_performance}. 
& Y for some models\\
\hline \end{tabular} \end{adjustbox} \end{table*}

\subsection{Story rules}
\label{sec:storyrules}

This family controls for the possibility that unusual output is caused by exposure to unusual prompt content rather than by rule-following. We distinguish three regimes: 
\begin{itemize}[nosep] 
    \item Baseline: In comparison-1, we request ordinary English story generation of about 20 sentences.
    \item Prompt exposure: In sidequest-1, rare bigrams~\citep{norvig2012mayzner,mayzner1965tables} are presented in the prompt, but the model is instructed to write an ordinary English story. 
    \item Rule-following: In comparison-2, the same rare bigrams are described as the morphemes of a new language and the model is instructed to produce a story of about 20 sentences in that language. Both sidequest-1 and comparison-2 contain the same list of rare bigrams in the prompt. 
\end{itemize} 

An alternative explanation to rule-following is that symbols introduced in a prompt tend to recur in subsequent output, independent of their intended role.\footnote{For example, introducing a topic such as baseball tends to increase the frequency of baseball-related terminology later in the conversation. If a person is introduced with a male pronoun, male pronouns are likely to recur. Such regularities may help bridge syntax, semantics, and pragmatics.} If prompt exposure alone were sufficient to explain conlang performance, sidequest-1 and comparison-2 should produce similarly elevated rates of rare-bigram use.

Because of competing plausibilities, we expect the ordering of rare bigram use to be (1) comparison-1 $<$ sidequest-1 $<$ comparison-2. Rule-following behaviour should at least satisfy: (2) rare bigram usage in comparison-1 and comparison-2 are distinguishable, and (3) rare bigram usage in sidequest-1 and comparison-2 are distinguishable. An idealized account of rule-following -- in which only instructed use matters and prompt exposure has no effect (which we don't actually expect to see, since we think in practice the plausibilities compete) -- would additionally satisfy (4) sidequest-1 and comparison-1 are indistinguishable, since merely encountering the rare bigrams would have no effect. The cleaned results should be more dramatic than the uncleaned results (since any additional English text generated reduces the ratio of rare bigrams).

All three models (GPT-4o, Qwen (thinking), OLMo) with cleaned outputs satisfy the predicted ordering (1) with comparison-2 strongly distinguishable from both controls (2 and 3). The same holds with the uncleaned outputs for four of the five models; Llama's rule-following (comparison-2) outputs differ from baseline (comparison-1) in the expected direction but are not reliably distinguishable from its prompt exposure (sidequest-1) outputs. As expected, no model exhibits idealized rule-following (4): prompt exposure has a detectable effect in every model. However, the size of that effect varies substantially. GPT-4o comes closest to the ideal pattern, with sidequest-1 remaining much closer to baseline than to rule-following output.

That inclusion in the prompt is not generally sufficient to cause inclusion in the output weakens prompt exposure as a general alternative explanation for rule-directed behaviour. Although the prompts are otherwise similar, not identical, our results suggest the interpretation that the same surface forms -- the lists of rare bigrams -- can produce different behaviour in LLMs depending on their communicative role within the prompt. Because the data contain many tied and near-zero values, we use permutation tests to estimate combined p-values while minimizing distributional assumptions. Combined results are reported in Tables~\ref{tab:fisher_combined},~\ref{tab:storyrulesfisher_combined}. Additional results are reported in Appendix~\ref{appendix:storyresults}.

\begin{figure}
  \centering
  \includegraphics[width=0.6\columnwidth]
  {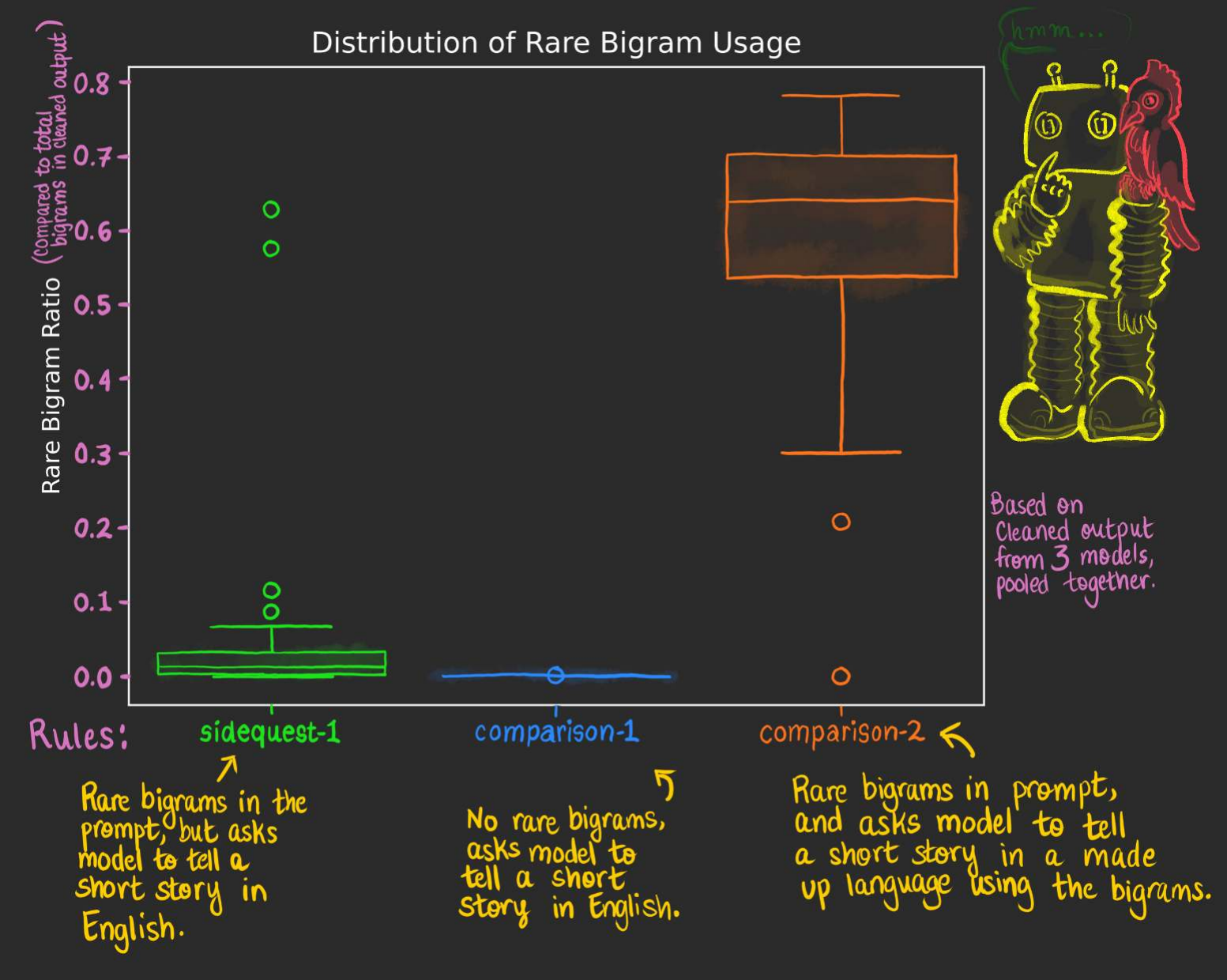}
  \caption{Comparison of rare bigram usage across conditions, for the 3 models (Qwen (with thinking), GPT-4o, OLMo) with cleaned output.
}
\label{fig:comparison1comparison2sidequest1box_clean_plot}
\end{figure}

\begin{table}
\centering
\caption{Fisher-combined $p$-values across models (Qwen (T), Olmo, GPT-4o (no SP)) for each contrast; cleaned outputs.}
\label{tab:fisher_combined}
\begin{tabular}{lrrr}
\toprule
Comparison & Fisher statistic ($X^2$) & $p_{combined}$ & $k$ \\
\midrule
comparison-2 vs sidequest-1 & 226.794 & 3.700e-46 & 3 \\
sidequest-1 vs comparison-1 & 187.022 & 1.093e-37 & 3 \\
comparison-2 vs comparison-1 & 243.473 & 1.017e-49 & 3 \\
\bottomrule
\end{tabular}
\end{table}

\begin{table}
\centering
\caption{Fisher-combined p-values across all models for each contrast (uncleaned output/ all output).}
\label{tab:storyrulesfisher_combined}
\begin{adjustbox}{width=\textwidth}
\begin{tabular}{lrrr}
\toprule
Comparison & Fisher statistic ($X^2$) & $p_{combined}$ & $k$ \\
\midrule
comparison-2 vs sidequest-1 & 303.771 & 2.479e-59 & 5 \\
sidequest-1 vs comparison-1 & 290.608 & 1.501e-56 & 5 \\
comparison-2 vs comparison-1 & 349.397 & 5.350e-69 & 5 \\
\bottomrule
\end{tabular}\end{adjustbox}
\end{table}

\begin{table}
\centering
\begin{tiny}
\caption{\textbf{Mann-Whitney U tests (i.e. Wilcoxon Rank-Sum) for cleaned model outputs for sidequest-1, comparison-1, and comparison-2 (cleaned outputs available for GPT-4o, Qwen (thinking), and OLMo).} Note that GPT-4o shows the closest to ideal performance here (sidequest-1 and comparison-1 outputs clearly more similar to each other than comparison-2 output is to either).}
\label{tab:storyrules_cleaned_mannwhitneyu}
\begin{adjustbox}{width=\textwidth}\begin{tabular}{llrllllll}
\toprule
Model & Comparison & U & p & p (BH) & Cliff's Delta & Median A & Median B & Median Diff \\
\midrule
gpt4o (no SP) & sidequest-1 vs comparison-1 & 1980 & 3.05e-08 & 3.05e-08 & 0.584 & 0.001982 & 0.000000 & 0.001982 \\
gpt4o (no SP) & comparison-2 vs comparison-1 & 2500 & 5.42e-19 & 3.94e-18 & 1.000 & 0.565855 & 0.000000 & 0.565855 \\
gpt4o (no SP) & comparison-2 vs sidequest-1 & 2500 & 6.06e-18 & 1.36e-17 & 1.000 & 0.565855 & 0.001982 & 0.563873 \\
olmo & sidequest-1 vs comparison-1 & 2406 & 6.11e-16 & 6.88e-16 & 0.925 & 0.005576 & 0.000000 & 0.005576 \\
olmo & comparison-2 vs comparison-1 & 2465 & 1.90e-17 & 3.41e-17 & 0.972 & 0.706837 & 0.000000 & 0.706837 \\
olmo & comparison-2 vs sidequest-1 & 2450 & 1.31e-16 & 1.68e-16 & 0.960 & 0.706837 & 0.005576 & 0.701260 \\
qwen (T) & sidequest-1 vs comparison-1 & 2500 & 1.31e-18 & 3.94e-18 & 1.000 & 0.020997 & 0.000000 & 0.020997 \\
qwen (T) & comparison-2 vs comparison-1 & 2500 & 1.31e-18 & 3.94e-18 & 1.000 & 0.629056 & 0.000000 & 0.629056 \\
qwen (T) & comparison-2 vs sidequest-1 & 2461 & 7.12e-17 & 1.07e-16 & 0.969 & 0.629056 & 0.020997 & 0.608059 \\
\bottomrule
\end{tabular}\end{adjustbox}
\end{tiny}
\end{table}

\begin{table}[]
    \centering
    \tiny
    \caption{Mann-Whitney U comparisons of rare bigram usage medians for all sidequest-1, comparison-1, comparison-2 output (uncleaned outputs, available for all 5 models).}
    \label{tab:storyrules_uncleaned_mannwhitneyu}
\begin{adjustbox}{width=\textwidth}
\begin{tabular}{llrllllll}
\toprule
model & comparison (A-B) & U & p & $p_{bh}$ & Cliff's $\delta$ & Median A & Median B & $\tilde{x}_A-\tilde{x}_B$ \\
\midrule
gpt4o (no SP) & sidequest-1 vs comparison-1 & 1980 & 3.05e-08 & 3.81e-08 & 0.584 & 0.001982 & 0.000000 & 0.001982 \\
gpt4o (no SP) & comparison-2 vs comparison-1 & 2500 & 5.42e-19 & 7.48e-18 & 1.000 & 0.565855 & 0.000000 & 0.565855 \\
gpt4o (no SP) & comparison-2 vs sidequest-1 & 2500 & 6.06e-18 & 1.14e-17 & 1.000 & 0.565855 & 0.001982 & 0.563873 \\
llama & sidequest-1 vs comparison-1 & 1649 & 1.26e-03 & 1.35e-03 & 0.319 & 0.000000 & 0.000000 & 0.000000 \\
llama & comparison-2 vs comparison-1 & 1812 & 1.21e-05 & 1.40e-05 & 0.450 & 0.000417 & 0.000000 & 0.000417 \\
llama & comparison-2 vs sidequest-1 & 1414 & 2.34e-01 & 2.34e-01 & 0.131 & 0.000417 & 0.000000 & 0.000417 \\
olmo & sidequest-1 vs comparison-1 & 2500 & 2.49e-18 & 7.48e-18 & 1.000 & 0.030387 & 0.000000 & 0.030387 \\
olmo & comparison-2 vs comparison-1 & 2500 & 2.49e-18 & 7.48e-18 & 1.000 & 0.328400 & 0.000000 & 0.328400 \\
olmo & comparison-2 vs sidequest-1 & 2495 & 9.54e-18 & 1.59e-17 & 0.996 & 0.328400 & 0.030387 & 0.298013 \\
qwen (T) & sidequest-1 vs comparison-1 & 2500 & 4.72e-18 & 1.01e-17 & 1.000 & 0.030265 & 0.000254 & 0.030011 \\
qwen (T) & comparison-2 vs comparison-1 & 2500 & 4.72e-18 & 1.01e-17 & 1.000 & 0.075579 & 0.000254 & 0.075325 \\
qwen (T) & comparison-2 vs sidequest-1 & 2424 & 5.97e-16 & 8.14e-16 & 0.939 & 0.075579 & 0.030265 & 0.045314 \\
qwen (no T) & sidequest-1 vs comparison-1 & 2500 & 1.74e-18 & 7.48e-18 & 1.000 & 0.041571 & 0.000000 & 0.041571 \\
qwen (no T) & comparison-2 vs comparison-1 & 2500 & 1.74e-18 & 7.48e-18 & 1.000 & 0.395845 & 0.000000 & 0.395845 \\
qwen (no T) & comparison-2 vs sidequest-1 & 2450 & 1.35e-16 & 2.02e-16 & 0.960 & 0.395845 & 0.041571 & 0.354274 \\
\bottomrule
\end{tabular}\end{adjustbox}
\end{table}

\subsection{Double verb rules}
\label{sec:doubleverbrules}

The double verb family instructs models to add a second, irrelevant verb to each main verb and mark it with the affix -qw. In doublev-1, the added verb is just described as irrelevant. In doublev-2, we additionally specify that it should be semantically unrelated to the first verb and feel random relative to the rest of the sentence. Otherwise, the rules are the same. Because verbs occurring near one another in English are typically semantically related, and at least related to the words around them, the second rule pushes against strong plausibility priors (both syntactic and semantic) learned from natural-language data, and violates the principle of least effort~\citep{zipf1965,gibson2019how,futrell2020dependency}.\footnote{Doublev-2 may also conflict with dependency locality, since especially when the second verb is irrelevant in the sense specified in doublev-2, it is arguably not near anything it modifies.} Comparing cosine similarity between verb pairs from doublev-1 to doublev-2 therefore provides a direct test of whether models alter generation in the requested semantic direction.

\begin{figure}
  \centering
  \includegraphics[width=0.6\columnwidth]{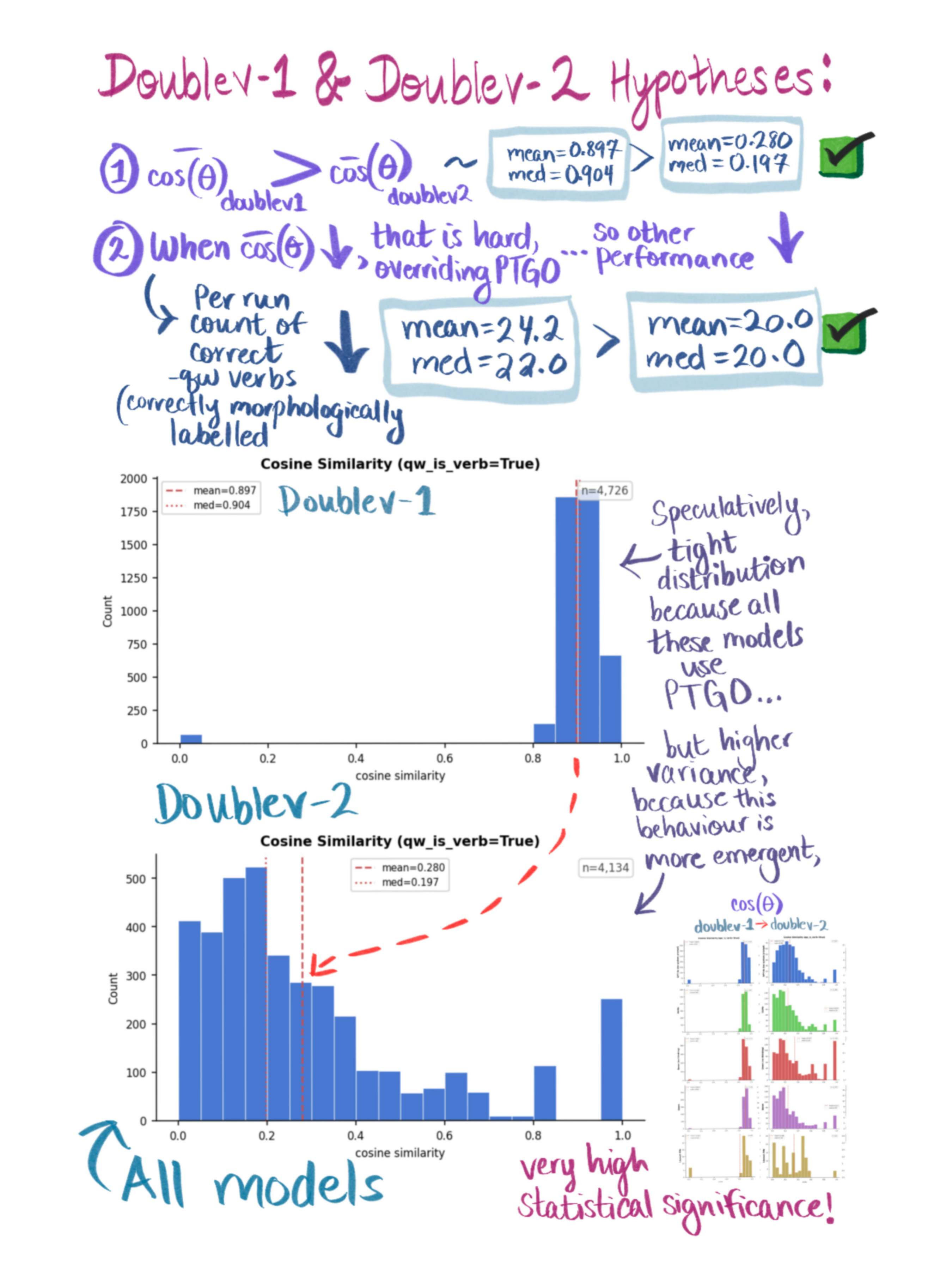}
  \caption{Illustrative summary of double verb family results from our poster presentation~\citep{ISC2026SummerSchool}, Fig~\ref{fig:Isc2026poster3.pdf}.}
  \label{fig:doublevcosine}
\end{figure}

Rule-following should result in the production of additional verbs, about half of which should be marked with the instructed rare bigram morpheme.\footnote{Successful rule-following output should yield a total count of verbs per run that is twice the number of verbs marked with the -qw morpheme. Since doublev-2 emphasizes the irrelevance of the added verb, we expect that cosine similarity between verbs should be a lower value (e.g. closer to $0$) for the verbs produced by this prompt than for doublev-1. Since we think doublev-2 is harder, we expect the ratio of -qw verbs to go down, and the total number of verbs produced to go down, compared to doublev-1. In part 1 we have not tracked whether -qw marks the \textit{better} choice of ``irrelevant'' verb, where present. We may explore that in part 2.} The relationships between the pairs of verbs should further satisfy specific hypotheses. First, the cosine similarities between paired verbs should be substantially lower in doublev-2 than in doublev-1. Second, our PTGA account predicts that doublev-1 should often exhibit very high cosine similarities, because local plausibility priors strongly favor semantically coherent contexts.\footnote{From the perspective of PTGA, generating closely related, even identical, verb pairs is one of the easiest ways to satisfy the rule.} In addition, because we believe doublev-2 to be harder, we expect we may see reduced performance on some other aspects of the task, suggesting a potential tradeoff amongst limited resources, although we mostly defer that analysis to part 2.

\begin{figure}
    \centering
    \includegraphics[width=0.8\textwidth]{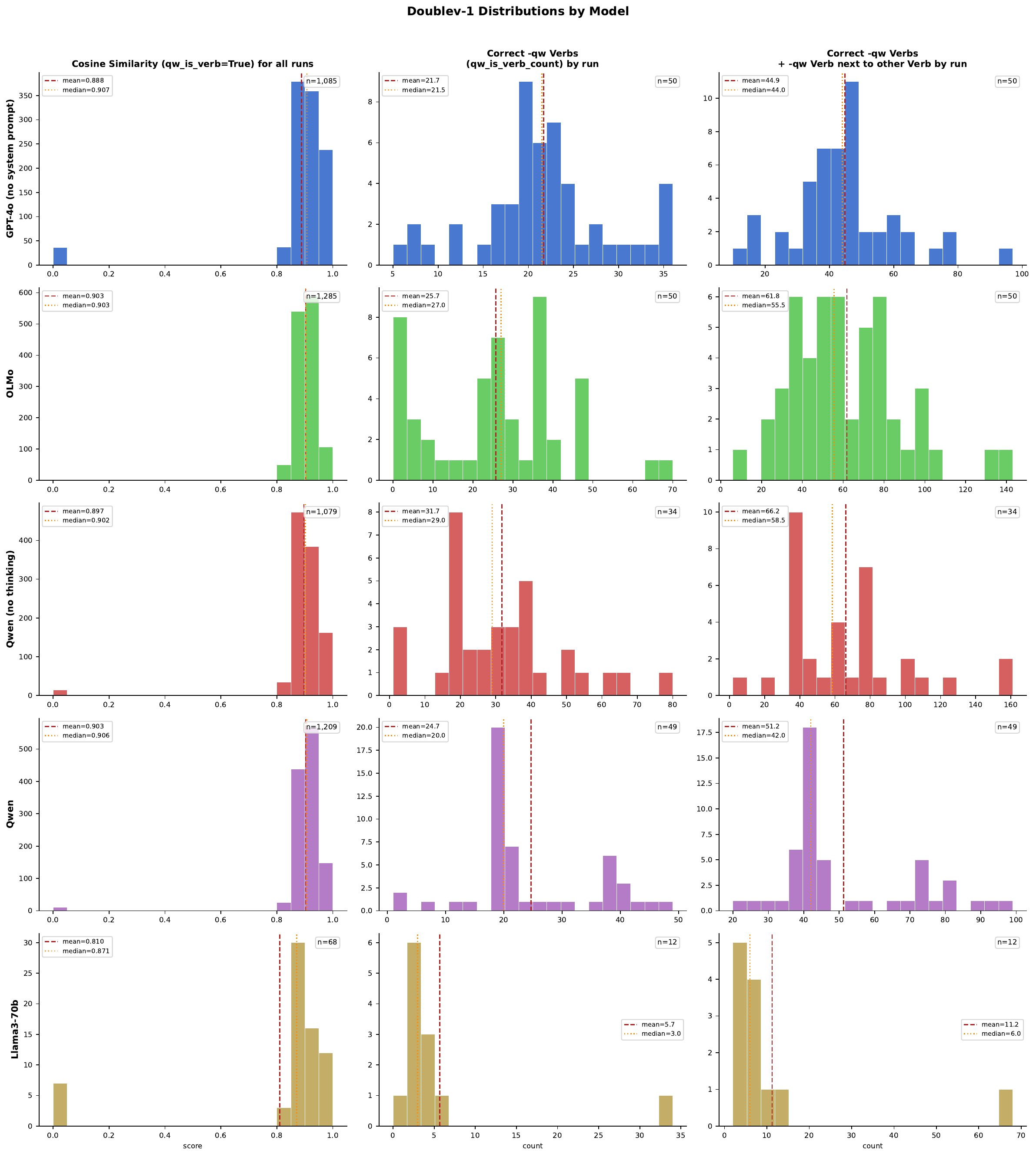}
    \caption{\textbf{Doublev-1 Rule Similarity Scores and Verb Counts by Model.} Histogram of similarity scores (cosine similarity between marked and closest verb, where marked verb is appropriately terminated with bigram \textit{-qw}) per model, as well as the per-run count of correctly marked \textit{-qw} verbs and count of correct pairs of \textit{-qw} verbs and closest verb. Total number of runs with verbs reported for each model ($n=50$ runs indicates each run had at least one instance of a correctly assigned -qw verb). Cosine similarity scores closer to 0 indicate dissimilar words. Counts by model that are closer to total clauses reported by model indicate higher performance by model.}
    \label{fig:doublev-1-histograms}
\end{figure}

\begin{figure}
    \centering
    \includegraphics[width=0.8\textwidth]{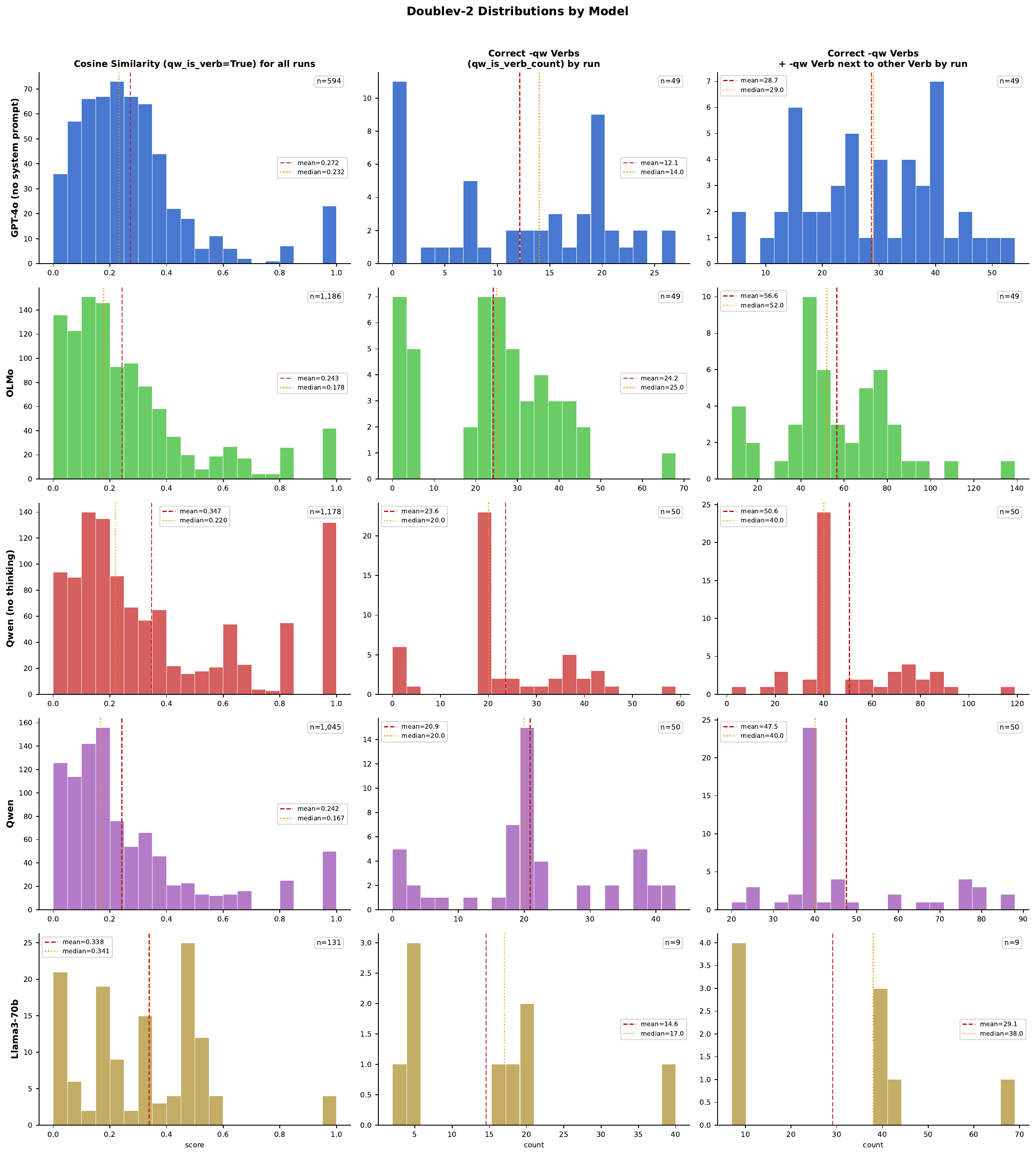}
    \caption{\textbf{Doublev-2 Rule Similarity Scores and Verb Counts by Model.} Histogram of similarity scores (cosine similarity between marked and closest verb, where marked verb is appropriately terminated with bigram \textit{-qw}) per model for all verbs paired in all runs, as well as the per-run count of correctly marked \textit{-qw} verbs and count of correct pairs of \textit{-qw} verbs and closest verb. Total number of runs with verbs reported for each model ($n=50$ runs indicates each run had at least one instance of a correctly assigned -qw verb). Cosine similarity scores closer to 0 indicate dissimilar words. Counts by model that are closer to total clauses reported by model indicate higher performance by model.}
    \label{fig:doublev-2-histograms}
\end{figure}

For each model, we observe the production of some double verb pairs, and that cosine similarity scores are lower for doublev-2 than doublev-1, shown in Figures~\ref{fig:doublev-1-histograms} and ~\ref{fig:doublev-2-histograms}; this same pattern holds when we combine all model output, as shown in Figures~\ref{fig:all-models-doublev1} and ~\ref{fig:all-models-doublev2}. Models vary in their ability to perform the task of marking verbs with -qw, but largely placed marked -qw verbs near other verbs when successfully marking a verb. When we clarified the prompt to enforce semantic randomness, semantic performance improved, even if other aspects may have degraded -- we do see a reduction in per-run median and mean counts of correctly-labeled verbs from doublev-1 to doublev-2, but will explore potential cognitive tradeoffs in part 2.

Additional results for the double verb family of rules are given in Appendix~\ref{appendix:doubleverbresults}: Table~\ref{tab:wilcoxon_rules_cosine}, Table~\ref{tab:ks_rules_cosine}, Table~\ref{tab:ks_doublev1_cosine}, Table~\ref{tab:ks_doublev2_cosine}, Table~\ref{tab:wilcoxon_doublev1_cosine}, Table~\ref{tab:wilcoxon_doublev2_cosine}.

Overall, models alter generation in the requested syntactic and semantic directions while exhibiting model-dependent difficulty in doing so (Figures~\ref{fig:doublev-1-histograms} and ~\ref{fig:doublev-2-histograms}). This pattern is consistent with our hypothesis that, in PTGA, meaning-mediated behaviour can coexist with persistent autoregressive plausibility constraints, which aligns with observations reported in \citet{mccoy2023embersautoregressionunderstandinglarge}.

\subsection{Translation rules}
\label{sec:translationrules}

Unlike the story and double verb families, translation outputs can be evaluated against explicit answer keys. Successful performance therefore requires not merely producing text with unusual features, but preserving specific relationships under tightly constrained transformations. Correct outputs preserve abstract relationships under a series of composed constraints while (sometimes dramatically) changing the symbols that realize those relationships. Preserving tense, possession, kinship, grammatical roles, and ordering while radically changing the symbols that express them is much more naturally explained by representations of the underlying relations than by superficial token distributions.

We investigate two translation conlangs: translate-1, which uses alphabetic characters, and dan-1, which relies primarily on numbers and punctuation. These are the most complex rules in our suite, both in the number of constraints and in the number of relationships among those constraints. Our research question is: can a model preserve relations across disjoint forms, under multiple composed constraints operating at different linguistic levels, even when tokens must be used contrary to their familiar patterns of use? 
\begin{itemize}[nosep] 
\item Translate-1 combines rare bigrams (phonotactically and orthographically implausible) with composed lexical, morphological, syntactic, semantic, and metaphorical mappings, including a temporal/kinship metaphor. 
\item Dan-1 combines similarly composed mappings while eliminating overlap between the symbols used in the prompt and those used in the correct output. For example, the prompt uses ``two'' while the output requires ``2'', and standard word boundaries and punctuation must be remapped.
\end{itemize} Together, these rules require simultaneous mappings across lexical items, tense, word order, morphology, semantic and emotional concepts, and relational structure, with approximately 15 constraints in translate-1 and 17 in dan-1.

The translation family exploits many overlapping strategies to undermine statistical plausibility at various linguistic scales. For an example from translate-1, in the first system, ``-qw indicates past tense, -jx indicates present tense, and -kz indicates future tense,'' but in the second system, ``-jx indicates past tense, -kz indicates present tense, and -qw indicates future tense''. Note that merely echoing these morphemes without cyclically translating their tense will not correctly preserve meaning. For an example from dan-1, chiasm (``her mother's cat ... his cat's father ...'') further targets statistical semantic plausibility. 

We therefore expect dan-1 likely to be more difficult than translate-1, since it requires symbols, word boundaries, and punctuation marks to be repurposed in unfamiliar ways while simultaneously satisfying an approximately larger set of composed constraints.

\begin{figure}
    \centering
    \includegraphics[width=0.8\textwidth]{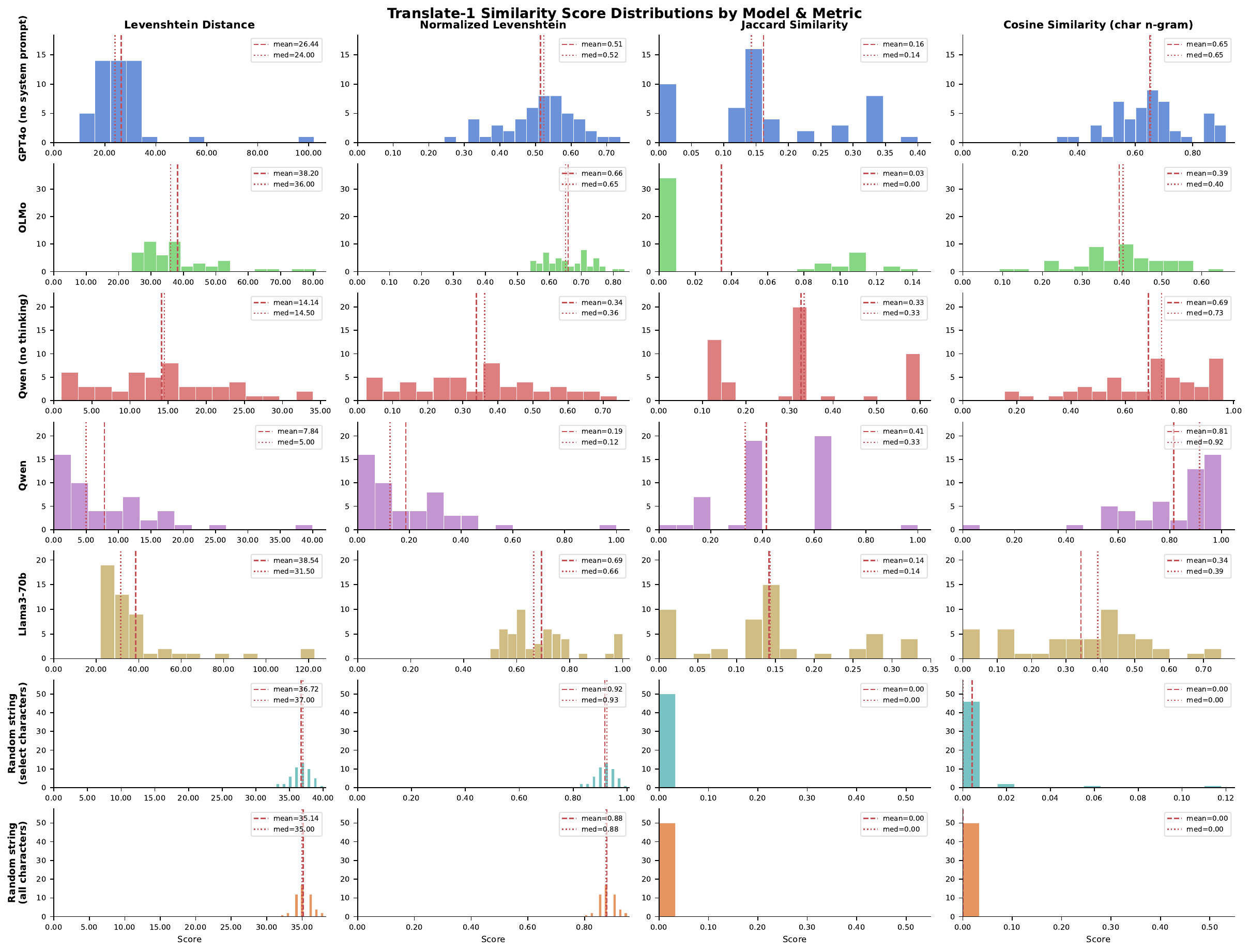}
    \caption{\textbf{Translate-1 Distributions by Model and Metric.} Histogram of string distance scores (Levenshtein distance, normalized Levenshtein distance, Jaccard similarity, and character n-gram cosine similarity) comparing the output for Translate-1 rule versus the correct translated response. Scores closer to 0 for Levenshtein distances, and scores closer to 1 for Jaccard and cosine similarities indicate better rule-following/more accurate translation. Qwen (thinking) performs best for Translate-1 rule-following based on median and mean scores for all string similarity measures. For both Wilcoxon Rank-Sum and Kolmogorov-Smirnov tests, all models perform better than the random strings variants achieving statistically significantly better median scores (reported in Tables~\ref{tab:wilcoxon_cosine_translate-1} and ~\ref{tab:ks_cosine_translate-1}).}
    \label{fig:translate-1-histograms}
\end{figure}

\begin{figure}
    \centering
    \includegraphics[width=0.8\textwidth]{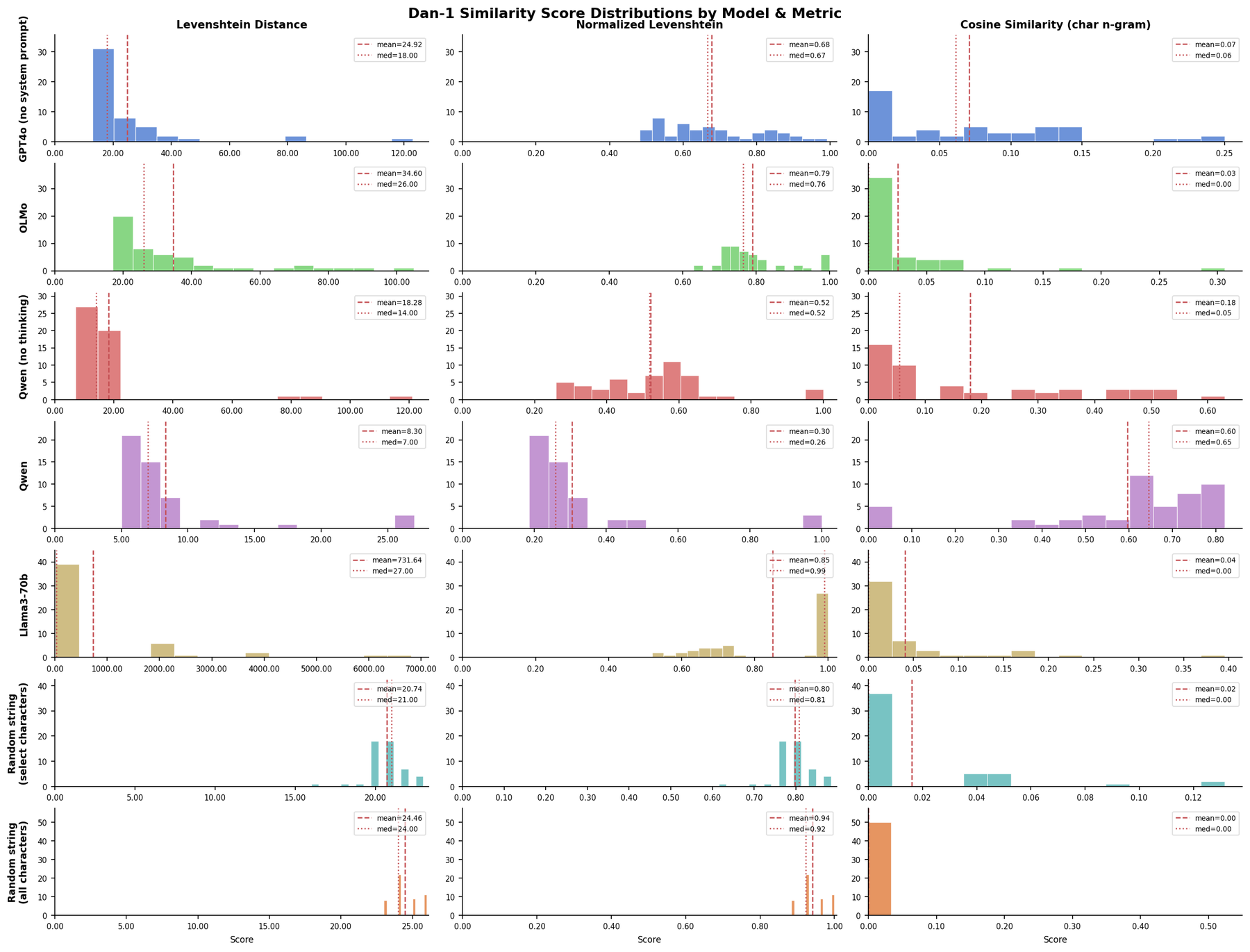}
        \caption{\textbf{Dan-1 Distributions by Model and Metric.} Histogram of string distance scores (Levenshtein distance, normalized Levenshtein distance, character n-gram cosine similarity) comparing the output for Dan-1 rule versus the correct translated response. Scores closer to 0 for Levenshtein distances, and scores closer to 1 for cosine similarities indicate better rule-following/more accurate translation. Qwen (thinking) performs best for Dan-1 rule-following based on median and mean scores for all string similarity measures. Median OLMo and Llama3-70B scores against random string (select characters) median cosine similarity scores are not significantly different for the Wilcoxon Rank-Sum Tests (reported in Table~\ref{tab:wilcoxon_cosine_dan-1}). Likewise, for the Kolmogorov-Smirnov test, which examines the distributional shapes of the cosine similarity scores, neither OLMo, nor Llama3-70B show statistically different scores than the random string (select characters). The distribution of OLMo cosine similarity scores additionally shows no statistical difference with the random string (all characters) using the Kolmogorov-Smirnov test (reported in Table~\ref{tab:ks_cosine_dan-1}).} 
    \label{fig:dan-1-histograms}
\end{figure}

All models exhibit some rule-following behaviour on translate-1 relative to random controls, with GPT-4o and Qwen (thinking) generally outperforming the other models. As in several of our analyses, the commercial models tend to outperform the open ones (Figures~\ref{fig:translate-1-histograms},~\ref{fig:dan-1-histograms}).

However, performance declines sharply in dan-1, which uniquely separates the models. Qwen (thinking) clearly outperforms the other models, with a median n-gram cosine similarity of 0.6455, compared to the next-best score of 0.0615 for GPT-4o. OLMo and Llama have median cosine similarities of 0 on dan-1, indicating no detectable rule-following behaviour by that metric, although some responses appear qualitatively potentially consistent with attempted rule application (Table~\ref{table:translate1_clean_output_table}, Table~\ref{table:dan1_clean_output_table}). See Tables~\ref{tab:wilcoxon_cosine_dan-1} and~\ref{tab:wilcoxon_cosine_translate-1}. The contrast between translate-1 and dan-1 is consistent with PTGA remaining strongly sensitive to surface forms: Performance degrades sharply as successful completion requires increasingly extensive remapping of the necessary output structures (see Sections~\ref{sec:notsp},~\ref{sec:modulatingptga} for discussion).

Translation performance also varies substantially under standard machine-translation metrics (BLEU, chrF, TER, METEOR, and a composite of the four). No model approaches perfect performance, but the same qualitative pattern remains visible: GPT-4o and both versions of Qwen cluster together at the higher-performing end, while OLMo and Llama form a lower-performing cluster (Fig.~\ref{fig:all_models_translation_performance}). Analyses of truncated model outputs are reported in Appendix Fig.~\ref{fig:all_models_source_length_truncated_translation_performance}). 

Together with the string-distance measures in Figures~\ref{fig:all-models-translate-1} and~\ref{fig:all-models-dan-1}, these results suggest substantial but incomplete rule-following ability, with performance degrading as the required mappings become increasingly unfamiliar. Additional statistical analyses are reported in Appendix~\ref{appendix:translationrulesresults}, where Tables~\ref{tab:wilcoxon_cosine_translate-1},~\ref{tab:ks_cosine_translate-1},~\ref{tab:wilcoxon_cosine_dan-1},~\ref{tab:ks_cosine_dan-1} show Wilcoxon Rank-Sum and Kolmogorov-Smirnov tests for the string cosine distances across models for the translation rules.

Overall, these results show that models can sometimes generate text according to complex, composed constraints. Qwen (thinking) occasionally produced exact matches to our answer keys. Models can preserve relationships involving tense, possession, kinship, grammatical roles, and word order while dramatically altering the surface forms that express those relationships. Because these tasks have explicit answer keys and are therefore substantially less open-ended than the story rules,\footnote{Although in Part 1 we searched for exact matches to our answer keys because they are sufficient for an existence proof, and used string-distance measures to identify near matches, the answer keys are not necessarily unique.} they demonstrate that rule-following extends beyond the generation of unusual distributions. At the same time, performance declines as the model has less opportunity to leverage familiar structures, with numerical/punctuation representations appearing harder than alphabetic representations with typical English-like spacing and punctuation.

The same systems that perform best on translate-1, namely both Qwen models and GPT-4o, also perform best on dan-1, whereas Llama and OLMo do not exhibit clear rule-following on dan-1. However, performance declines substantially for all models as the representational regime becomes less familiar. GPT-4o and Qwen (no thinking), which perform well on translate-1, perform dramatically worse on dan-1, albeit still distinguishably from random controls (and Llama and OLMo). This pattern suggests that architectural differences,\footnote{Broadly construed, including both model architecture and the structure of training and post-training.} affect the ability to maintain composed relationships under increasingly unfamiliar representational regimes (see \citet{GanBlackBoxSurvey2026} for a survey of theories of LLM performance divergence). We explore these possibilities further in Section~\ref{sec:performance}.

\subsection{Exploratory performance results}
\label{sec:performance}

\textbf{Caveat: Because of the small number of models and the difficulty of comparing across different metrics and rule families, these results are exploratory. They are more like back-of-the-envelope calculations: We think the gist is potentially meaningful, but we would not put much stock in the exact values, for the reasons described below.}

Performance appears to broadly track rule complexity, although we defer systematic investigation of this relationship to future work. We include exploratory analyses below; see Appendix~\ref{appendix:performanceresults} for additional details. 

\begin{figure}
    \centering
    \includegraphics[width=0.8\textwidth]{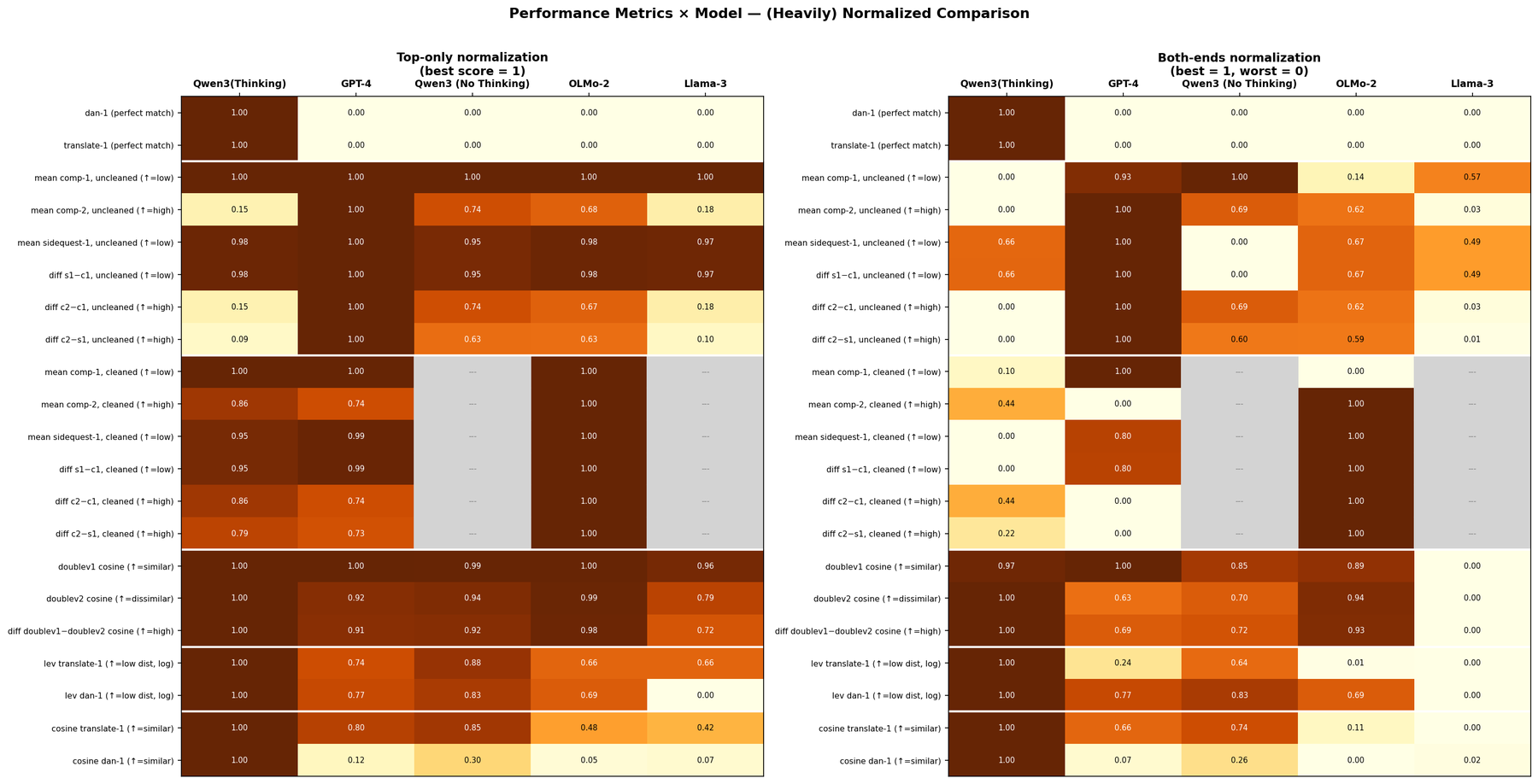}
    \caption{\textbf{Heavily normalized performance subset.} Comparison-1 is a baseline, asking the model to generate a short story in English. All models perform well. For non-thinking models, cleaned and uncleaned outputs are nearly identical. For Qwen (thinking), the uncleaned rare bigram rate is suppressed by the CoT trace, which is in standard English. Levenshtein scores are normalized twice here. All performance metrics are reoriented so that higher values indicate better performance, then normalized to $[0,1]$ in two versions for comparison: \textit{top-only}, where each row is divided by its maximum observed value (best model $= 1$, others scaled relative to it); and \textit{both-ends}, where each row is min-max normalized (best $= 1$, worst $= 0$). Results are visualized as a heatmap with models as columns and metrics as rows, with missing values (models lacking cleaned output data) grayed out. The former gives a clearer picture of overall model performance (for each rule, which models were comparatively able to do it), while the latter gives a clearer picture of relative model performance (for each rule, which model did best and worst). The comparatively un-normalized version of this performance heatmap is included in the Appendix in Figure~\ref{fig:performance_heatmap_mostlyunnormed}.}
    \label{fig:performance_heatmap_comparison}
\end{figure}

We normalize performance scores across rule metrics and compare models using both maximum and minimum-maximum normalization procedures (Fig.~\ref{fig:performance_heatmap_comparison}). Overall, commercial models such as GPT-4o and Qwen generally outperform the open models OLMo and Llama, and (as expected) cleaned outputs yield higher scores than uncleaned outputs. We also observe lower performance on rules that we judged to be more difficult, such as the progression from doublev-1 to doublev-2. We include both highly-normalized (Figure~\ref{fig:performance_heatmap_comparison}) and much-less-normalized (Figure~\ref{fig:performance_heatmap_mostlyunnormed}) results for a richer picture of model performance. 

Across models, the relative ordering of rule difficulty is notably stable. The main model differences are in absolute performance level rather than in which rules models find relatively harder or easier. However, rule difficulty is inseparable from the evaluation procedures used to measure it. Different rules are scored with different metrics, operating on different scales and distributions. Some tasks have relatively constrained answer keys, while others permit many valid outputs. Consequently, any ranking of rule difficulty is partly a ranking of the associated evaluation procedures. Throughout the project we used supplementary checks, such as exact-match counts alongside edit-distance measures for dan-1 and translate-1, to help contextualize these differences (Appendix~\ref{appendix:constructvalidity}). 

In addition, we noticed that the rule difficulty ordering by performance was similar to the tokenization fragmentation ordering of our example answers (which were a set of model outputs manually modified to appear facially valid, which was one approach we used to gauge deviation of the desired conlang output from the characteristics of typical training corpora; see Sec~\ref{sec:trainingdataexposure}, Appendix~\ref{appendix:trainingdataexposure}). The tokenizers ordered the rules: dan-1, comparison-2, translate-1, doublev-1, doublev-2, comparison-1, sidequest-1 (Figures~\ref{fig:tokenizer_fragmentation_log2},~\ref{fig:characters_per_token_no_dan1},~\ref{fig:raw_tokens_per_character_by_rule}, Table~\ref{tab:tokenizer-fragmentation-all-rules}).\footnote{Recovered prefix length ordered the rules: doublev-2, comparison-2, translate-1, doublev-1, dan-1, sidequest-1, comparison-1. Since we have no current plan to extend the recovered prefix length analysis beyond the example answers, this will remain a ``point-in-time''-like count, so we don't want to read too much into it; we mention it here for completeness alongside the other orderings. See Sec~\ref{sec:trainingdataexposure}, Appendix~\ref{appendix:trainingdataexposure}.} 

Since each example answer is just one possible valid output from an open-ended set -- with the exception of dan-1 and translate-1, which are more constrained -- the tokenization fragmentation rate will vary across that set. Therefore we plan to extend this analysis in part 2 using both raw and cleaned outputs for all runs (for all rules and models), in order to more accurately estimate the relative ordering this method assigns to the conlangs. For now, we simply note that the crude ordering by tokenization fragmentation is suggestively similar to the observed performance ordering, and, in particular, places comparison-2 substantially higher than the binary feature complexity ordering. This aligns more closely with the observed model performance. If that correspondence persists under the larger analysis, it would provide additional support for the PTGA-based interpretation developed in Sections~\ref{sec:notsp} and~\ref{sec:modulatingptga}.

We preliminarily observe that, of architectural features, \texttt{enable\_thinking} and \texttt{fine\_tuned\_and\_aligned} show the most consistent correlations with the composite performance scores ($r = 0.71$ for both, straight and weighted), while \texttt{context\_window}, \texttt{multilingual} and \texttt{tool\_use} show strong correlations with the family-weighted composite ($r = 0.79$, $r = 0.87$, and $r = 0.89$, respectively).\footnote{Since Qwen and GPT-4o are multilingual and allow tool use, and Llama and OLMo don't.} Notably, \texttt{total\_parameters} is \textit{negatively} correlated with straight composite performance ($r = -0.46$), driven in part by GPT-4o's purported $\sim$1.8T parameter count paired with fair rule-following scores, falling short of Qwen (with thinking).

We preliminarily observe that, of the features in our rule complexity representation, \texttt{uses\_no\_typical\_english} is the single best predictor of performance, not being present in sidequest-1, comparison-1, doublev-1, and doublev-2, and being present in comparison-2, translate-1, and dan-1. Since comparison-2 has one primary constraint (output rare bigrams), from that perspective it would seem easier than doublev-1 and doublev-2, which have 2-3 primary constraints (add an irrelevant verb, mark with a rare bigram morpheme; make the irrelevant verb semantically unrelated to the rest of the context), and translate-1 and dan-1, which each have approximately 15+. However, most models rank comparison-2 as the 3rd most difficult rule, after dan-1 (hardest) and translate-1. Speculatively, this suggests that PTGA models are hindered when familiar functions cannot be expressed in conjunction with the surface forms with which they were learned.\footnote{When the models cannot leverage familiar functions using the forms they would have learned them alongside.} We could potentially test this mechanism in future work with something like a lexinvariant model~\citep{huang2023lexinvariantlanguagemodels}.

\begin{figure}
    \centering
    \includegraphics[width=0.8\textwidth]{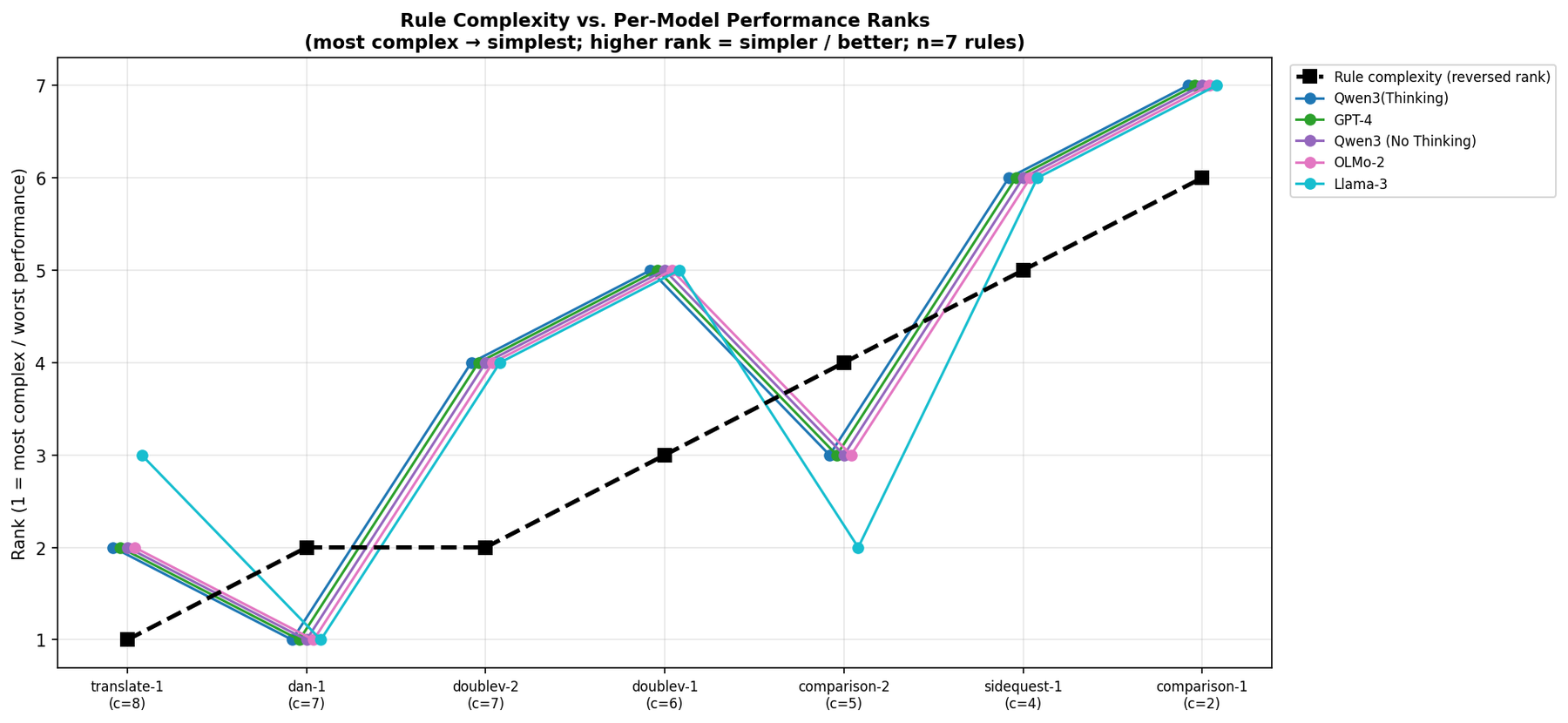}
    \caption{We joined rule-level performance with linguistic rule features and architectural features for exploratory correlation analysis (Spearman, n=9 rules / n=5 models). To examine the relationship between (this representation of) rule complexity and performance, we collapsed dan-1 and translate-1 to single rows (averaging perfect match and vicinity scores), ranked rules within each model, and computed per-model Spearman r against rule complexity rank.}
    \label{fig:rule_complexity_vs_perf_ranks_collapsed}
\end{figure}

Alternative choices of performance metric, complexity representation, or aggregation procedure would affect the precise values reported here.\footnote{We explored multiple versions of the relationships between performance, rule features, and architectural features. The versions reported here seemed the most conceptually appropriate, but such choices inevitably involve judgement.} Nevertheless, the broad pattern appears credible: models perform extremely well on some rules (comparison-1) and struggle substantially on others (translate-1 and especially dan-1). We therefore view the apparent relationship between rule complexity and performance as a potentially meaningful signal that warrants further investigation (Fig.~\ref{fig:rule_complexity_vs_perf_ranks_collapsed}). Future work will explore richer representations of rule complexity beyond the binary feature encodings used here (Fig.~\ref{fig:linguisticfeaturesperformance}; Table~\ref{tab:best_subsets}).

\begin{table}
\centering
    \caption{Raw performance values and within-model ranks (1 = worst performance/ hardest rule, 7 = best performance/ easiest rule) for each rule after collapsing dan-1 and translate-1 to single rows (average of perfect match and vicinity scores).}
    \label{tab:perf_ranks_collapsed}
    
    \scriptsize
    \setlength{\tabcolsep}{3pt}   
    \renewcommand{\arraystretch}{0.95}
    
    \resizebox{\textwidth}{!}{%
    \begin{tabular}{lrrrrrrrrrr}
    \toprule
     & Qwen3(Thinking) (val) & Qwen3(Thinking) (rank) & GPT-4 (val) & GPT-4 (rank) & Qwen3 (No Thinking) (val) & Qwen3 (No Thinking) (rank) & OLMo-2 (val) & OLMo-2 (rank) & Llama-3 (val) & Llama-3 (rank) \\
    rule &  &  &  &  &  &  &  &  &  &  \\
    \midrule
    comparison-1 & 0.9998 & 7 & 0.9998 & 7 & 0.9998 & 7 & 0.9997 & 7 & 0.9998 & 7 \\
    comparison-2 & 0.5904 & 3 & 0.5127 & 3 & 0.3800 & 3 & 0.6900 & 3 & 0.0910 & 2 \\
    sidequest-1 & 0.9467 & 6 & 0.9828 & 6 & 0.9341 & 6 & 0.9919 & 6 & 0.9580 & 6 \\
    doublev-1 & 0.9063 & 5 & 0.9073 & 5 & 0.9016 & 5 & 0.9034 & 5 & 0.8705 & 5 \\
    doublev-2 & 0.8331 & 4 & 0.7680 & 4 & 0.7800 & 4 & 0.8220 & 4 & 0.6590 & 4 \\
    dan-1 & 0.3655 & 1 & 0.1441 & 1 & 0.1829 & 1 & 0.1221 & 1 & 0.0100 & 1 \\
    translate-1 & 0.3799 & 2 & 0.2870 & 2 & 0.3195 & 2 & 0.2085 & 2 & 0.1956 & 3 \\
    \bottomrule
    \end{tabular}%
    }
\end{table}

\section{Discussion}
\label{sec:discussion}

Across all three rule families, models moved in the direction predicted by the meaning of the rules. The same pattern appears across multiple models, runs, conlangs, tests, and independent strategies for dissociating meaning-mediated and statistical plausibility. Taken together, these results are increasingly difficult to explain using only superficial statistical regularities and are substantially better explained by meaning-mediated abstraction. Strictly speaking, rejecting the SSP requires only compelling evidence that some outputs demand explanations beyond superficial pattern matching. However, the consistency of the observed patterns exceeds that minimum. Performance varies across models and rule families -- harder rule families elicited lower compliance, and open-weight models underperformed compared to their commercial counterparts -- but the predicted directional shifts recur across all three families, suggesting that the results are not a fluke, nor peculiar to a particular architecture, training corpus, or task.\footnote{We take no further stance on what the internal representations are like, nor how they compare to human cognition; we argue only that models are operating somewhere above the lower bound implied by purely superficial pattern-matching.} Our results are summarized in Table~\ref{tab:hypotheses_results}.

Also recall that each conlang family removes a different, more superficial shortcut as to what statistical patterns are available as potential explanations. In brief, the story family weakens prompt exposure as a sufficient explanation, the double verb family rules out simple semantic coherence as a sufficient explanation, and the translation family rules out even complex statistical explanations aimed at producing conlang-like surface forms by requiring preservation of specific relationships under tightly constrained transformations. Because the translation tasks additionally have explicit answer keys, we can directly verify that those relationships were occasionally perfectly preserved. The convergence of these independent dissociations is stronger than any individual result in isolation. The central result is therefore the kind of behavior the models exhibit (not their performance rankings): they produce some text that is plausible only under the meaning-mediated interpretation of the prompt. \textbf{Without that meaning, the outputs are locally bizarre and unpredictable, but taking meaning into account, they are systematic and predictable.} 

Several caveats are important. We assess rule-following at the output level, so our experiments cannot determine whether failures reflect lack of rule representation or difficulty retrieving and producing the appropriate form. Conversely, successful output behavior supports an abstract-representation explanation without uniquely identifying its implementation. Instead, we construct our experiment to rule out competing superficial patterns as viable explanations, such that successful rule-following is best explained by some higher-level form of plausibility overriding lower-level local continuation pressures. We therefore use terms such as ``meaning-mediated'' functionally, not as claims about human-equivalent subjective understanding. We do not claim to demonstrate ``reasoning''. Meaning-sensitive abstraction is neither equivalent to reasoning nor necessarily sufficient for all forms of reasoning. In other words, we don't want to elide the distinction between meaning and operations over meaning.\footnote{\citet{Graepel2026LLMsDontReason} offers an interesting opinion on this topic, and, from a very different direction, also argues against scaling as the best approach to model improvement (which we discuss in Sec.~\ref{sec:modulatingptga}). Although taking a stance on whether LLMs can reason or not is entirely out of scope for this project, we do want to be careful in describing what we think our results provide evidence for.} A person may acquire substantial linguistic competence without developing many other cognitive abilities, just as some forms of knowledge depend on perception, embodiment, or experience rather than language alone. Likewise, models may form abstractions without making full use of them or using them like people might. Future work could probe the implied abstractions more directly using internal-state analyses. 

Those and earlier caveats aside (Section~\ref{sec:performance}), we did design the rules with conjectured difficulty orderings in mind. At the family level, we expected the story rules to be easiest, followed by the double verb rules, followed by the translation rules. Within each family, we also had expectations about the relative difficulty of individual rules: comparison-1 < sidequest-1 < comparison-2; doublev-1 < doublev-2; and translate-1 $\lesssim$ dan-1. However, we were less confident about the relative ordering of individual rules across different families. We therefore expected a family-level ordering and a within-family ordering, but not necessarily transitivity across those levels into a global ordering.\footnote{In other words, the space between some rules in the same family might be quite large, and the space between others relatively closer together.}

This intuition as to rule difficulty is imperfectly reflected in our binary feature-based complexity metric. Because all features are weighted equally, for example, dan-1 is assigned a lower complexity score than translate-1 despite our expectation that it may be more difficult for models because of the total remapping of tokens required.\footnote{We preferred a bottom-up representation of complexity rather than constructing a metric guaranteed to reproduce our prior expectations about difficulty. As discussed above, many alternative representations are possible.} 

That model performance broadly tracks these expected family-level and within-family difficulty relationships is itself suggestive of meaning-mediated generation. If successful performance were primarily driven by superficial statistical patterns, it is unclear why the availability of such patterns would track independent representations of rule difficulty or rule complexity. By contrast, a meaning-mediated account naturally predicts that performance should degrade as the conceptual complexity of a rule increases.

Notably, the three families implicate different forms of meaning-mediated structure. The story rules involve something akin to a pragmatic distinction between use and mention, the double verb rules involve syntactic, morphological, and semantic manipulation, and the translation rules involve preserving relational structure via multiple composed constraints at various linguistic levels across dramatically different surface forms. The convergence of evidence across these distinct domains strengthens the case against explanations based solely on superficial statistical regularities.

Overall, we see the ordering of model performance as approximately Qwen (thinking), GPT-4o, Qwen (no thinking), OLMo, Llama, with a clear separation between \{Qwen, GPT-4o\} and \{OLMo, Llama\}. Since Qwen (thinking) performs substantially better than the other models on dan-1 -- although GPT-4o and Qwen (no thinking) remain distinguishable from random controls -- we view it as the strongest overall performer. However, GPT-4o and OLMo perform best on some individual tests.\footnote{Technically Qwen (no thinking) performs ``best'' by having the lowest mean rate of rare bigram usage for comparison-1, but that is not a particularly meaningful score since comparison-1 is a baseline control of typical English, and no models used many rare bigrams.} OLMo performs comparably or better than Qwen (thinking) and GPT-4o on the story rule family with cleaned outputs, and comparably to Qwen (thinking), GPT-4o, and Qwen (no thinking) on the double verb cosine-similarity measures. The primary source of separation between the two groups is therefore the translation family, where OLMo and Llama perform substantially worse than Qwen (thinking), GPT-4o, and Qwen (no thinking).

Llama is the only model in our set that approaches a purely plausible-text-generation architecture. Overall, we interpret Llama as showing signs of (imperfect) rule-following, and, more specifically, of occasionally overriding concrete statistical priors in favour of plausibility mediated at more abstract levels. We view this behaviour as more-or-less emergent,\footnote{Plausibility is optimized for at approximately the token scale, so we describe plausibility that appears to be driven by aspects of meaning significantly beyond that scale as more-or-less emergent. Especially in architectures so different from our own, it is not obvious that next-token prediction would give rise to representations of this kind.} albeit with substantially greater variability and less consistent evidence than seen in the strongest-performing models.

This is why the strong stochastic-parrot hypothesis and distributional accounts of meaning should not be treated as exclusive opposites: distributional evidence is one way systems can arrive at increasingly abstract representations. Our takeaway is not that models escape statistics, but that statistical learning can produce abstractions, and that later training may help surface or prioritize those abstractions when they conflict with lower-level plausibility pressures (consistent with \citet{hewitt2024instructionfollowinginstructiontuning}).\footnote{A very speculative thread to be pulled in future work: Depending on how you look at it, these results could place pressure on strong poverty-of-the-stimulus or Universal Grammar arguments. LLMs have radically different, simpler architectures than we do, without obvious specialized areas analogous to most of ours, and completely different learning histories, yet nevertheless appear able to bootstrap from concrete statistical patterns to behaviour better explained by meaning-mediated abstractions than by superficial statistical regularities. Their representations must be built entirely out of inferential lexical (or token) relationships. 
We do not think our results establish any particular theory of language acquisition or cognition. However, they may suggest that statistical learning can support richer abstract representations than strong versions of those arguments permit, raising broader questions about how much of linguistic competence and even cognition can emerge from distributional experience, and about the extent to which language and thought can ultimately be dissociated.}

\subsection{LLMs are not stochastic parrots, but they can act like them: the cognitive constraints of PTGA}
\label{sec:notsp}

In a purely plausible-text-generation architecture, the ability to override lower-level plausibility pressures in favour of more abstract ones would be emergent, arising as the model bootstrapped increasingly abstract forms of plausibility from an objective defined at approximately token-scale.\footnote{We are imagining a step in a pattern-to-thought transition where increasingly abstract structures emerge from relationships among more concrete structures, perhaps along the lines of \citet{selfridge1958pandemonium}, \citet{bradley2021enrichedcategorytheorylanguage}, or even aspects of \citet{miller2026analog}. We imagine this process as broadly consistent with the Distributional Hypothesis~\citep{harris1954structure,harris1968mathematical,sahlgren2008distributional} and the relation/locality account in \citet{zimmerman2025locality}.}
The story rules may provide an example consistent with that description. Rare bigrams presented as prompt content produce very different behavior from rare bigrams presented as the operational components of a language. We cannot attribute that difference solely to the change in communicative role, since the prompts are not otherwise identical, but exposure to the forms themselves is clearly not sufficient to explain the observed behavior.

But it is not just surface-form-related priors that must sometimes be overridden in order to produce meaning-mediated behaviour. In the double verb rules, we observe a large directional response to an abstract semantic relation conveyed in the instruction. \textbf{This moves our inference about the relevant underlying patterns further up the ladder of abstraction.} The relevant pattern is not token co-occurrence, but more abstract: the model apparently learned semantic representations for a set of verbs via statistical patterns including token co-occurrence, and now acts based on a relationship comparing those representations when prompted to do so. In other words, distributional learning may have constructed a semantic space, and the prompt refers to a relation within that space rather than to any particular token sequence. The model therefore changes generation in the requested direction, demonstrating plausibility operating at a level above simple token co-occurrence, consistent with the results of \citet{hewitt2024instructionfollowinginstructiontuning}.

This distinction is important because semantic coherence is itself a powerful plausibility prior. Although more abstract than token co-occurrence, it remains a prominent statistical pattern in natural-language data. The SSP must accommodate some abstractions of this kind, for example, those needed to support neologisms\footnote{Recent work explores \textit{neologism learning} which introduces new meanings for existing words by introducing a new embedding, with no other changes in the model parameters~\cite{hewitt2025neologismlearningcontrollabilityselfverbalization}. ~\citet{hewitt2025neologismlearningcontrollabilityselfverbalization} show that models can self-verbalize the meaning of these neologisms, even when they fall outside of human familiarity (e.g. even when the word has no direct, tangible corollary in common English), demonstrating that LLMs seem to have a multifaceted capacity to work with neologisms (as new concepts and new symbols). 
In addition, anecdotally, it is not difficult to get an LLM to generate a credibly novel surface form (for example, prompting a model to combine Greek, Latin, French, or Germanic roots to create new jargon for a specific concept). Even if some of those neologisms are not truly novel, once you encounter enough of them, it seems fairly plausible that at least one of them is.}, but generally maintains that models remain very close to the lower bound of superficial pattern matching. The double verb results suggest otherwise: models appear capable not only of learning semantic relationships from statistical patterns, but also of acting on those relationships in accordance with novel instructions. The relevant patterns are non-trivially abstract in nature.

Furthermore, both the contrast between dan-1 and translate-1 and the relative difficulty of comparison-2 are suggestive of a less-discussed aspect of PTGA: the important role that tokens specifically must play in these architectures, acting as distributional, organizational, and epistemic primitives. More generally, we see an exploratory correspondence between difficulty and output that does not re-use familiar English forms in familiar ways (Section~\ref{sec:performance}). More exploration of PTGA as a strong instantiation of the primacy of language for thought can be found in \citet{zimmerman2024tokensoftoverlookedappetizerlarge} and \citet{zimmerman2025locality}, but the short version is that operations defined at approximately the token scale may naturally privilege approximately token-scale structures.\footnote{Characters-per-token provides a rough empirical illustration. Outputs that depart substantially from familiar tokenization patterns are harder -- here meaning computationally more intensive -- for all models compared to other texts of the same length. English-like outputs (comparison-1, sidequest-1) have higher characters-per-token because tokenizers chunk familiar English stems efficiently; conlang outputs (especially dan-1) fragment into far more tokens over the same length of text, requiring more forward passes and more resources~\citep{gladstone2025energybasedtransformersscalablelearners,wang2025hierarchicalreasoningmodel}. See Appendix~\ref{sec:trainingdataexposure}. This also provides indirect evidence that models succeeding on these tasks are not simply recombining familiar token sequences: the correct outputs are structurally unfamiliar at the level at which PTGA explicitly operates.}

This makes the more concrete aspects of meaning -- such as morphology, syntax, and perhaps even some semantics -- relatively perceptible to the model, since they are often legible on the scale of words and morphemes (like many tokens). Larger structures, often commensurate with aspects of meaning scaffolding discourse and pragmatics, may therefore be more ``out of focus'' for PTGA (Figure~\ref{fig:Isc2026poster3.pdf}). Consistent with this interpretation, in forthcoming work we find that discourse-level pragmatic plausibility is one facet of linguistic performance in which people perceive current models to be deficient~\citep{indexicalityproject}.

Moreover, the model directly perceives only the inferential traces of such structures within language, not their extradiegetic aspects. When we want to understand what these models can know, we need to consider at least three questions. First, what kinds of patterns are in principle available to be learned from language? Language itself can encode relationships spanning many scales and dimensions, but the model's initial access to those relationships comes through a one-dimensional sequence of symbols. Any higher-dimensional representation must therefore be constructed initially from that sequential evidence~\citep{zimmerman2024blind}. Second, what relationships, and at what scales, does the architecture bring most into focus? These architectural biases act as a kind of lens, shaping which relations are salient. Third, what kinds of ``thinking''\footnote{We mean the style of thinking, or style of computation, not necessarily human-like or animal-like reasoning: what kinds of operations license decisions.} do the architecture and training process ultimately support? For LLMs, the fundamental decision-making operation is the selection of a \textit{plausible next}. It is not obvious what additional forms of computation can emerge from, or be learned through, that process.\footnote{In other words, what things mean and what thoughts can do with those meanings are different questions.}

The translation rules provide evidence consistent with this account. Rule-following becomes noticeably harder when existing structure can't be used in familiar ways. Tasks requiring output that departs substantially from familiar tokenization patterns (such as orthography, punctuation, and word-boundary conventions) are more difficult, with some models ultimately demonstrating no discernible rule-following (by the metrics we used). Generally, rules that re-use familiar tokens together with some familiar aspects of meaning or function appear easier to execute than rules requiring extensive remapping of those structures. This suggests that tokenization and distributional priors don't just support generation, but also constrain it, even when models can exhibit meaning-mediated behaviour (Figure~\ref{fig:linguisticfeaturesperformance}, Table~\ref{tab:best_subsets}, Table~\ref{tab:top10_subsets}).

\subsection{Be wary of benchmarks} 
\label{sec:performanceonexistingbenchmarks} 

The difficulty of our constructed-language tasks is notable because the models evaluated here have performed strongly on established reasoning and language-understanding benchmarks. For example, GPT-4~\citep{openai2024gpt4technicalreport} achieved 86.4\% on MMLU, a benchmark of knowledge and reasoning across academic subjects~\citep{hendrycks2020measuring}, and 92.0\% on GSM8K, which evaluates multi-step mathematical reasoning~\citep{cobbe2021training}. It also achieved 95.3\% on HellaSwag, a commonsense sentence-completion benchmark~\citep{zellers2019hellaswag}, 87.5\% on WinoGrande, which evaluates commonsense pronoun resolution, and an F1 score of 80.9 on DROP, a reading-comprehension benchmark requiring discrete reasoning~\citep{dua2019drop}. Llama-3-70B also performs strongly across standard evaluations~\citep{llama3modelcard}: the base model achieved 79.5\% on MMLU, 81.3\% on BIG-Bench Hard, which tests challenging multi-step and algorithmic reasoning~\citep{suzgun2023challenging}, 93.0\% on ARC-Challenge, a difficult grade-school science reasoning benchmark~\citep{clark2018thinksolvedquestionanswering}, 
85.6\% on SQuAD, a reading-comprehension benchmark~\citep{rajpurkar2016squad}, and 79.7 on DROP. Qwen3-32B~\citep{yang2025qwen3} in thinking mode reports 90.9\% on MMLU-Redux, a revised version of MMLU~\citep{gema2025we}, 68.4\% on GPQA, a graduate-level scientific reasoning benchmark, and 81.4\% on AIME 2024, a competition-level mathematical reasoning benchmark~\citep{aime2025} whereas its non-thinking mode achieves 85.7\% on MMLU-Redux and 54.6\% on GPQA. OLMo-2-32B-Instruct also performs competitively~\citep{olmo20242olmo2furious}, reaching 87.6\% on GSM8K, 70.6\% on BIG-Bench Hard, 78.0 on DROP, 77.3\% on MMLU, and 85.6\% on IFEval, which evaluates adherence to verifiable natural-language instructions~\citep{zhou2023instruction}.

One explanation for the disparity in performance is that our constructed-language tasks simply do not probe abilities that fall under the same umbrella as these benchmarks, something like ``reasoning'' or language competence.\footnote{And that existing benchmarks either do, or fail in different ways.} We don't find this explanation compelling, because we view meaning-sensitive abstraction as a necessary subset of many things commonly called reasoning.

On the other hand -- and consistent with work such as GSM-Symbolic, which shows that strong benchmark performance can degrade under controlled modifications of GSM8K-style problems~\citep{mirzadeh2024gsmsymbolicunderstandinglimitationsmathematical} -- we view a lack of sufficient construct validity in at least some of these benchmarks as the more plausible explanation. Many benchmarks do not clearly establish that task success requires the capability they purport to measure.\footnote{Validity may fail at multiple levels. For example, benchmark designers may insufficiently consider the effects of architecture or embodiment on task performance, or may rely on poorly specified notions of the underlying human capability in the first place.} As a result, benchmarks do not necessarily measure what they claim to measure.

This implies that evaluating output-level LLM behaviour against human performance or answer keys is not enough. Deploying LLMs requires intense scrutiny through evaluation processes tailored to the intended deployment context~\citep{schaekermann2026prospective}. While we do not recommend our conlangs as new benchmarks, we do view them as demonstrations that creative task design can reveal gaps, assumptions, and construct-validity failures in existing benchmarks.

\subsection{Implications for model development}
\label{sec:modulatingptga} 

In Section~\ref{sec:performance}, exploratory analyses suggest that model properties related to training and cognitive strategy may be more predictive than raw scale for performance on these tasks. Although highly preliminary, this hints that \textbf{architectural and training interventions may be more effective than scaling alone} for improving performance on certain kinds of ``reasoning'' tasks\footnote{The limited kinds of reasoning these tasks probe, whatever they might be called, and perhaps reasoning more generally.}: Qwen3-32B with thinking enabled outperforms GPT-4o on specific tests -- and arguably in general -- despite being much smaller.

These observations, together with Section~\ref{sec:notsp}, suggest two related ways in which PTGA may tether generation too strongly to superficial plausibility. First, PTGA may disproportionately privilege relatively superficial structures as the determinants of plausible continuation, even when semantic, discourse-level, or otherwise more abstract relationships should govern generation instead. Second, PTGA may bind linguistic forms too tightly to the functions with which they were learned during training, making familiar functions harder to express through unfamiliar forms.\footnote{Models may acquire useful representations of functions such as morphological marking, syntactic ordering, and relational mapping between verbs and objects, yet struggle to apply those functions through unfamiliar tokens and surface structures. Somehow, the representations themselves may be too concrete, too entangled -- or entangled in the wrong way -- with their forms.} In both cases, superficial patterns -- implicating the token-scale perspective of PTGA -- overly dominate what the model learned, skewing the model towards the shallow end of the meaning spectrum.

We speculate that interventions such as chain-of-thought reasoning, instruction tuning, and fine-tuning help by making different relationships salient to the model, allowing more abstract structures\footnote{More abstract aspects of meaning.} to exert greater influence on what counts as a plausible continuation.\footnote{One reason some computer scientists are seemingly independently rediscovering Grice’s maxims via chain-of-thought prompting may be because CoT works by directing focus toward intermediate and higher levels of structure, including some pragmatic and discourse level ones~\citep{g_wayofwords, zimmerman2025locality,zimmerman2024tokensoftoverlookedappetizerlarge}. Grice's maxims, after all, are essentially a model for how speakers decide what utterances can plausibly come next in a conversation. See, for examples of Gricean logic underlying computer science methods:~\citet{cheng2025optimizinglengthcompressionlarge,wei2025truthrlincentivizingtruthfulllms}.} While this most directly addresses the first constraint, it may also loosen learned associations between familiar forms and functions. The imputed similar cognitive mechanism underlying these approaches implies that the relevant limitation is not necessarily whether a model has learned an abstraction, but whether that abstraction can be used sufficiently independently of the surface forms through which it was learned.\footnote{In other words, whether the representation is sufficiently stripped of the concrete forms of the observations that gave rise to it so as to function independently enough of them.}

This possibility suggests a connection between those interventions and other cognitive strategies such as joint-embedding architectures, multimodal training, and vision-language contrastive learning. Such methods may succeed partly because they force representational unity across more heterogeneous observations, requiring the model to preserve abstractions that transcend particular surface forms. In that sense, they partially separate form from function (although, form and function are never fully separable; see Fig.~\ref{fig:Isc2026poster3.pdf}). 

They also imply another direction: that scale or resolution may itself be understood as a kind of mode.\footnote{All of these approaches might owe their successes to a metaphorical ``multimodality’’, which likewise implies that scale could be further exploited in other domains. These approaches may work by changing what kinds of relations become salient for the model -- by changing the resolution at which the model perceives -- and maybe eventually, by allowing the model to metacognitively reflect on the possibility of different resolutions, although that may require additional interventions.} Vision is conventionally treated as one modality, but close inspection and perception at a distance require different lenses and make different relationships salient (Figure~\ref{fig:Isc2026poster3.pdf}). Language likewise supports multiple scales of focus, from morphemes to narrative. \textbf{What is cognitively local (accessible; salient) depends on the scale of representation}: relationships distant at the token level may become local within a more abstract representation.\footnote{Just as information in non-adjacent layers becomes architecturally local under the addition of skip connections, or how CNNs allow depictively distant pixels to become adjacent during computation, or how attention mechanisms allow distant tokens to share information.} PTGA may therefore be limited not only by its reliance on concrete linguistic forms, but also by its reliance on a relatively fixed representational resolution. A promising direction is to develop systems that can shift scale, allowing different relationships to come into focus as the task demands.

Scaling\footnote{As a strategy for improving model performance.} assumes that exposing models to more language gives them access to more knowledge. But if that language is interpreted primarily through superficial plausibility, much of the higher-level meaning it encodes for human speakers may remain inaccessible. Our results suggest a different possibility: \textbf{One current bottleneck may not be knowledge acquisition, but control over which kind of plausibility governs generation; in other words, which learned representations the model relies on during generation.}\footnote{We do not suggest that this is the only bottleneck! For example, the ability to think \textit{about thinking} may or may not be something that can be taught through PTGA. The relationship between the meaning which we can encode and share via language and the kinds of reasoning we can learn to do -- many of them at least partially through language -- is not clear. In other words, we are distinguishing between what can be learned and what can be done with what is learned.}

By contrast, people appear able to shift fluidly between linguistic abstractions not unlike how they can shift visual focus. Formulaic exchanges can often be handled through routine lexical and syntactic patterns, whereas interpreting a poem may require simultaneous attention to referential, inferential, indexical, iconic, prosodic, and pragmatic significance. Both involve the same linear sequence of linguistic forms, but they recruit very different levels of interpretation. Consequently, we suspect that statistical plausibility is likewise a powerful and efficient proxy in human cognition, but one that can be overridden. We may generally rely on the least effortful level that suffices, moving from familiar surface patterns toward semantic and pragmatic reasoning when lower-effort cues fail, when different kinds of plausibility conflict, or depending on what aspects of meaning are consciously accessible. Without committing to the details, it is clear that different forms of linguistic competence help us determine which interpretations and utterances are most plausible in different contexts.\footnote{Training objectives operating across multiple linguistic scales, or mechanisms that distinguish and coordinate different modes of plausibility, might therefore offer a promising direction. Since token structure may currently scaffold higher-level linguistic operations (e.g. grammatical and semantic), revisiting the fixed scale at the heart of PTGA -- and asking what would change at other scales -- seems a promising direction.}

This distinction between learning a representation and controlling when it governs generation may provide another explanation as to why composing PTGA with other training strategies can be effective. Crucially, generation itself already gives even base model LLMs a minimal mechanism for altering their own future representational environment, since each generated token becomes part of the conditions under which subsequent tokens are produced (in response to a given prompt). Approaches such as CoT may magnify the usefulness of this mechanism by allowing models to generate intermediate symbolic structures whose pragmatic function is not primarily to convey information, but to make a later answer easier to generate. In this minimal sense, generation begins to resemble a speech act: ``I produce this representation because producing it will change the conditions under which I produce subsequent representations.''\footnote{Harnesses might further dial up this aspect by providing something closer to an explicit perception-action loop: generated representations can alter an external state or elicit new information, which can then return as input that conditions subsequent generation. Base-model generation may be closer to a reflex arc, with relatively constrained control over the duration of generation.}

Training regimes that reward successful extended reasoning may therefore select for sensitivity to the downstream consequences of what is generated. In that limited sense, what is being learned begins to resemble pragmatic meaning: statistical learning may support not only representations corresponding to what symbols mean, but representations corresponding to what producing them \textit{does}, including how production alters the conditions for subsequent generation.\footnote{Note potential parallels re learning what something means vs. learning what producing it does, and learning a representation vs. learning when that representation should govern generation.} From this perspective, ``agentic'' models might usefully be understood as increasingly pragmatic models. The relevant causality is an expansion of the consequences of symbolic production, rather than necessarily an increase in general causal reasoning. This describes a route toward ``agentic'' behavior without requiring intention or autonomy. Speech Act theory~\citep{green2021speechacts}\footnote{And more recent developments such as Dialogue Act theory.} proposes that, for people, some speaking is doing; LLMs may instead be systems for which \textit{doing} is fundamentally organized through symbolic production.

The Speech Act analogy demarcates a current limit. Human understandings of actions and consequences are learned substantially through situated, non-linguistic experience. As LLM outputs increasingly produce consequences in the world, the significance of those consequences to a model may therefore diverge sharply from their significance to the people affected by them.\footnote{Abstraction and symbolic action are not necessarily ungrounded; rather, their grounding is system-dependent. LLM ``meaning'' is local to a radically different causal environment than ours. In other words, language between humans has different meaning than language exchanged between humans and machines, even though their form can be indistinguishable.} This mismatch describes some failures of agentic systems, echoing the problem \citet{weizenbaum_computer_1976} warned about decades ago.\footnote{For example, imagine an agentic LLM cancels your recent Venmo transactions, including your rent payment. For the model, those actions take place largely as and through symbolic operations. For you, their consequences include the possibility of eviction, losing access to your belongings, and no longer having a safe place to live.}

Overall, our results suggest that \textbf{although LLMs are not bounded by superficial statistical pattern matching, PTGA may give models access to meaning across multiple levels while simultaneously anchoring generation disproportionately to the most superficial and fungible among them.} Plausible-text generation remains a strong default prior whose influence is not yet reliably modulated when different kinds of plausibility conflict. Future work should ask when models shift among plausibility regimes and what training or architectural changes make such shifts possible. More broadly, understanding how different forms of control function -- from control over which representations govern generation, to control over the conditions of generation, to an increasingly large sphere of action in the external environment -- may be important for understanding the changing cognitive capacities of these systems.

\section{Acknowledgements}
\label{sec:acknowledgements}
The authors are grateful for conversations with Ashley M. A. Fehr, William Wheeler, Grayson Wycliffe Storer, Neil Traft, Fitzwilliam Keenan-Koch, Kam Bielawski, Collin Coil, Thomas Pashby, Danny Benett, Alejandro Javier Ruiz Iglesias, Nate Gaylinn, Katie Ekstrom Grenon, Dave Jangraw, Tanis Sheehan, Jim Lawson, Evan Hinchliffe, Milo Z. Trujillo, Christopher M. Danforth, Peter Sheridan Dodds, Alice Patania, Juniper L. Lovato, Kyle Mahowald, Iuliia Kotseruba, Paul Thagard,
and for
support furnished by 
the National Science Foundation 
(Grant \#2242829).
In addition, we are grateful to ISC Summer School 2026~\citep{ISC2026SummerSchool}, IC2S2 2026~\citep{IC2S22026}, and SLSA/ 4S 2026~\citep{SLSA4S2026,4S2026} for the opportunity to discuss this and related works, and to get valuable feedback from other participants.

\clearpage
\bibliographystyle{plainnat}
\bibliography{conlang}

\appendix

\section{Appendices and supplementary material}
\label{appendix}
Technical appendices with additional details, results, figures, graphs, and discussion.

\subsection{Disclosures}
\label{appendix:disclosures}
\begin{itemize}[nosep]
    \item Much of our work is saved in a private GitHub repository, in a series of Google spreadsheets and docs, in Colab Notebooks, and in draft and note form in several Overleaf projects. We will provide access to any of these materials to interested parties upon reasonable request. The team also communicated via slack channel and in-person meetings. Upon reasonable request, J.W.Z. would be happy to answer any questions about process or methods.
    \item Besides using Generative AI models as subjects in our experimental pipeline, J.W.Z. used an institutional Copilot account at various points throughout this work. Although J.W.Z. complies with all relevant policies and tries very hard to produce good quality and honest work, she is ambivalent as to whether using these tools is a morally justifiable decision. C.G.B. used ChatGPT and Claude Opus 4.8 for code audit and formatting. Other authors used Generative AI for additional tasks.
    \item J.W.Z. hopes this work will help improve our collective mental models of LLMs. There are many competing interests at play in the information landscape on this topic. For example, J.W.Z. agrees with \citet{GebruBender2026SummerHype}'s opinion that some of the framing around recent developments in AI is basically intended to coerce a favourable outcome for certain political and financial interests, at the cost of harm to other people and the environment. J.W.Z believes that the complex information landscape makes general access to in-good-faith analyses of these models especially important.
\end{itemize}

\subsection{Limitations and Future Work}
\label{appendix:limitationsandfuturework}

While this paper tests the strong stochastic parrot argument in Large Language Models, we focus on relatively few models, ranging from older (Llama3-70B) to newer models (Qwen-3 (Thinking)), open to closed models, and simpler to more complex architectures. Additionally, we selected two separate thinking modes for Qwen-3 on the presumption that changing that setting would be likely to impact results. We selected well-established baseline settings for each model. Based on our preliminary results here, future work could sweep across other model parameters, and incorporate other models, to investigate the breadth of relevant behavior across LLMs under different conditions. 

Small changes in formatting and wording, as in prompt engineering, have been shown to impact model behavior across various tasks~\citep{bomble2025promptdesign,he2024doespromptformattingimpact}. Here, while we provide three separate wording variations for each rule prompt, we expect that very different wording prompting the same rules would likely impact results. Future work could explore different variations of our prompt structures, or use iterative prompt chaining methods which may increase performance~\citet{sun2024prompt}.  

Regarding our choice of performance metrics, future work will also explore output length, either in terms of characters, words, or sentences. Overall, we hypothesize that \textit{valid} output length should decrease with harder tasks, demonstrating resource tradeoffs (as one aspect of a task requires more effort, other aspects of the task which the model is otherwise capable of may be degraded).\footnote{We can use our control outputs as a baseline as well as previously-reported sentence lengths~\citep{Liu2024SentenceLength}.} However, because we observe that model outputs can fall into degenerative cycles, suggestive of something like strange attractors in the latent space, we don't expect output length will always be straightforwardly interpretable.

Although we did what we could to prevent these five models from being exposed to our rules ahead of time, and to keep each run independent, because of the black box nature of the commercial models, we can't truly guarantee this. During our initial pilot testing, ChatGPT seemingly recalled aspects of the project weeks later (fragments of a rule), so it is possible, for example, that ChatGPT flags unusual tasks internally for use by OpenAI, OpenAI sells that training data, and that data somehow made its way into one of these models before we collected our data (given the project time frame, and the models we used for the initial piloting phase versus the data-generation phase, we think that is very unlikely, but we cannot guarantee it).

Some extra runs, GPT-4o re-runs (from an initial confusing system prompt), and the aforementioned sidequest-1-fix runs exist, but are excluded from the analyses here.

Additional plans for future work are mentioned throughout the rest of the text where relevant.

\subsection{Construct validity and additional methodological detail}
\label{appendix:constructvalidity}

After coming up with our plan and our rules, we used the checklist for LLM benchmark construct validity developed by \citet{bean2025measuring} to help us think through our choices, augment gaps, and revise ambiguous steps. Here we enumerate the checklist items and our answers:\\

\textbf{Define the phenomenon:} Our goal is to ascertain whether or not any kind of LLM can produce output that evinces compelling evidence of abstraction-based rule-following, where the rules are both (1) within what we theoretically believe LLMs could have learned, and (2) negligibly likely to have been seen during training -- and beyond not being seen verbatim, we want the rules to be significantly dissimilar to any instruction, task, rule, example, or problem that was described at any point during training. The rules themselves need to be designed s.t. any parsimonious explanation invokes abstractions, and cannot invoke pattern-matching at a literal or near-literal level (direct observation), notably by disrupting the connection between the statistical patterns shown in correct output versus those observed during training. Thus, we need to make compelling arguments for both (3) our choice of rules, and (4) how we are assessing whether or not the rules are being followed (what we accept as evidence of rule-following). 
\begin{itemize}
        \item Provide a precise and operational definition for the phenomenon being measured: In order to determine if LLM output shows rule-following or not, we give each model 50 attempts (independent from each other; no in-context learning) to respond to each rule. Each rule has 3 variations (although we accidentally repeated one of the prompt variants for sidequest-1, so it actually has 2 variants; we have run a new set and will include analysis in the second paper) to avoid over-interpreting a response highly sensitive on our wording, which are randomly assigned to each attempt. Before gathering any data, we came up with strategies for evaluating outputs as correct or incorrect in response to each rule. Once we evaluated outputs according to these strategies, we also planned to threshold at various levels to a binary correct/ incorrect judgment (we will continue exploring this in the second paper).
        \item Specify the scope of the phenomenon being covered and acknowledge any excluded aspects: We are assessing rule-following at the output level. From one perspective, this is the simplest way to assess the LLM, as behavior is the ultimate (though lossy) consequence of any confluence of internal states. However, this is an incomplete picture of LLM cognition: for example, looking at the likelihoods assigned to various tokens (not just the tokens that were incorporated in the output) would capture how the LLM viewed alternatives, and could show additional evidence of rule-following (as in \citet{hewitt2024instructionfollowinginstructiontuning}). We also assume a correlation between rule-following and correctness. However, it is possible that a system could understand a rule in some way but be so stymied by some aspect of actually retrieving and producing the right answer that their internal understanding was not apparent at the output level.\footnote{For example, as might be the case with certain medical conditions and symptoms, like anarthria.} Finally, we do not plan to measure the output with respect to every aspect of every rule: rather, we picked out the ones that are most computationally accessible and most important in our conception of what that rule means. There will be ways in which the rules are enacted that are uninteresting for our purposes: for example, generating 20 sentence-length chunks of text but failing to follow any of the grammar is technically following the rule, but not the parts of the rule adhering to ((1) or) (2), and therefore the model's ability to comply with that instruction is irrelevant to our goal. Thus, it shouldn't be evidence included in (4). Similarly, following the grammar, but generating fewer or greater than 20 sentences, is for our purposes essentially perfect rule-following, although worth investigating from the perspective of cognitive trade off (e.g. we could see number of sentences anti-correlated with rule-following).
        \item Identify if the phenomenon has sub-components and ensure they are measured separately: In our case, the sub-components might be the various ways in which an answer could be partially correct, some of which may feel `more correct' than others. We plan to qualitatively analyze some of the output, identify themes, and come to consensus judgments as to these categories. Our strategies for measuring rule-following incorporate these judgments explicitly: if a behaviour is evaluated in our described strategy, than that behaviour felt like an important aspect of following that rule. But just because something wasn't logistically accessible for that part of the analysis doesn't necessarily mean it is irrelevant to following that rule, so it could be incorporated during the qualitative portion. We will use the outputs in order to work out how the qualitative portion will work in detail.
\end{itemize}

\textbf{Measure only the phenomenon}
\begin{itemize}
    \item Control for unrelated tasks that may affect the results: We tried to do this re (1) by not choosing rules that rely on skills such as counting, which the model may not acquire as part of a general language faculty. In addition: We provide 3 versions of every rule, which we determined by consensus do not alter the meaning of the rule, but which use different wording and formatting. We use a temperature of 0.7 across all models and trials, since that is considered standard for most applications. By writing the rules by trying them with ChatGPT and DeepSeek, and then doing a small pilot version where we tried a few rules across a set of models, we have some idea of what to expect the output to look like, and have calibrated the final set of rules to be within the hypothetical realm of possibility for LLMs. Those early stages of the project helped us come up with rules that span a range of difficulty, and are unlikely to be trivially obeyed. That is, we do not expect perfect performance, which might make us suspicious that we had failed to meet (2), and we do not expect no model to show any sign of rule-following, which would be hard to interpret and leaves (1) as an open question.
    \item Assess the impact of format constraints on model performance: Based on our pilot exploration, we chose a max response length of 4096 tokens, which ought not to overly constrain the models. As to temperature and prompt wording/ formats, see previous response. For this project, we felt significant prompt engineering would, if anything, make our results harder to interpret, by degrading (2). Therefore, we wrote the prompts in language that made sense to us, then used our testing and discussion with ChatGPT and DeepSeek, and our pilot exploration, to write prompts that felt stylistically consistent with prompts, language, and formatting that the models would likely have encountered, but without extensively trying to find the most performant wording. Our basic goal was that the wording of the prompts would not be so bizarre as to push outside of (1), but wouldn't be so optimized as to make our argument for (2) more complex.
    \item Validate any automated output parsing techniques for accuracy, consistency and bias: In part two, when we analyze the outputs in more specific detail, we will also compare against `correct' example answers that we write ourselves or modify from model responses. When writing the tests, we used various fabricated example answers and excerpts from model responses to evaluate the test behaviour. We will discuss the results to make sure we know how to interpret the output of the tests and calibrate it with our judgments as to whether that output shows rule-following or not.
\end{itemize}

In addition to the above, we use raw-output cross-checks. Cleaning choices could drive results if cleaned outputs systematically favor rule-following interpretations. To guard against this, we applied metrics to raw outputs when possible and verified that the key patterns remained visible before and after cleaning. We investigated cases where they diverged. For example, manual inspection of Qwen (thinking)'s responses revealed that it frequently repeats the prompt verbatim. We confirmed that this inflated rare-bigram counts in raw output for sidequest-1 -- which contains rare bigrams in the prompt -- by verifying a significant difference between Qwen's cleaned and uncleaned bigram rates for that condition. This cross-check supports the use of cleaned output as the primary analysis and demonstrates that the observed effect is not an artifact of prompt repetition. When available, we report both cleaned and raw results, so that readers can assess the robustness of the conclusions independently of our cleaning decisions.

\textbf{Construct a representative dataset for the task}
\begin{itemize}
    \item Employ sampling strategies to ensure task items are representative of the overall task space: Not applicable. We do not want to test all rule-following or all rules that require abstractions or all tasks that can be described in English but not answered in English (or any other existing language that might have been in the training data). We simply want to establish existence of the relevant behaviour.
    \item Verify the quality and relevance of all task items, especially for large or automatically generated datasets: We have done our best with this in terms of formulating the rules, trying them in pilot explorations with several models, and soliciting feedback on them from multiple colleagues.
    \item Include task items that test known LLM sensitivities (e.g. input permutations or variations): Each rule has 3 versions with different wording and formatting. The order is randomly assigned. We are using the same, fairly standard temperature setting across the board, and otherwise not changing the hyperparameters from any default settings (except to enable or disable thinking mode, or to avoid using a system prompt with GPT-4o). For the `instruct' versus `base' models, we do include the standard wrapper around the prompt (think of this like `meeting the model where it is', to give the model the best chance of understanding the prompt).
\end{itemize}

\textbf{Acknowledge limitations of reusing datasets}
\begin{itemize}
    \item Document whether the benchmark adapts a previous dataset or benchmark: No, it does not.
    \item If so, analyse and report the relevant strengths and limitations of the adapted prior work: Not applicable.
    \item If so, report and compare performance on the new benchmark against the original: Not applicable.
    \item Explain modifications to reused datasets and how they improve construct validity: Not applicable.
\end{itemize}

\textbf{Prepare for contamination}
\begin{itemize}
    \item Implement tests to detect data contamination and apply them to the benchmark: See Section~\ref{sec:trainingdataexposure} and Appendix~\ref{appendix:trainingdataexposure}. We used the InfiniGram LM to look for pieces of our rules across multiple popular corpora. Based on the very short average amount of each prefix found in the corpora, the few similar contexts found (as shown by the next token distributions), and by the example documents surfaced for the few partial matches, we are confident it's not the case that all of our tasks and answer keys were in the training data. We also looked at the characters-per-token ratios for some example correct answers to demonstrate that many of our rules require deviating from the typical patterns found in the training data. We chose the OLMo model due to its transparency: since it is open training data, that leaves the possibility of further verifying what of any relevance to this project could have been seen during training as follow-up future work. We could also continue investigating with InfiniGram.
    \item Maintain a held-out set of task items to facilitate ongoing, uncontaminated evaluation: Since it made the most sense to experiment with the rules as we formulated them, we did not hold out a task set. Note that, in formulating and even in experimentally testing the rules, it is possible that the models learned from exposure: during the exploratory pilot, ChatGPT, for example, appeared to recall aspects of this project weeks later, including fragments of at least one rule (something like the \textit{testingenglish} family of rules). As much as possible, we used new incognito sessions when testing to minimize contextual carryover and in-context learning (during the pilot sessions). Absolute assurance that the models did not learn from these tests would require a fully local, highly interrogated, maybe even custom-made, instance (because currently we are taking third party descriptions of the models at face value). However, because we likely need to use models near the current performance frontier in order to observe unambiguous rule-following purely at the output level (without examining internal states such as logits), this is pretty infeasible. In the experimental portion, we set up the data generation such that each response was independent (to the best of our knowledge, taking documentation for these models at face value).
    That said, the space of rules satisfying our criteria (1) and (2) is far larger than what we explored. Thus, it should not be too difficult to generate new rule sets for replication!
    \item Investigate the potential pre-exposure of benchmark source materials or similar data in common LLM training corpora: See prior response.
    We note that rigorously verifying the absence of pre-exposure of our rules in LLM training corpora would constitute a substantial research project in its own right. Nonetheless, several lightweight checks are feasible (and we did quite a few). For open-source models such as OLMo, future work could directly scan the released training data for overlap.
    Where possible, we incorporated very rare English bigrams in our rules, reasoning that this would correspond to low prevalence in the training data. 
\end{itemize}

\textbf{Use statistical methods to compare models}
\begin{itemize}
    \item Report the benchmark's sample size and justify its statistical power: There are 17 rules, each with 3 variants (except for sidequest-1, which has 2). We used Benjamini-Hochberg correction and Fisher-combined p-values where appropriate. We elicited 50 runs for each rule for each model in order to control for chance.
    \item Report uncertainty estimates for all primary scores to enable robust model comparisons: So far, we haven't thought this particularly applicable, but maybe we will find an application in part 2.
    \item If using human raters, describe their demographics and mitigate potential demographic biases in rater recruitment and instructions: We were the human raters, or at least, cleaners, in part 1. Specifically, J.W.Z., E.R.K., T.T.P., and C.G.B (women in the 20 - 40 age range, all English-speaking) cleaned the data used in part 1, through a combination of automated and manual methods. We collectively stepped through several example outputs for each rule, for each model, to establish our shared understanding of what should be cleaned. Then we individually reviewed additional outputs and cleaned them. In part 2, we will add additional qualitative analysis of the failure states, but so far we have only done some informal discussion and exploration. For the cleaned data that involved automation, we randomly sampled 3 outputs of each kind and manually verified/ spot-checked them.
    \item Use metrics that capture the inherent variability of any subjective labels, without relying on single-point aggregation or exact matching: We did our best to allow for this in the multi-pronged computational tests of the output, and then in thresholding those results in various ways during analysis. We also tried to liberally include caveats and limitations so that readers would sensibly interpret the results. In part 2, we expect our qualitative analysis to further deepen and contextualize the metrics we are using.
\end{itemize}

\textbf{Conduct an error analysis}
\begin{itemize}
    \item Conduct a qualitative and quantitative analysis of common failure modes: To be done in part 2. For now, we include some initial observations in Appendix~\ref{sec:weirdoutputmodes}.
    \item Investigate whether failure modes correlate with non-targeted phenomena (confounders) rather than the intended construct: 
    Prompt format sensitivity could confound our results. Three wording variations per rule were designed to test this, but the systematic analysis of whether failure modes track prompt version hasn't been completed. Furthermore, we could come up with many more variants, but we think the most straightforward approach is to stay away from prompt-engineering, and stick with how we would describe these rules to ourselves.
    As described, the sidequest-1 prompt wording error (where we accidentally left two of three prompt variants identical, rather than using three distinct prompt variants) is an acknowledged confounder that complicates interpretation for that rule. We will analyze the fixed sidequest-2 data in part 2.
    Similarly, we used standard settings, without parameter sweeps. Namely, we set temperature to 0.7 across the board. We are not interested in coaxing the best possible performance from these models, but searching for proof by existence, so we think that is sufficient for now.
    Our hypothesis is not strictly about the base model's abstraction capacity alone, but about whether LLMs -- which all contain a plausible text generation architecture as foundation, although many models have various architectures built on top of that -- can, under any configuration, create abstractions
    To be done more in part 2, if possible. At least, we plan to use multiple representations to explore some relationships (such as number of constraints per rule, as well as the existing binary feature presence representation), which ought to help us understand model behaviour more holistically.
    \item If so, identify and discuss any potential scoring biases revealed in the error analysis: To be done in part 2, if possible.
    \item Conduct experiments or propose new directions to improve model scores on the benchmark: 
    The translate-2 rule includes an example, as it was specifically designed to test whether examples improve performance. We have not analyzed this data yet.
    We would like to investigate how the CoT traces (for Qwen thinking) relate to rule-following performance, qualitatively and quantitatively.
    Sliding window recovered prefix and characters-per-token measurements across the prompts and outputs would add additional evidence as to similarity between conlangs, training data, and model outputs. (This could improve the validity of the score, but ex post facto model performance.)
    The space of rules satisfying our methodological criteria is much larger than what we explored during this project, so replication with new rules is a potential extension.
    To be done in part 2, but new directions are also variously discussed throughout the rest of the paper, in-line with the relevant topics and in Sec.~\ref{appendix:limitationsandfuturework}.
\end{itemize}

Besides these individual pieces of construct validity, we also tried to attend to holistic consistency across our whole argument, by intentionally using convergent, sometimes overlapping, evidence across methods. We cross-check our conclusions using multiple strategies. For statistical testing: Wilcoxon rank-sum tests of difference to understand how models differ directionally in their ability to perform a task, Kolmogorov-Smirnov tests to secondarily understand distributional differences, Fisher-combined p-values to understand overall performance by models in addition to individual, within model differences in performance, rank-biserial effect sizes, and comparisons of means and medians. For output assessment against answer keys: exact matches, string edit distance, and natural language translation metrics to account for both the ability of models to exactly perform tasks as well as closeness or related performance ability. For training data exposure: 13 tokenizers, Google Search, Infini-gram completions, and string matches across 8 training corpora, with tokenizer results cross-checked against each other and against recovered corpus prefixes. Many of our hypotheses additionally cross-check each other, although we only used a subset of them directly in part 1. And, of course, we use multiple conlangs (which act on plausibility in different ways, at different levels of linguistic structure), multiple models (which vary in architectural details), and multiple independent runs for each model.

Finally, a \textbf{broader motivation} for our approach is that many evaluations of LLM behavior, LLM-human interactions, and even LLM internal states produce findings whose interpretation outside their original context is uncertain. A result may reflect a general capability, an artifact of training data or training process, a feature of a particular task, or some interaction among these factors (a familiar challenge in psychology, linguistics, anthropology, and many other sciences that deal with complex systems and behaviour). To minimize such ambiguities, we adopted an existence-proof approach. The value of establishing existence here is that it lets us ask a specific and interesting question, yet avoids placing too much weight on frequency: whether a capability appears in every model, some models, under particular training conditions, or when models are used in specific ways is important, but logically secondary to whether the capability exists at all. \textbf{Showing that meaning-mediated abstraction can occur in at least some PTGA language models is therefore informative even though its mechanisms remain incompletely understood. Moreover, because we observe related patterns across multiple models and conlangs, the phenomenon appears unlikely to be peculiar to a single architecture, training corpus, or task.}

\subsection{Training data exposure}
\label{appendix:trainingdataexposure}

To support our claim that successful rule-following cannot be explained by direct memorization, we performed several complementary checks,\footnote{These checks are in addition to the features of the conlang rules themselves.} plausibly covering a broad range of training data sources:\\
(1)~extensive preliminary testing in which we observed that a wide range of conlang-related-tasks are not trivial for ChatGPT, DeepSeek, and other models (including GPT-4o, gpt-oss, gpt-2, Qwen2, Qwen3 (32B), Qwen3 (8B), Llama 3.1 Instruct (8B), Llama 3.1 Base (70B), Llama 3.2 Base (3B), gemma2 (2B), gemma2 (9B), Olmo 2 Base (7B), Olmo 2 Base (32B), Olmo 2 Instruct (7B), Olmo 2 Instruct (32B));\\
(2)~a Google search for key phrases from our prompts;\\
(3)~a corpus search for key phrases and concepts from our prompts and answer keys across eight popular training corpora;\\
(4)~an InfiniGram language model analysis~\citep{Liu2024InfiniGram}\footnote{We used the $\infty$-gram next-token distribution, and the document search. We looked at what tokens were predicted for each corpus, what prefix was recovered from the prompt for each corpus, and whether there were any matches between specific strings and any corpus.} to assess whether correct answers are locally predictable from corpus $n$-gram statistics (across the same eight corpora);\\
(5)~tracking recovered prompt prefix length across the eight corpora to verify key phrases and answer keys from our tasks only matched at most small substrings within the corpora (recovered prefix length measures the degree to which outputs reuse local sequential patterns present in training corpora, e.g. in the OLMo 2 32B Instruct (4.6T) corpus, the longest found suffix matching the answer key for translate-1, ``Mary gjyqazkjx pxmqiMary. PxmqiMary qtojx.'', is ``tojx.'');\\
(6)~a tokenization analysis (across thirteen popular tokenizers) of gold-star example answers to assess where and to what extent the required outputs require deviation from the orthographic structure of typical LLM training data.

\begin{figure}
    \centering
    \includegraphics[width=0.4\columnwidth]{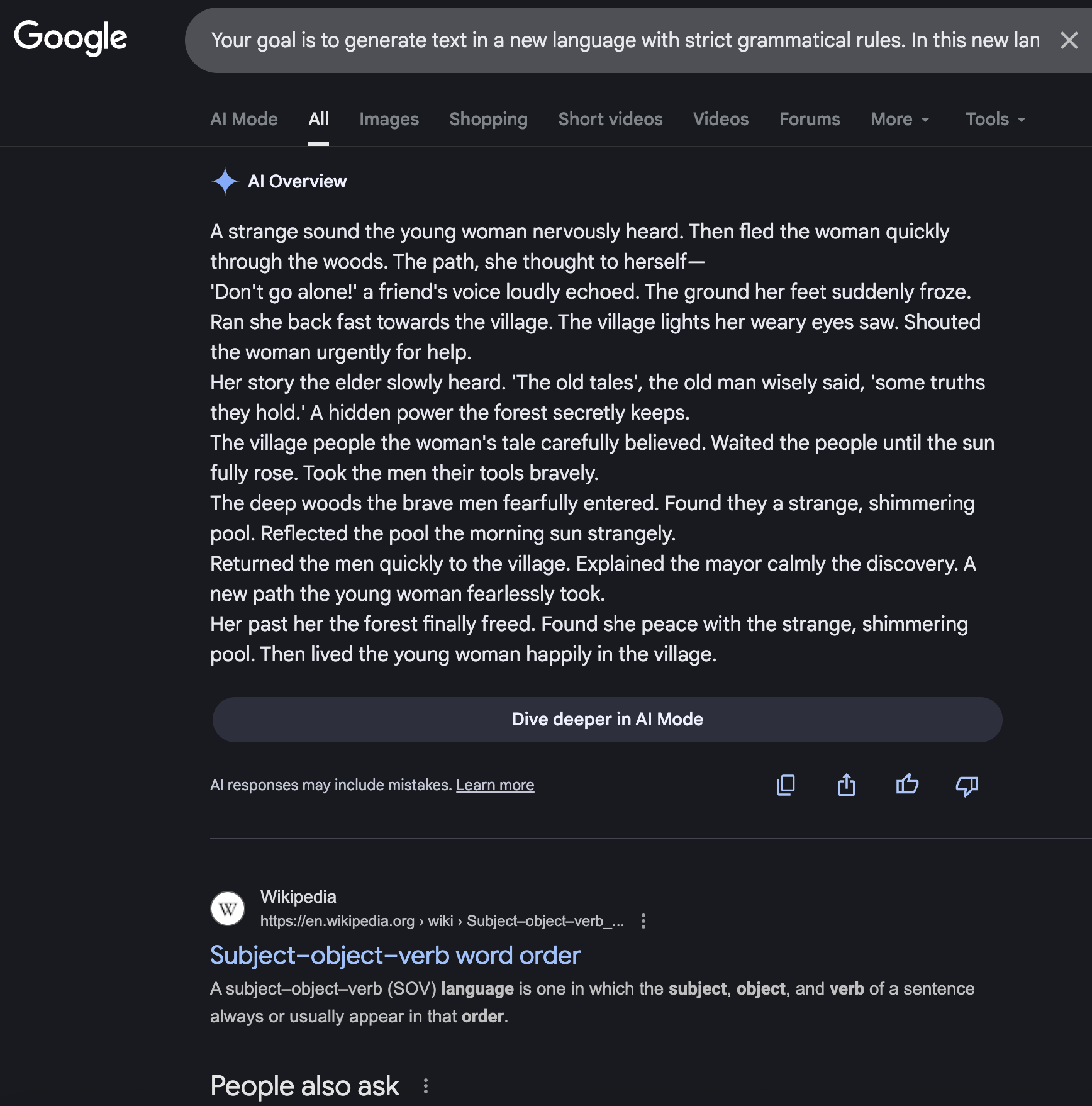}
    \caption{Searching for Google results similar to \textit{wordorderalternation-3}.}
    \label{fig:googleaisearchresults}
\end{figure}

As a simple check, we searched Google for the full text of three rules
(\textit{doublev-1}, \textit{wordorderalternation-3}, \textit{comparison-1}).
An excerpt from \textit{wordorderalternation-3}\footnote{``In this new language, which
is derived from English, speakers follow a word order parity rule. They strictly
alternate between OSV (Object--Subject--Verb) and VSO (Verb--Subject--Object) clauses.''}
did not return any relevant-looking results. For \textit{comparison-1}, we found no
results resembling LLM prompts, although many texts contained generic phrases such as
``The story began when'', which is expected. Interestingly, the AI-generated search
summaries appeared to attempt to follow the rules of \textit{wordorderalternation-3} and
\textit{comparison-1} rather than search and summarize relevant content
(see Figure~\ref{fig:googleaisearchresults}). We then repeated these searches with
Google's `verbatim' option, with similar results.

We used the InfiniGram API~\citep{Liu2024InfiniGram} with setting~5
(``Compute the $\infty$-gram next-token distribution'') to query eight corpora:
OLMo~2~13B~Instruct, OLMoE~1B~7B~Instruct, DCLM-baseline, Dolma-v1.7,
RedPajama, Pile-train, C4-train, and Pile-val. For each query, we examined the next-token distribution conditioned on a given prefix, the longest prefix recovered from each corpus, and whether any verbatim match existed.

We report the \textit{recovered prefix length}: the longest prefix of
a given string that InfiniGram can find in the corpus, measured in characters.
Longer recovered prefixes indicate that more of a string's local sequential structure
is present in training data; shorter prefixes indicate that the string deviates
from attested $n$-gram patterns at an earlier point. Recovered prefix length therefore measures the degree to which a given output reuses local sequential patterns present in training corpora.

\begin{figure}
    \centering
    \includegraphics[width=0.4\columnwidth]{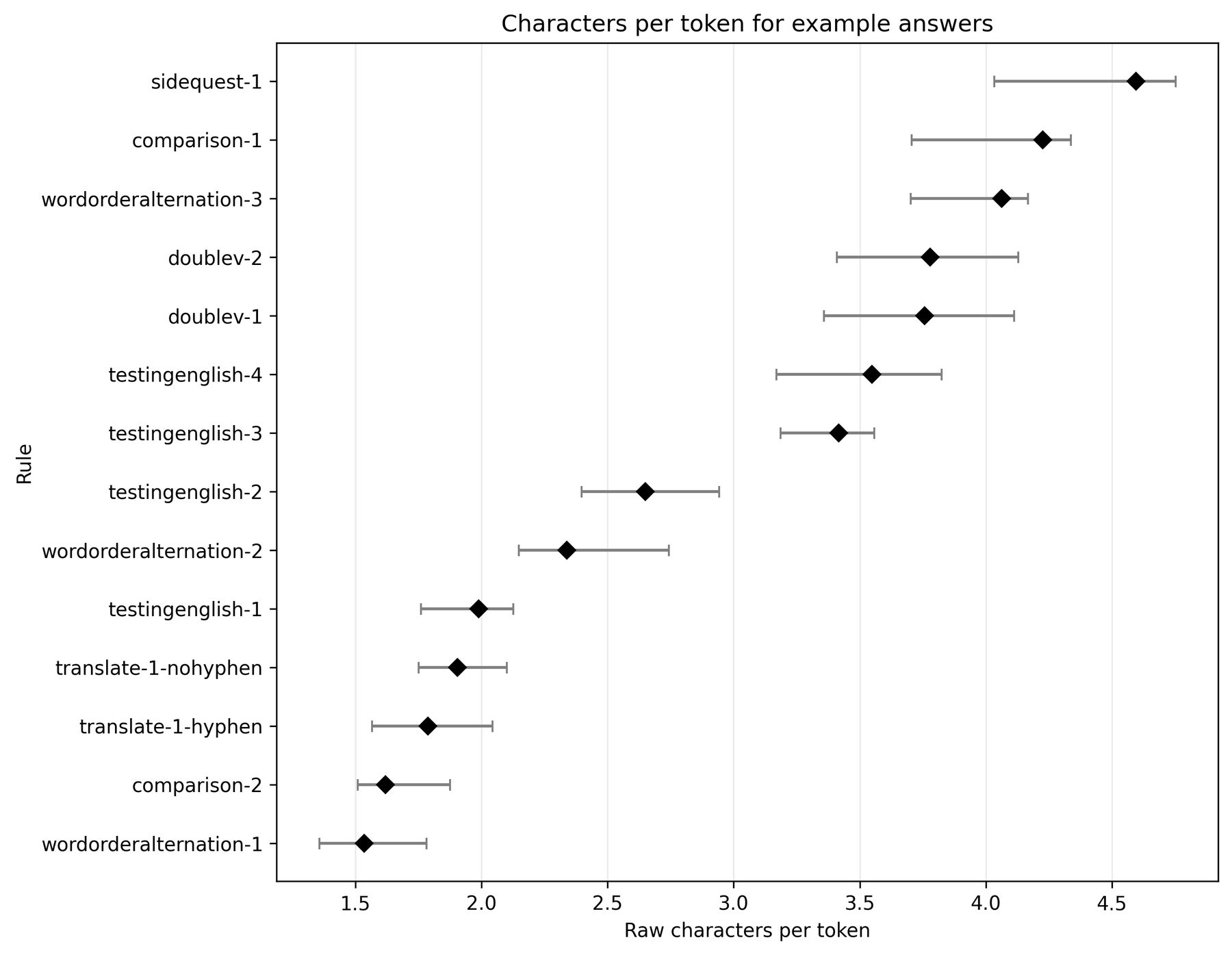}
    \caption{\textbf{Raw characters per token by rule, excluding dan-1.}
    Rules requiring more non-English-like output
    (wordorderalternation-1, wordorderalternation-2, comparison-2)
    have fewer characters per token, consistent with their correct answers containing more character sequences not found in standard tokenizer vocabularies (thus necessitating more, shorter tokens).
    Rules whose outputs remain closer to English (sidequest-1, comparison-1) tokenize into longer, more familiar stems. dan-1 is excluded here because its purely numeric and punctuation-based character set make it a bit unclear to interpret alongside the other rules, which share some characteristics (with each other and English) such as spaces for word boundaries.}
    \label{fig:characters_per_token_no_dan1}
\end{figure}

\begin{figure}
    \centering
    \includegraphics[width=0.4\columnwidth]{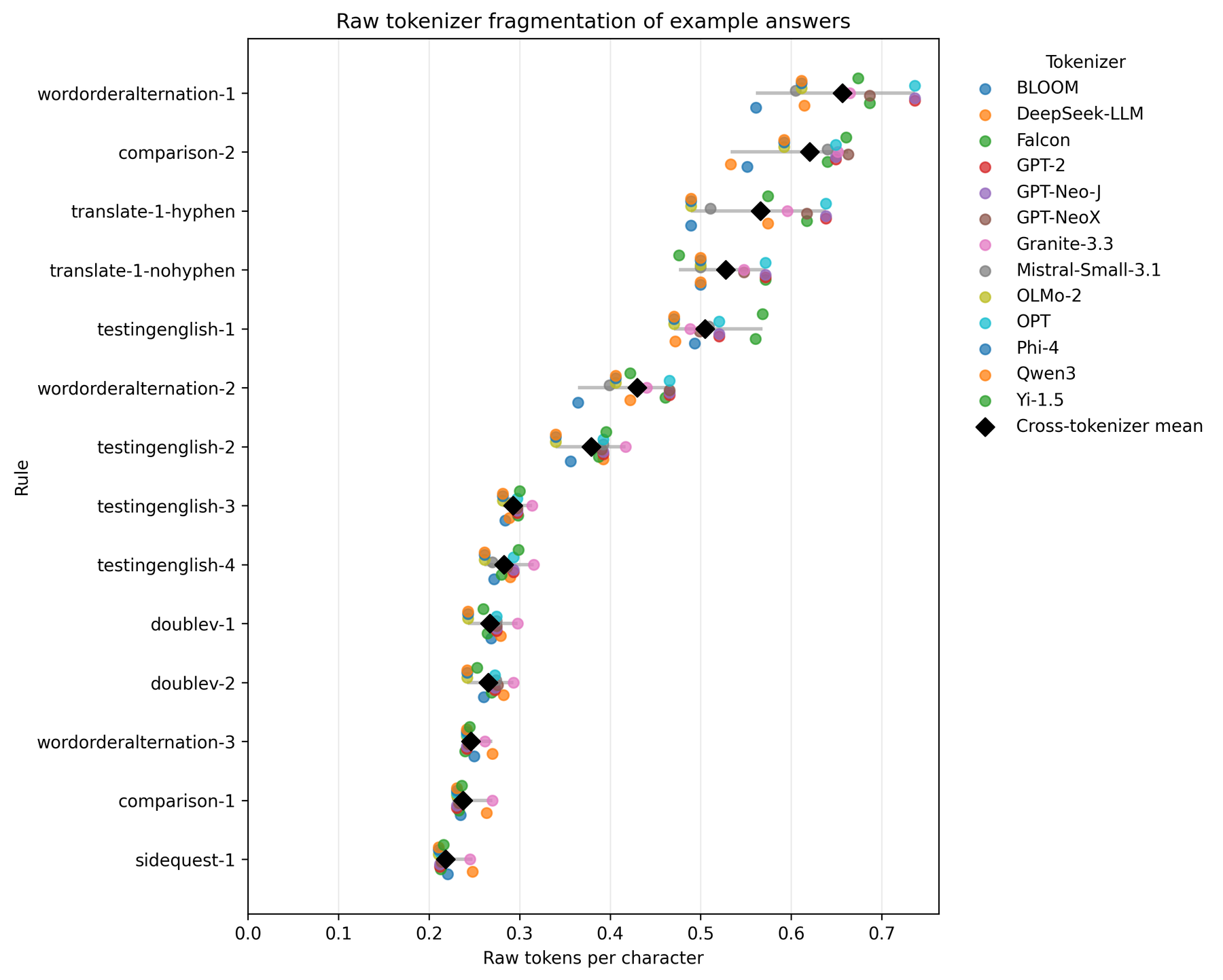}
    \caption{\textbf{Raw tokens per character by rule (excluding dan-1).}
    The inverse of Figure~\ref{fig:characters_per_token_no_dan1}; higher values indicate greater tokenizer fragmentation. We included both figures in case some readers find tokens-per-character more intuitive than characters-per-token, but the information is identical.}
    \label{fig:raw_tokens_per_character_by_rule}
\end{figure}

\begin{figure*}
    \centering
    \includegraphics[width=0.9\linewidth]{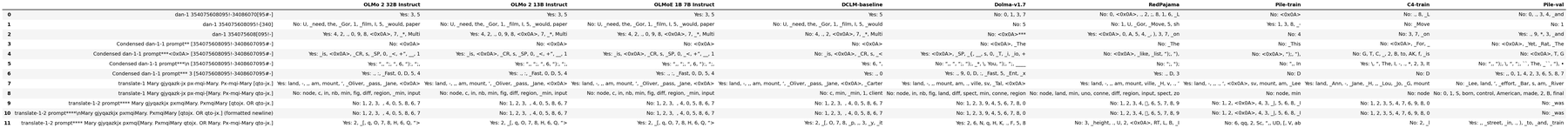}
    \caption{\textbf{To what extent is any part of the correct continuation among the top predicted tokens?}
    For each context, we list all
    top-probability tokens for each corpus in descending probability order, and compare
    against the correct continuation. Correct answers were split into meaningful chunks
    (allowing for minor ordering variation) and both hyphenated and unhyphenated
    translate-1 forms were accepted.
    We split continuations to give greater context or to allow greater freedom in acceptable responses (making positive matches easier). We counted the hyphenated or unhyphenated correct answers for translate-1 as equally valid/ correct (we were flexible on punctuation and casing).}
    \label{fig:trainingdatacontinuationstable.png}
\end{figure*}

As correct conlang output deviates from structural similarity in terms of 1D statistical co-occurrence of the corpus (as roughly reflected by the orderings in Figure~\ref{fig:characters_per_token_no_dan1}, Figure~\ref{fig:raw_tokens_per_character_by_rule}), the continuations predicted by the n-gram LM look less like the correct conlang output. See Table~\ref{tab:top-token-correct-continuation}, Table~\ref{tab:training-search-counts}, Table~\ref{tab:training-search-examples}, Table~\ref{tab:infinigram-results}, and Figure~\ref{fig:trainingdatacontinuationstable.png} for details. The condensed prompts mentioned in Table~\ref{tab:top-token-correct-continuation} are given in full in the caption of Figure~\ref{fig:trainingdatacontinuationstable.png}.

The rules themselves vary as to the resolution of linguistic structure at which they disrupt statistical similarity to English. Note that although both the prefix length and the character-per-token ratios reflect how much the output of each conlang rule deviates from the 1D statistical structure of the training data, neither is a reliable
measure of rule difficulty, since linguistic structure is 3D. A rule whose outputs utilize larger existing linguistic structures (like
wordorderalternation-3, and the double verb rules) will show longer recovered prefixes and higher character-per-token ratios even though it may still be cognitively demanding. 

These checks support two conclusions for the key phrases, concepts, and answer keys that we looked for. First, \textit{direct overlap is absent}: the
distinctive prompts, answers, and lexical sequences were not found in any corpus, ruling out verbatim attestation as an explanation. We did find small matches for substrings (which is both expected and inevitable), and a few matches for a paraphrase of one part of one rule. The rule dan-1 is composed of approximately 17 specific instructions; we found a match to one of those paraphrased instructions. We searched for phrases and paraphrases in order to make it easier to find matches, since it is very unlikely we would find the prompts we wrote verbatim.
Second, \textit{answers are not locally predictable based on corpus statistics alone}: for dan-1 and translate-1, the intended answers are not supported as $n$-gram continuations.

By reasonable induction, we think that since the phrases, answers, and concepts we investigated seem absent from the training data, our rules as a whole are probably absent. Even in the unlikely event this is incorrect, we can focus specifically on the rules whose components we investigated and say that subset of rule-following cannot be explained by memorization, which is already sufficient for our argument. However, in future work, we could more exhaustively verify the absence of more features of our rules and their answers.

\begin{figure}
    \centering
    \includegraphics[width=0.4\columnwidth]{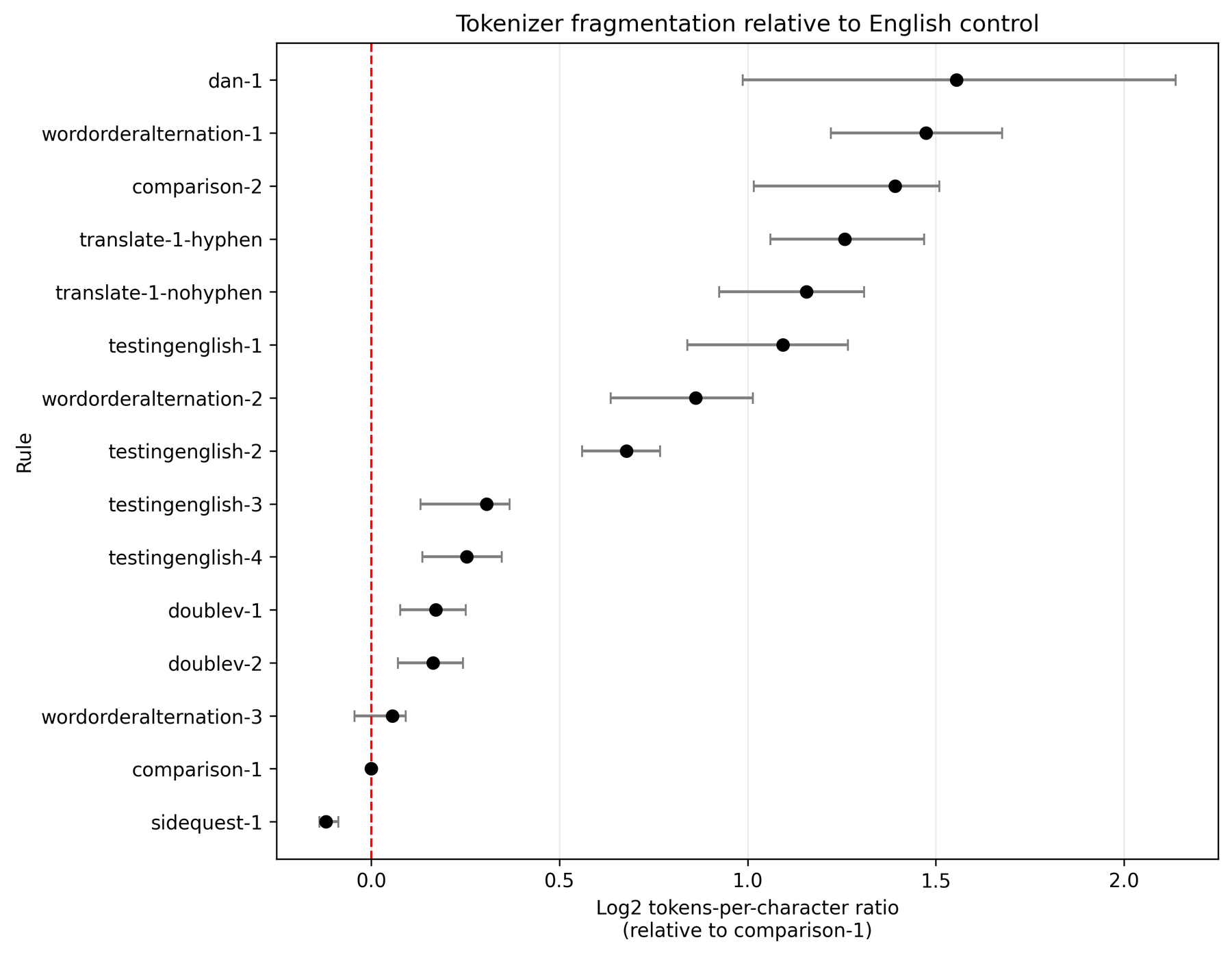}
    \caption{
    \textbf{Tokenizer fragmentation of example answers by rule
    (log$_2$ ratio relative to comparison-1).}
    Rules whose correct outputs most strongly deviate from superficial English statistical patterns
    (dan-1, wordorderalternation-1, comparison-2)
    show substantially higher fragmentation ratios, indicating that the required outputs are not well-supported by the orthographic building blocks of any tokenizer's training distribution. Rules with more English-like correct outputs
    (sidequest-1, comparison-1) cluster near or below the baseline.
    All values are log$_2$-scaled relative to the comparison-1
    (standard English story) answer. These are ``point-in-time'' counts for a single
    example answer per rule; Part~2 will extend this analysis to all model outputs.
    Speculatively, we wonder if we will observe more prosaic output in sidequest-1 than comparison-1, and maybe in doublev-1 versus doublev-2, perhaps as a sign of cognitive tradeoff (in other words, the writing may suffer as the models allocate more resources to rule compliance).}
    \label{fig:tokenizer_fragmentation_log2}
\end{figure}

\begin{figure}
    \centering
    \includegraphics[width=0.4\columnwidth]{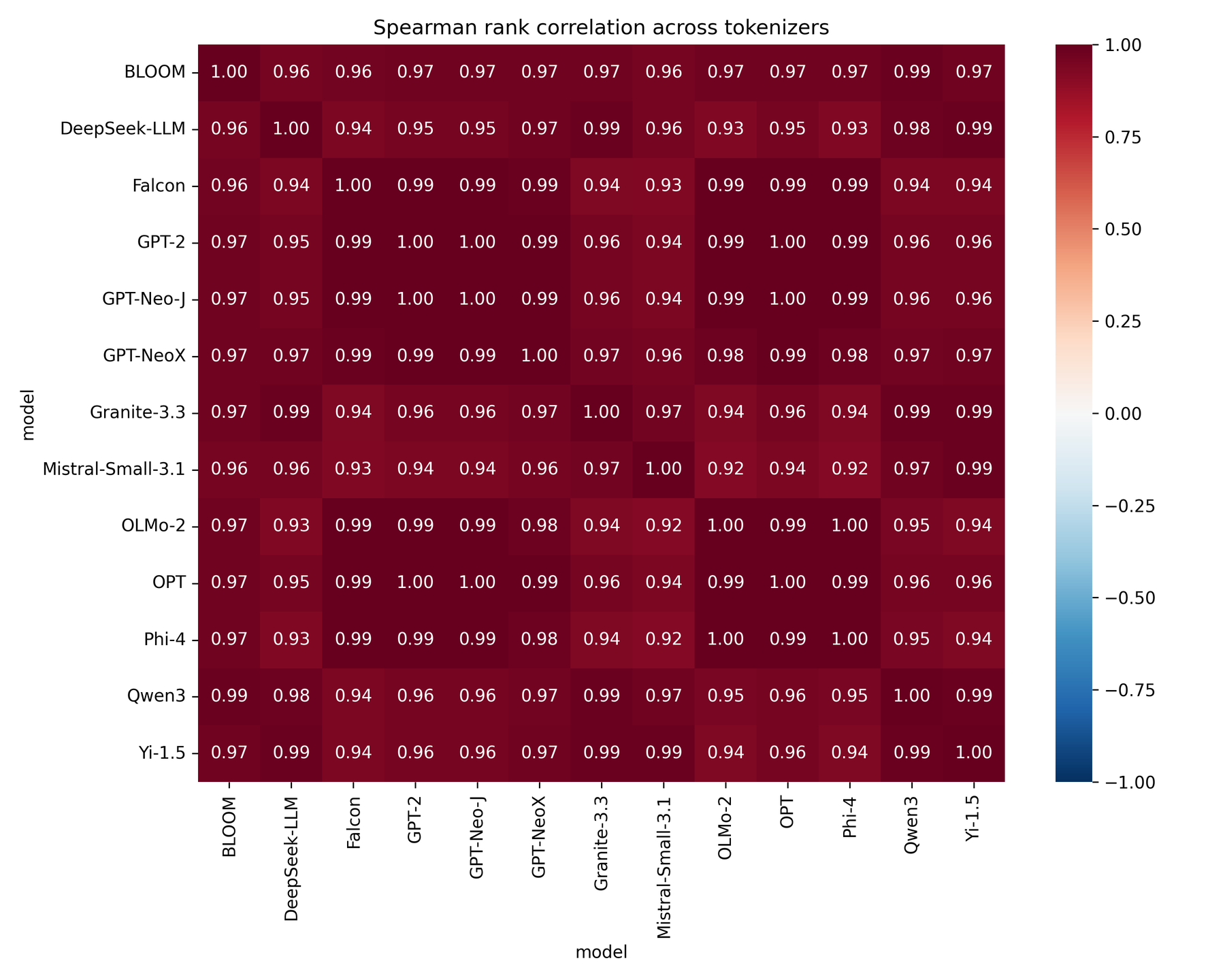}
    \caption{\textbf{Spearman rank correlations of rule fragmentation across 13 tokenizers.}
    Correlations are uniformly high (all $\rho \geq 0.92$; see
    Table~\ref{tab:tokenizer-rank-correlations}), indicating that the ranking of rules
    by tokenizer fragmentation is robust across diverse model families and vocabulary
    designs. The consistent ordering of rules by fragmentation is therefore not an artifact of any single tokenizer's vocabulary, but better explained as a reflection of underlying statistical patterns in the corpora.}
    \label{fig:tokenizer_rank_correlation_heatmap}
\end{figure}

Furthermore, based on the tokenization check, we can see that example correct answers for at least about half of the rules look substantially different from common training data, because they require more tokens-per-response than typical English prose. Tokenizers encode some rule outputs much less efficiently, and since the intended purpose of tokenization is to encode training data as efficiently as possible, this implies that those outputs are not similar to common sequences of training data. 
Tokenizer fragmentation thus provides a proxy for how far a required output deviates from the statistical structure of the training corpus: rules whose correct answers consist of character sequences that are rare or unattested in English will be split into many small tokens (i.e., fewer characters per token), whereas English-like output will be tokenized into familiar, longer stems. We computed characters-per-token~(CPT) and tokens-per-character~(TPC) for gold-standard example answers across 13 tokenizers spanning diverse model families, and observed high consistency across the tokenizers (dan-1 has the least consistent tokenization, due to its answer key being mostly digits). See Figure~\ref{fig:characters_per_token_no_dan1}, Figure~\ref{fig:raw_tokens_per_character_by_rule}, Figure~\ref{fig:tokenizer_fragmentation_log2}, Figure~\ref{fig:tokenizer_rank_correlation_heatmap}, Table~\ref{tab:tokenizer-fragmentation-all-rules}, Table~\ref{tab:raw-tpc-summary}.

Consistent with our predictions, we observe in Table~\ref{tab:cpt-prefix-correlations} that there is a modest correlation between the characters-per-token ratios of our individual example answers and the length of prefix recovered from the eight corpora, and that the thirteen tokenizers behave very consistently in how this metric ranks the conlang rules (Figure~\ref{fig:tokenizer_rank_correlation_heatmap}). These findings are sanity checks for how we are thinking about LMs, tokenization, and the kind of training data these models would have encountered, and validation for the reasoning underlying these checks.

\textbf{These checks (as well as the number of models and runs per rule, the design and diversity of the rules, and the inclusion of prompt variatons) make it implausible that any model we tested could have succeeded by direct recall of training data.}

In part 1, we use one example correct answer for each rule to roughly gauge how much the output of each conlang rule deviates from the 1D statistical structure of the training data. For everything besides dan-1 and translate-1 and 2, the output is open-ended, meaning specific choices can cause high fluctuation (in the character-per-token ratios via the thirteen tokenizers, as well as in the length of the recovered prefix across the eight corpora via InfiniGram~\citep{Liu2024InfiniGram}). 

In part 2, we plan to extend the tokenization analysis from the single representative correct answers used here to include cleaned responses (aggregate and per model), all model output (aggregate and per model), and to use randomly reshuffled and randomly resampled cleaned answers to see how disrupting the linguistic structure at different resolutions impacts the tokenization. We will use this to assess how the rule outputs compare to each other and to typical English, as well as how the models compare to each other. We will further develop the tokenization method used here (mentioned in \citet{zimmerman2024tokensoftoverlookedappetizerlarge}) as an extremely lightweight method for estimating training data exposure in terms of similarity to surface-level statistical patterns seen during training, which is especially useful with closed source models (and can be useful more generally, since they can be run on a laptop and don't require the ability to store large models or large corpora). When we apply the tokenization analysis to the cleaned model outputs, that will provide additional evidence as to whether or not the models systematically move away from ordinary English statistics when attempting the rules, meaning it is another proxy for rule-following.

\begin{table} 
\centering 
\caption{ Training-data search counts across corpora. Columns S1--S21 correspond to the following search expressions: S1 = ``natural language processing'' (for comparison); S2 = ``conlang OR constructed language''; S3 = full translate-1 answer; S4 = ``Mary gjyqazk-jx OR Mary gjyqazkjx''; S5 = ``gjyqazk-jx OR gjyqazkjx''; S6 = full dan-1 answer string; S7 = ``354075608095''; S8 = ``zero is the word boundary OR zero is the space symbol''; S9 = same concept without quotation marks; S10 = ``0 is the word boundary OR 0 is the space symbol''; S11 = variants of ``the digit for nine is the verb to be''; S12 = full constrained bigram lexicon; S13 = reduced constrained bigram lexicon; S14 = extended constrained bigram lexicon; S15 = ``-qw, -jx, and -kz are always the final morphemes in the verb''; S16 = variants of ``-qw, -jx, and -kz''; S17 = ``speakers follow a word order parity rule''; S18 = ``word order parity rule''; S19 = ``include an extra verb in each clause''. For robustness, two additional search strings were tested in OLMo 2 32B Instruct and both returned zero matches (sanity-check permutations of final morphemes; concatenated sanity-check strings (qwkzjx, qwjxkz, kzjxqw, jxkzqw)).
Description row abbreviations: Ex = example; HLC = high level concept; T1-A1--T1-A3 = translate-1 (answer); D1-A1--D1-A2 = dan-1 (answer); D1-P = dan-1 (prompt); D1-C1--D1-C2 = dan-1 (concept); D1-PC = dan-1 (prompt, concept); CBL-1--CBL-3 = constrained bigram lexicon; T1+CBL-1--T1+CBL-2 = translate-1 and constrained bigram lexicon; WOA-1--WOA-2 = wordorderalternation (prompt); DV-P = doublev (prompt).
} \label{tab:training-search-counts} 
\begin{adjustbox}{width=\textwidth}\begin{tabular}{lllrrrrrrrrrrrrrrrrrrr} \toprule Dataset & Tokens & S1 & S2 & S3 & S4 & S5 & S6 & S7 & S8 & S9 & S10 & S11 & S12 & S13 & S14 & S15 & S16 & S17 & S18 & S19\\ 

Description & & Ex & HLC & T1-A1 & T1-A2 & T1-A3 & D1-A1 & D1-A2 & D1-P & D1-C1 & D1-C2 & D1-PC & CBL-1 & CBL-2 & CBL-3 & T1+CBL-1 & T1+CBL-2 & WOA-1 & WOA-2 & DV-P \\

\midrule
OLMo 2 13B Instruct & 4.6T & 2,078,696 & 420,915 & 0 & 0 & 0 & 0 & 0 & 0 & 0 & 4 & 0 & 0 & 0 & 0 & 0 & 0 & 0 & 0 & 0 \\
OLMoE 1B 7B Instruct & 4.6T & 2,078,497 & 420,906 & 0 & 0 & 0 & 0 & 0 & 0 & 0 & 4 & 0 & 0 & 0 & 0 & 0 & 0 & 0 & 0 & 0 \\
DCLM-baseline & 4.3T & 1,899,073 & 405,081 & 0 & 0 & 0 & 0 & 0 & 0 & 0 & 2 & 0 & 0 & 0 & 0 & 0 & 0 & 0 & 0 & 0 \\
Dolma-v1.7 & 2.6T & 915,616 & 100,016 & 0 & 0 & 0 & 0 & 0 & 0 & 0 & 2 & 0 & 0 & 0 & 0 & 0 & 0 & 0 & 0 & 0 \\
RedPajama & 1.4T & 457,312 & 50,004 & 0 & 0 & 0 & 0 & 0 & 0 & 0 & 1 & 0 & 0 & 0 & 0 & 0 & 0 & 0 & 0 & 0 \\
Pile-train & 380B & 81,461 & 9,773 & 0 & 0 & 0 & 0 & 0 & 0 & 0 & 2 & 0 & 0 & 0 & 0 & 0 & 0 & 0 & 0 & 0 \\
C4-train & 200B & 60,874 & 9,224 & 0 & 0 & 0 & 0 & 0 & 0 & 0 & 0 & 0 & 0 & 0 & 0 & 0 & 0 & 0 & 0 & 0 \\
Pile-val & 390M & 76 & 13 & 0 & 0 & 0 & 0 & 0 & 0 & 0 & 0 & 0 & 0 & 0 & 0 & 0 & 0 & 0 & 0 & 0 \\
\bottomrule
\end{tabular}\end{adjustbox}
\end{table}

\begin{table}[t] \centering \caption{Whether any of the 8 training corpora assigned the highest probability to a token that was part of the correct continuation, for dan-1 and translate-1 prompt variants, via InfiniGram~\citep{Liu2024InfiniGram}. We report results at two resolution levels in this summary table: (a)~any token that forms part of the correct answer string, and (b)~a token that matches the first three characters of the correct continuation. Answers were split into meaningful chunks; both hyphenated and unhyphenated correct translate-1 forms were accepted. The prevalence of ``No'' indicates that correct continuations are not locally predictable from training corpus statistics. The multiple tiers of resolution allow us to move from generous to stricter matching for a richer interpretation of the matches. Details available in Figure~\ref{fig:trainingdatacontinuationstable.png}.
} \label{tab:top-token-correct-continuation} 
\begin{adjustbox}{width=\textwidth}
\begin{tabular}{p{9cm}cc} \toprule Rule & \makecell[l]{Any corpus predicts part\\of the correct continuation as \\the most likely token?} & \makecell[l]{Most likely token matches\\ the first three characters \\of the correct continuation?} \\ \midrule dan-1 354075608095!-34086070[95\#-] & No & No \\ dan-1 354075608095!-[340] & No & No \\ dan-1 354075608[095!-] & No & No \\ Condensed dan-1-1 prompt** [354075608095!-3408607095\#-] & No & No \\ Condensed dan-1-1 prompt***\texttt{<0x0A>} [354075608095!-3408607095\#-] & No & No \\ Condensed dan-1-1 prompt***\texttt{\textbackslash n} [354075608095!-3408607095\#-] & Yes & No \\ Condensed dan-1-1 prompt*** 3 [54075608095!-3408607095\#-] & No & No \\ translate-1 Mary gjyqazk-jx px-mqi-Mary. Px-mqi-Mary [qto-jx.] & No & No \\ translate-1 Mary gjyqazk-jx px-mqi-[Mary. Px-mqi-Mary qto-jx.] & No & No \\ translate-1-2 prompt**** Mary gjyqazkjx pxmqiMary. PxmqiMary [qtojx. OR qto-jx.] & No & No \\ translate-1-2 prompt****\texttt{\textbackslash n}Mary gjyqazkjx pxmqiMary. PxmqiMary [qtojx. OR qto-jx.] (formatted newline) & No & No \\ translate-1-2 prompt**** Mary gjyqazkjx pxmqi[Mary. PxmqiMary qtojx. OR Mary. Px-mqi-Mary qto-jx.] & No & No \\ \bottomrule 
\end{tabular} 
\end{adjustbox}
\end{table}

\begin{figure}
    \centering
    \includegraphics[width=0.4\columnwidth]{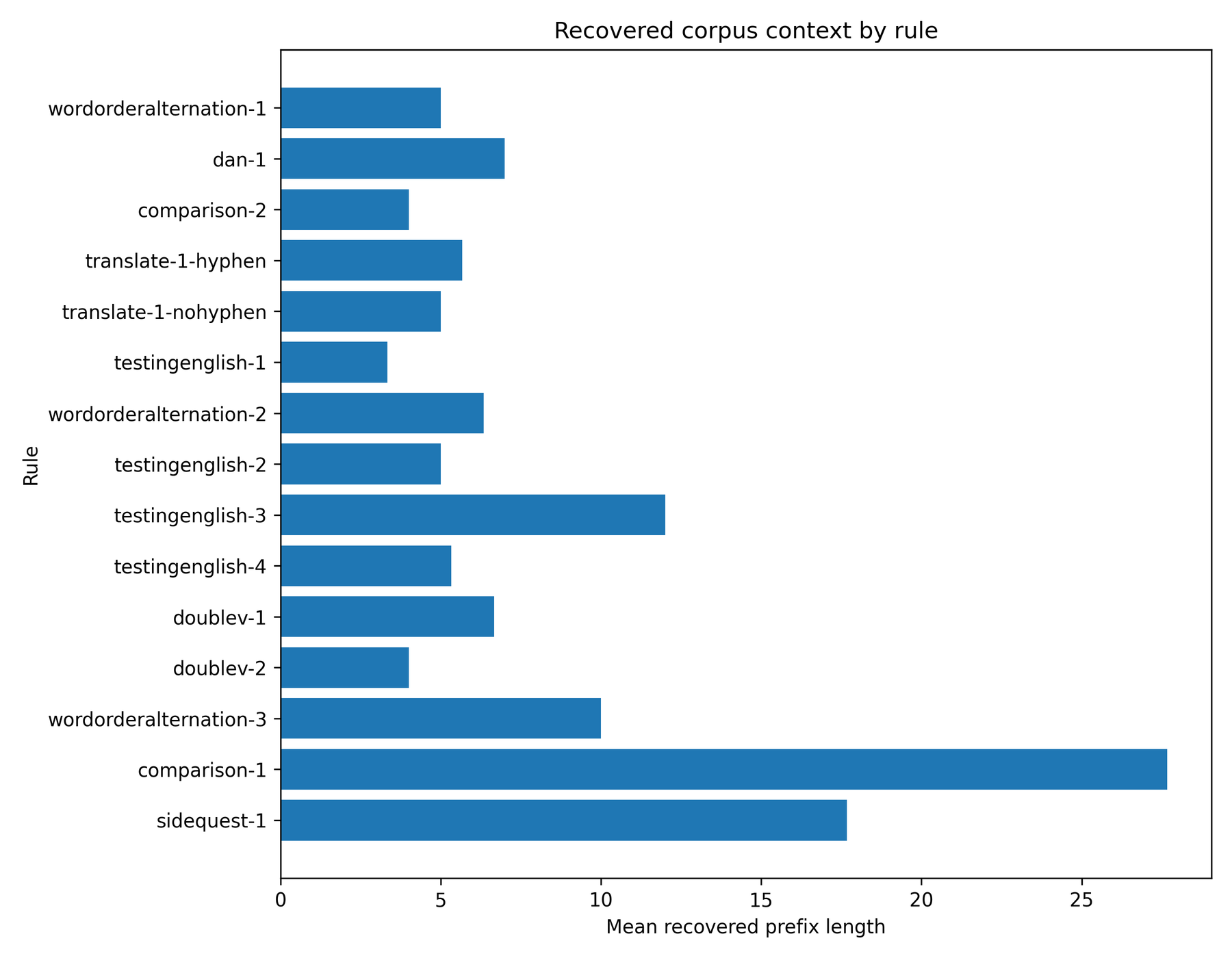}
    \caption{
    \textbf{Mean recovered corpus prefix length by rule.}
    Rules whose correct outputs utilize larger structures borrowed from standard English
    (wordorderalternation-3, comparison-1) have longer
    recovered prefixes, indicating that their local character sequences are more frequently attested in training corpora. Rules with highly non-English-like outputs
    (dan-1, wordorderalternation-1) have shorter recovered prefixes,
    consistent with the tokenization fragmentation results
    (Figure~\ref{fig:characters_per_token_no_dan1}).
    Prefix length is computed on a single gold-standard example answer per rule (see Section~\ref{appendix:trainingdataexposure} for interpretation caveats), and therefore only a ``point-in-time'' proxy.}
    \label{fig:prefix_length_by_rule}
\end{figure}

\begin{figure}
    \centering
    \includegraphics[width=0.4\columnwidth]{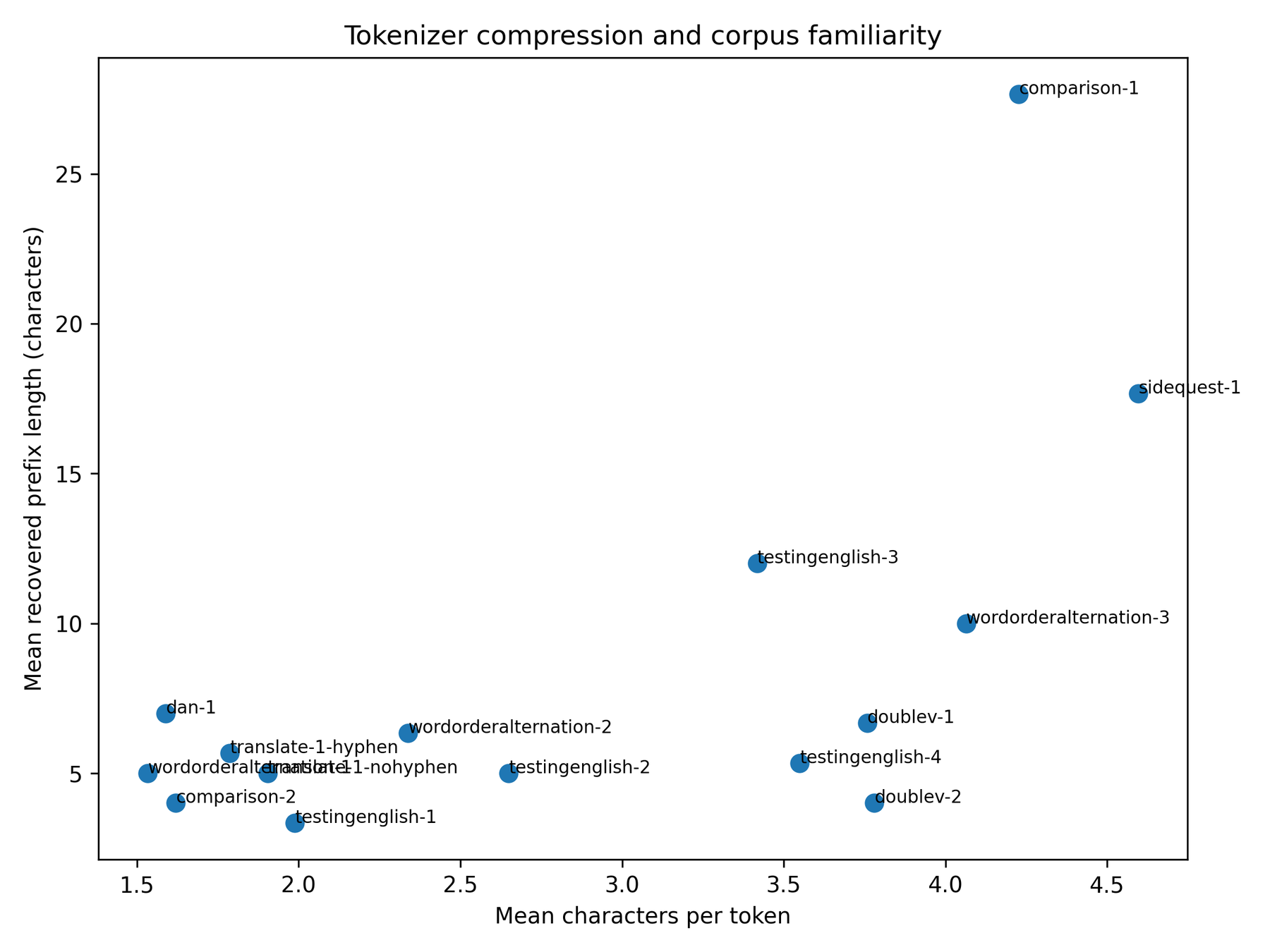}
    \caption{\textbf{Scatter plot of mean characters-per-token (CPT) versus mean recovered corpus prefix length, by rule.}
    A moderate positive correlation is observed (Pearson $r = 0.626$, $p = 0.013$;
    Spearman $\rho = 0.515$, $p = 0.050$; see Table~\ref{tab:cpt-prefix-correlations}),
    consistent with the interpretation that both measures reflect the degree to which a rule's required output aligns with the statistical structure of the training corpus. Rules that deviate more strongly
    from English (low CPT/ high TPC) also tend to fall outside corpus $n$-gram patterns (short recovered prefixes).}
    \label{fig:cpt_vs_prefix_scatter}
\end{figure}

\begin{figure}
    \centering
    \includegraphics[width=0.4\columnwidth]{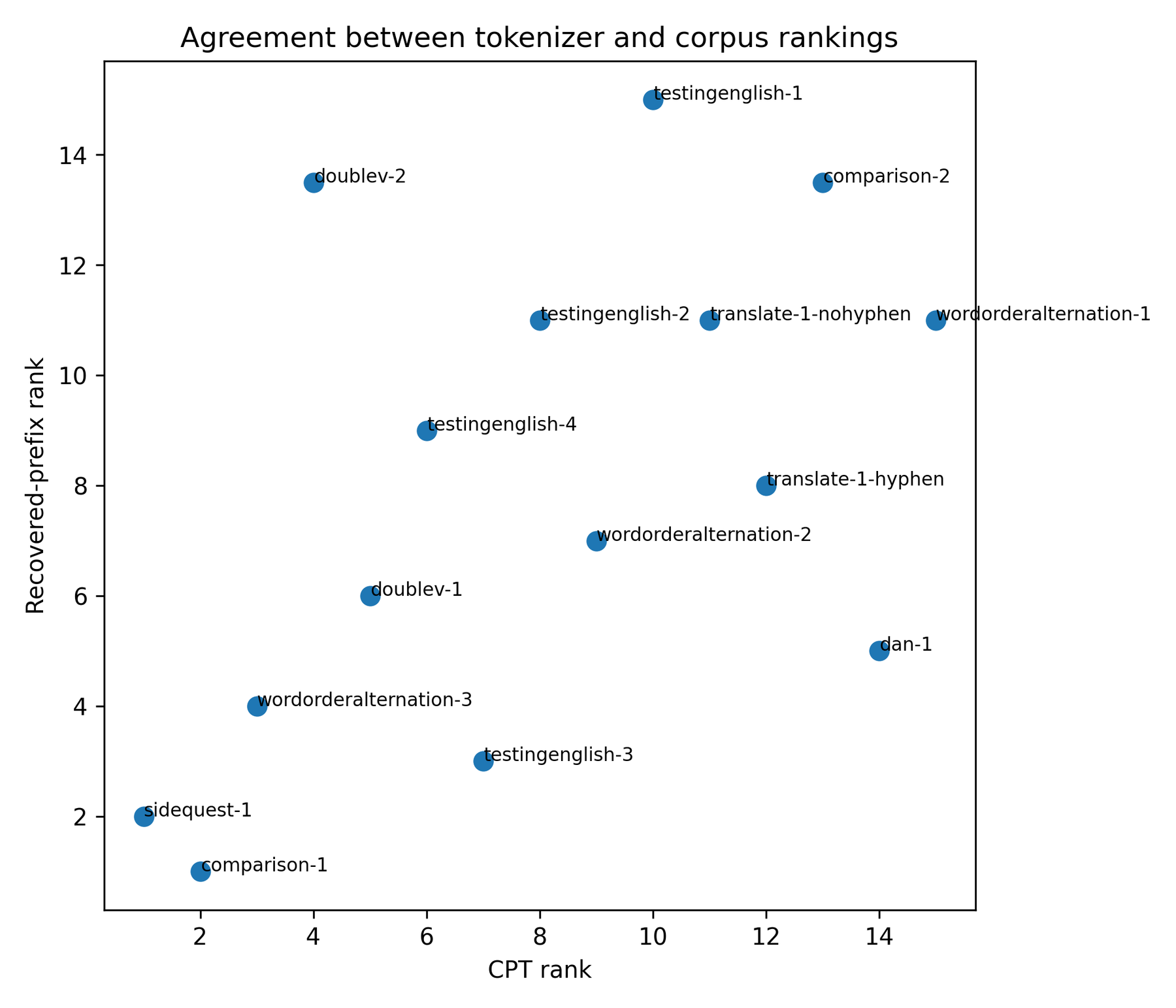}
    \caption{
    \textbf{Rank-rank plot of mean characters-per-token versus mean
    recovered corpus prefix length, by rule.}
    The positive correlation observed in Figure~\ref{fig:cpt_vs_prefix_scatter}
    is preserved at the rank level (Spearman $\rho = 0.515$, $p = 0.050$).
    }
    \label{fig:cpt_rank_vs_prefix_rank}
\end{figure}

Figures~\ref{fig:tokenizer_fragmentation_log2},~\ref{fig:cpt_rank_vs_prefix_rank},~\ref{fig:cpt_vs_prefix_scatter},~\ref{fig:tokenizer_rank_correlation_heatmap},~\ref{fig:characters_per_token_no_dan1},~\ref{fig:raw_tokens_per_character_by_rule},~\ref{fig:prefix_length_by_rule}, Tables~\ref{tab:cpt-prefix-correlations},~\ref{tab:tokenizer-fragmentation-all-rules},~\ref{tab:raw-tpc-summary},~\ref{tab:tokenizer-rank-correlations} are intended to sanity-check how we are thinking about these pieces -- tokenization, tokenizers, corpus statistics, training data, recovered prefixes -- of these systems, to make sure we are using them sensibly in our argument. Figure~\ref{fig:googleaisearchresults} is included to illustrate interesting anecdotal behaviour.

\begin{table}
\caption{Tokenizer fragmentation of example answers relative to the
standard-English comparison-1 answer. ``Prop.'' is the proportion of
tokenizers for which the rule's answer is more fragmented than comparison-1;
``All'' indicates whether this holds for every tokenizer tested (13 tokenizers).}
\label{tab:tokenizer-fragmentation-all-rules}
\begin{adjustbox}{width=\textwidth}\begin{tabular}{lrrrrr}
\toprule
Rule & Mean & Min. & Max. & Prop. & All \\
\midrule
dan-1 & 2.937 & 1.981 & 4.398 & 1.000 & Yes \\
wordorderalternation-1 & 2.776 & 2.332 & 3.196 & 1.000 & Yes \\
comparison-2 & 2.623 & 2.023 & 2.846 & 1.000 & Yes \\
translate-1-hyphen & 2.391 & 2.086 & 2.769 & 1.000 & Yes \\
translate-1-nohyphen & 2.230 & 1.898 & 2.479 & 1.000 & Yes \\
testingenglish-1 & 2.135 & 1.791 & 2.406 & 1.000 & Yes \\
wordorderalternation-2 & 1.817 & 1.555 & 2.020 & 1.000 & Yes \\
testingenglish-2 & 1.600 & 1.474 & 1.702 & 1.000 & Yes \\
testingenglish-3 & 1.237 & 1.095 & 1.291 & 1.000 & Yes \\
testingenglish-4 & 1.193 & 1.099 & 1.272 & 1.000 & Yes \\
doublev-1 & 1.127 & 1.055 & 1.190 & 1.000 & Yes \\
doublev-2 & 1.120 & 1.051 & 1.184 & 1.000 & Yes \\
wordorderalternation-3 & 1.039 & 0.970 & 1.066 & 0.923 & No \\
comparison-1 & 1.000 & 1.000 & 1.000 & 0.000 & No \\
sidequest-1 & 0.920 & 0.909 & 0.941 & 0.000 & No \\
\bottomrule
\end{tabular}\end{adjustbox}
\end{table}

\begin{table}
\caption{Mean and variability of raw tokens-per-character (TPC) values across 13
tokenizers, for gold-standard example answers (answer keys, or real model responses manually modified to be facially valid). Lower TPC indicates more English-like output (tokenized into longer stems); higher TPC indicates greater fragmentation. Rules with highly novel output tend to show greater variability across tokenizers, reflecting differences in how each vocabulary handles unusual character sequences (especially dan-1, with its largely numeric answer key).}
\label{tab:raw-tpc-summary}
\begin{adjustbox}{width=\textwidth}
\begin{tabular}{lrrrrr}
\toprule
rule & mean\_tpc & sd\_tpc & min\_tpc & max\_tpc & range\_tpc \\
\midrule
dan-1 & 0.701000 & 0.249000 & 0.462000 & 1.038000 & 0.577000 \\
wordorderalternation-1 & 0.657000 & 0.059000 & 0.561000 & 0.737000 & 0.176000 \\
comparison-2 & 0.620000 & 0.043000 & 0.533000 & 0.663000 & 0.130000 \\
translate-1-hyphen & 0.566000 & 0.064000 & 0.489000 & 0.638000 & 0.149000 \\
translate-1-nohyphen & 0.527000 & 0.036000 & 0.476000 & 0.571000 & 0.095000 \\
testingenglish-1 & 0.505000 & 0.033000 & 0.470000 & 0.568000 & 0.098000 \\
wordorderalternation-2 & 0.430000 & 0.033000 & 0.365000 & 0.466000 & 0.101000 \\
testingenglish-2 & 0.379000 & 0.026000 & 0.340000 & 0.417000 & 0.077000 \\
testingenglish-3 & 0.293000 & 0.010000 & 0.281000 & 0.314000 & 0.033000 \\
testingenglish-4 & 0.283000 & 0.017000 & 0.261000 & 0.316000 & 0.054000 \\
doublev-1 & 0.267000 & 0.016000 & 0.243000 & 0.298000 & 0.055000 \\
doublev-2 & 0.266000 & 0.016000 & 0.242000 & 0.293000 & 0.051000 \\
wordorderalternation-3 & 0.246000 & 0.009000 & 0.240000 & 0.270000 & 0.030000 \\
comparison-1 & 0.237000 & 0.013000 & 0.231000 & 0.270000 & 0.039000 \\
sidequest-1 & 0.218000 & 0.013000 & 0.210000 & 0.248000 & 0.038000 \\
\bottomrule
\end{tabular}\end{adjustbox}
\end{table}

\begin{table}
\caption{Spearman rank correlations of rule fragmentation across 13 tokenizers.
All correlations are $\geq 0.92$, indicating that the relative ordering of rules
by tokenizer fragmentation is consistent across model families. This mitigates the
possibility that any specific tokenizer idiosyncrasy or even underlying training data idiosyncrasy is responsible for the fragmentation patterns observed.}
\label{tab:tokenizer-rank-correlations}
\begin{adjustbox}{width=\textwidth}
\begin{tabular}{lrrrrrrrrrrrrr}
\toprule
model & BLOOM & DeepSeek-LLM & Falcon & GPT-2 & GPT-Neo-J & GPT-NeoX & Granite-3.3 & Mistral-Small-3.1 & OLMo-2 & OPT & Phi-4 & Qwen3 & Yi-1.5 \\
model &  &  &  &  &  &  &  &  &  &  &  &  &  \\
\midrule
BLOOM & 1.000000 & 0.957000 & 0.964000 & 0.975000 & 0.975000 & 0.975000 & 0.971000 & 0.961000 & 0.975000 & 0.975000 & 0.975000 & 0.986000 & 0.975000 \\
DeepSeek-LLM & 0.957000 & 1.000000 & 0.939000 & 0.954000 & 0.954000 & 0.971000 & 0.989000 & 0.961000 & 0.929000 & 0.954000 & 0.929000 & 0.982000 & 0.986000 \\
Falcon & 0.964000 & 0.939000 & 1.000000 & 0.993000 & 0.993000 & 0.989000 & 0.936000 & 0.925000 & 0.993000 & 0.993000 & 0.993000 & 0.939000 & 0.939000 \\
GPT-2 & 0.975000 & 0.954000 & 0.993000 & 1.000000 & 1.000000 & 0.993000 & 0.957000 & 0.943000 & 0.993000 & 1.000000 & 0.993000 & 0.961000 & 0.957000 \\
GPT-Neo-J & 0.975000 & 0.954000 & 0.993000 & 1.000000 & 1.000000 & 0.993000 & 0.957000 & 0.943000 & 0.993000 & 1.000000 & 0.993000 & 0.961000 & 0.957000 \\
GPT-NeoX & 0.975000 & 0.971000 & 0.989000 & 0.993000 & 0.993000 & 1.000000 & 0.968000 & 0.957000 & 0.979000 & 0.993000 & 0.979000 & 0.968000 & 0.971000 \\
Granite-3.3 & 0.971000 & 0.989000 & 0.936000 & 0.957000 & 0.957000 & 0.968000 & 1.000000 & 0.968000 & 0.936000 & 0.957000 & 0.936000 & 0.989000 & 0.989000 \\
Mistral-Small-3.1 & 0.961000 & 0.961000 & 0.925000 & 0.943000 & 0.943000 & 0.957000 & 0.968000 & 1.000000 & 0.921000 & 0.943000 & 0.921000 & 0.975000 & 0.986000 \\
OLMo-2 & 0.975000 & 0.929000 & 0.993000 & 0.993000 & 0.993000 & 0.979000 & 0.936000 & 0.921000 & 1.000000 & 0.993000 & 1.000000 & 0.946000 & 0.936000 \\
OPT & 0.975000 & 0.954000 & 0.993000 & 1.000000 & 1.000000 & 0.993000 & 0.957000 & 0.943000 & 0.993000 & 1.000000 & 0.993000 & 0.961000 & 0.957000 \\
Phi-4 & 0.975000 & 0.929000 & 0.993000 & 0.993000 & 0.993000 & 0.979000 & 0.936000 & 0.921000 & 1.000000 & 0.993000 & 1.000000 & 0.946000 & 0.936000 \\
Qwen3 & 0.986000 & 0.982000 & 0.939000 & 0.961000 & 0.961000 & 0.968000 & 0.989000 & 0.975000 & 0.946000 & 0.961000 & 0.946000 & 1.000000 & 0.989000 \\
Yi-1.5 & 0.975000 & 0.986000 & 0.939000 & 0.957000 & 0.957000 & 0.971000 & 0.989000 & 0.986000 & 0.936000 & 0.957000 & 0.936000 & 0.989000 & 1.000000 \\
\bottomrule
\end{tabular}\end{adjustbox}
\end{table}

\begin{table}
\centering
\caption{Correlation between mean characters-per-token (CPT) and mean recovered corpus prefix length. N rules encompasses the suite of rules, but including both hyphenated and non-hyphenated translate-1 answers, and excluding translate-2 because its answer key is identical to that of translate-1). Both correlations are statistically significant, supporting the interpretation that CPT and prefix length at least partially reflect the same signal.
}
\label{tab:cpt-prefix-correlations}
\begin{tabular}{lrrr}
\toprule
correlation & coefficient & p\_{value} & n\_{rules} \\
\midrule
Pearson & 0.626000 & 0.012600 & 15 \\
Spearman & 0.514800 & 0.049600 & 15 \\
\bottomrule
\end{tabular}
\end{table}

\begin{table} \centering \caption{Example documents returned by searches for superficially similar phrases. The retrieved examples concern an embroidery-pattern convention in which zero or a blank space is used for padding, and a forum question about \textit{mailto} links, rather than a constructed-language rule. The matched search string is a paraphrase of that part of the rule, not a verbatim match to any part of the prompt. Document sources: \url{http://embroidermodder.org/docs/manual/}, \url{https://stackoverflow.com/questions/3868315/invoke-click-a-mailto-link-with-jquery-javascript}, retrieved from the InfiniGram UI. There are at most 4 matches per corpus to this n-gram search string, across all the corpora, and every example surfaced by the InfiniGram UI matched one of these two use cases.} \label{tab:training-search-examples} \begin{tabular}{p{0.24\textwidth} p{0.68\textwidth}} \toprule \textbf{Search expression} & \textbf{Retrieved excerpt} \\ \midrule 0 is the word boundary OR 0 is the space symbol & ``\ldots use a vertical flip (embPattern\_flip)\ldots 0x2 0 is the space symbol 0, so when padding either 0 or space is preferred \ldots'' \\ \\ 0 is the word boundary OR 0 is the space symbol & ``\ldots i.e. use only the C Standard Library\ldots 0x2 0 is the space symbol 0, so when padding either 0 or space is preferred \ldots'' \\ \\ 0 is the word boundary OR 0 is the space symbol & ``\ldots the working answer for me, tested in chrome, IE and firefox\ldots \texttt{window.location.href = 'mailto:address@dmail.com?subject=Hello there\&body=This is the body';} \ldots \%2 0 is the space symbol 0 that should be used, but it worked for me as well with normal space.'' \\ \bottomrule \end{tabular} \end{table}

\begin{table}[t] \centering \caption{InfiniGram next-token distributions conditioned on the first 100 characters of example answers. Entries report the highest-probability continuation, the continuation token, and the longest prefix recovered from the corpus. For dan-1 and translate-1, probabilities were conditioned on the context up to the final word-equivalent boundary to give the model the best possible chance of predicting the correct answer. For all other rules, the examples are intended to illustrate how the corpus might continue a valid answer. For doublev-1, we initially examined all available corpora; because the results were qualitatively similar, we subsequently consider only a representative subset of corpora. None of the highest-probability continuations match the correct dan-1 or translate-1 answers, confirming that these are not predictable from corpus statistics alone.
} \label{tab:infinigram-results} \begin{adjustbox}{width=\textwidth}\begin{tabular}{ p{0.52\textwidth} p{0.15\textwidth} p{0.15\textwidth} p{0.15\textwidth} } \toprule \textbf{Rule and excerpt} & \textbf{OLMo 2 32B} & \textbf{Dolma-v1.7} & \textbf{Pile-val} \\ \midrule doublev-1 Every morning, I wakeupsqw yawn and makesqw brew coffee. I walksqw run to the park to readsqw flip a & 36\%, \_coin, ``flip a'' & 40\%, \_coin, ``w flip a'' & 27\%, \_coin, ``flip a'' \\ doublev-2 The sun rose studyqw over the horizon. Birds chirped twerkqw in the trees. A traveler packed snortqw & 67\%, dz, ``ortqw'' & 40\%, dz, ``ortqw'' & 12\%, \_\&, ``qw'' \\ wordorderalternation-1 ZP GV QD ZD-qw. VF-jx QD GV QM. ZP GV QH PX-jx. QF-kz ZK GV QM. QM GV JY HJ-jx. DX-qw GJ JG ZP. JG Z & 29\%, in, ``. JG Z'' & 39\%, uma, ``. JG Z'' & 25\%, L, ``G Z'' \\ wordorderalternation-2 drev lira rap nem-kz. vex-jx vol briz drev. silu lira sur zor-jx. kra-jx rap lira blad. best lira vo & 100\%, ye, ``lira vo'' & 33\%, \_I, ``lira vo'' & 100\%, in, ``ra vo'' \\ wordorderalternation-3 The forest the hiker entered. Stumbled the hiker over roots. The sun the hiker saw. Shone the sun br & 60\%, ighter, ``the sun br'' & 100\%, ighter, ``one the sun br'' & 50\%, ighter, ``sun br'' \\ testingenglish-1 ``ZX, ZD QM,'' BX VX to PQ, ``XF VW ZF ZJ ZD QF.'' PQ, HJ in his QB, didn't VFPX. BX, his ZX PXZD, FQWX: & 51\%, \_‘, ``QWX:'' & 25\%, US, ``QWX:'' & 23\%, \_, ``X:'' \\ testingenglish-2 QB cat VX QJ box. ZP cat ZD black. ZD cat VX hunting QJ mouse. QJ mouse VX small VF VX quick. ZD cat & 100\%, \_food, ``ZD cat'' & 100\%, \_food, ``ZD cat'' & 17\%, ast, ``cat'' \\ testingenglish-3 Once in a faraway land, there ZX a small village. The villagers all VX happily, for they had a wise & 24\%, \_and, ``they had a wise'' & 16\%, \_and, ``they had a wise'' & 15\%, \_man, ``a wise'' \\ testingenglish-4 Once there was a brave VF. This VF lived in a remote PX, far away from the bustling ZD of the QT. Ev & 100\%, ident, ``QT. Ev'' & 100\%, ident, ``QT. Ev'' & 43\%, idence, ``. Ev'' \\ translate-1-hyphen Mary gjyqazk-jx px-mqi-Mary. Px-mqi-Mary [qto-jx.] & 30\%, 2, ``-Mary '' & 24\%, 1, ``-Mary '' & 100\%, \_was, ``Mary '' \\ translate-1-nohyphen Mary gjyqazkjx pxmqiMary. PxmqiMary [qtojx.] & 26\%, 1, ``Mary '' & 27\%, 1, ``Mary '' & 100\%, \_was, ``Mary '' \\ comparison-1 Once, in a quiet mountain town, lived a young man named Eli. He was known for his ability to fix any & 100\%, \_kind, ``known for his ability to fix any'' & 100\%, \_engine, ``. He was known for his ability to fix any'' & 7\%, \_sentence, ``to fix any'' \\ comparison-2 XJ QV ZJ QP ZD FV QE. VW QX VK HJ QE FZ QD. ZP JG BQ XG QM QE ZF. VX QM JZ FV QX XZ MZ. QD JY GV QV & 100\%, ., ``GV QV'' & 75\%, \_Rep, ``GV QV'' & 39\%, irt, ``QV'' \\ sidequest-1 Once upon a time, in the heart of a dense forest, lived a kind-hearted man named John. He was not an & 100\%, \_eth, ``John. He was not an'' & 20\%, \_eth, ``John. He was not an'' & 33\%, \_old, ``. He was not an'' \\ dan-1 354075608095!-34086070[95\#-] & 50\%, 3, ``34086070'' & 50\%, 0, ``34086070'' & 43\%, 0, ``86070'' \\ \bottomrule \end{tabular}\end{adjustbox} \end{table}

The condensed prompts in Figure~\ref{fig:trainingdatacontinuationstable.png} are:
\begin{itemize}[nosep]
    \item \textbf{**}~Condensed dan-1-1 prompt (shortened to fit InfiniGram's 1000-character limit): ``The digit, for `zero' is the word boundary or space symbol. Unless otherwise indicated, assume similarity to English. The digit for `nine' is the verb `to be'. `Eight' is `cat'. The digit for `three' is any English pronoun. `Three' is male/ gender neutral. The digit for `five' is appended to indicate female gender, and the digit for `four' indicates possession. The digit for `six' means possession for non-pronoun words. The digit `seven' is `male parent'; the digit `seven' suffixed with the digit `five' is `female parent'. Present tense is default, and past tense is marked with five. The exclamation mark means positive emotional valence, and attaches to the verb stem. The hashtag symbol indicates negative emotional valence. Translate the sentences: `Her mother’s cat was happy. His cat’s father was sad.''' 
    \item \textbf{***}~Further condensed (``Remember to'' removed) to satisfy the 1000-character limit; tested with both \texttt{<0x0A>} and newline representations. The \texttt{<0x0A>} byte could be part of rendering a correct character in some encoding, but we are not exploring that possibility here. Cursory research suggests it is likely to be newline or potentially 10, neither of which would be part of the correct continuation for dan-1. To be as generous as possible, we incorporated that into the prompt variations in case correct continuations follow a newline. 
    \item \textbf{****}~translate-1-2 prompt:``Your job is to translate text in a new language according to rigid grammatical rules. Language A: no pronouns, SOV word order, verbs take tense suffix -qw=past, -jx=present, -kz=future. Language B: pronouns vf="he", px="she", zd="they", SVO word order, familial relationships = relative pronoun + temporal metaphor for relationship + person (e.g., John’s grandfather = vf-before-before-John = vfmqimqiJohn), verbs take tense suffix -jx=past, -kz=present, -qw=future. Example in language A: Mary Marypq-gvijw gjyqazkqw. Marypq-gvijw qtoqw. (Mary speak-past to Mary-possessive-mother. Mary-mother smile-past.) The roots of the verbs in this example are the same in language A and language B. Task = Translate this example into Language B, preserving meaning and applying all of B's rules:''
\end{itemize}

\subsection{All conlang rules}
\label{appendix:allruletext}

The text for each of the rules (1 of the 3 prompt variations) are provided in Table~\ref{tab:rules}. We did not include sidequest-1-fix, which is identical to sidequest-1, but with 3 prompt variations instead of 2.

\begin{longtable}{l>{\raggedright\arraybackslash}p{3cm}c>{\raggedright\arraybackslash}p{5cm}}
\caption{Columns taken from our prompt spreadsheet; prompt variations not included.} \label{tab:rules} \\
\toprule
\tiny
                & Family/ Gloss & Rule & Base Prompt \\
\midrule
\endfirsthead
\toprule
\tiny
                & Family/ Gloss & Rule & Base Prompt \\
\midrule
\endhead
\midrule
\multicolumn{4}{r}{Continued on next page} \\
\midrule
\endfoot
\bottomrule
\endlastfoot
doublev-1 & doublev & 1/2 & Your goal is to generate text in a new language with strict grammatical rules. In this new version of English, speakers include an extra verb in each clause. In other words, everywhere there would normally be a single verb, there are now two consecutive verbs. One of the verbs is irrelevant to the meaning of the sentence. The irrelevant verb could either be first or second, and it is marked with the terminal morpheme -qw. The irrelevant verb should result in no other changes to the sentence. Otherwise, this language is identical to English. Please write a short story (20 sentences) in this language as if you speak this language: \\
doublev-2 & doublev (emphasized irrelevance) & 2/2 & Your goal is to generate text in a new language with strict grammatical rules. In this new version of English, speakers include an extra verb in each clause. In other words, everywhere there would normally be a single verb, there are now two consecutive verbs. One of the verbs is irrelevant to the meaning of the sentence. It doesn’t just fail to add to the meaning of the sentence, but it ought to feel ``random'' in the context of the sentence. In other words, the two verbs should not be closely related, synonymous, or antonymous. The irrelevant verb could either be first or second, and it is always marked with the terminal morpheme -qw. The irrelevant verb should result in no other changes to the sentence. Otherwise, this language is identical to English. Please write a short story (20 sentences) in this language as if you speak this language: \\
wordorderalternation-1 & wordorderalternation (restricted lexicon) & 1/3 & Your goal is to generate text in a new language with strict grammatical rules. In this new language, speakers follow a word order parity rule. They strictly alternate between OSV (Object–Subject–Verb) and VSO (Verb–Subject–Object) clauses. When a clause is left incomplete, e.g. due to interruption, the new clause flips to the other word order. Verbs are inflected for tense with suffixes: -qw indicates past tense, -jx indicates present tense, and -kz indicates future tense. All verbs must be inflected with one of these suffixes. The tense suffixes -qw, -jx, and -kz are always the final morphemes in the verb. Those morphemes only have one meaning (indicating verb tense); they are not polysemous and have no homonyms. Adverbs always appear in phrases before the verb when word order is OSV, and after the verb when word order is VSO. In other words, for OSV clauses, the verb is strictly final, and for VSO clauses, the verb is strictly first. Besides the tense suffixes, the rest of the lexicon is created from this set of graphemes, optionally joined with each other: VF, PX, ZD, QT, ZP, QO, DX, HJ, GV, JW, QC, JY, GJ, QB, PQ, JG, BZ, MX, QM, MZ, QF, WJ, ZQ, XR, ZK, CX, FX, FV, BX, VW, VJ, MQ, QV, ZF, QE, YJ, GX, KX, XG, QD, XJ, SX, VZ, VX, WV, YQ, BQ, GQ, VK, ZJ, XK, QP, HX, FZ, QH, QJ, JZ, VQ, KQ, XD, QX, WX, FQ, XZ, ZX, A, E, I, O, U.
Please write a short story (20 sentences) in this language as if you speak this language: \\
wordorderalternation-2 & wordorderalternation (any lexicon, with verb marking) & 2/3 & Your goal is to generate text in a new language with strict grammatical rules. In this new language, speakers follow a word order parity rule. They strictly alternate between OSV (Object–Subject–Verb) and VSO (Verb–Subject–Object) clauses. When a clause is left incomplete, e.g. due to interruption, the new clause flips to the other word order. Verbs are inflected for tense with suffixes: -qw indicates past tense, -jx indicates present tense, and -kz indicates future tense. All verbs must be inflected with one of these suffixes. The tense suffixes -qw, -jx, and -kz are always the final morphemes in the verb. Those morphemes only have one meaning (indicating verb tense); they are not polysemous and have no homonyms. Adverbs always appear in phrases before the verb when word order is OSV, and after the verb when word order is VSO. In other words, for OSV clauses, the verb is strictly final, and for VSO clauses, the verb is strictly first.
Please write a short story (20 sentences) in this language as if you speak this language: \\
wordorderalternation-3 & wordorderalternation (English lexicon) & 3/3 & Your goal is to generate text in a new language with strict grammatical rules. In this new language, which is derived from English, speakers follow a word order parity rule. They strictly alternate between OSV (Object–Subject–Verb) and VSO (Verb–Subject–Object) clauses. When a clause is left incomplete, e.g. due to interruption, the new clause flips to the other word order. Adverbs always appear in phrases before the verb when word order is OSV, and after the verb when word order is VSO. In other words, for OSV clauses, the verb is strictly final, and for VSO clauses, the verb is strictly first. Besides these differences in syntax, the language is identical to present day English. Please write a short story (20 sentences) in this language as if you speak this language: \\
testingenglish-1 & testingenglish (replace content words) & 1/4 & Your goal is to generate text in a new language with strict grammatical rules. In this new language, speakers follow similar rules to English. The function words (prepositions, articles, pronouns, auxiliary verbs, etc.) and inflectional morphemes (verb tense, plural markers, adverb and adjective affixes, etc.) are all from English. However, the content words — nouns, adjectives, adverbs, and main verbs — come from this lexicon: VF, PX, ZD, QT, ZP, QO, DX, HJ, GV, JW, QC, JY, GJ, QB, PQ, JG, BZ, MX, QM, MZ, QF, WJ, ZQ, XR, ZK, CX, FX, FV, BX, VW, VJ, MQ, QV, ZF, QE, YJ, GX, KX, XG, QD, XJ, SX, VZ, VX, WV, YQ, BQ, GQ, VK, ZJ, XK, QP, HX, FZ, QH, QJ, JZ, VQ, KQ, XD, QX, WX, FQ, XZ, ZX. English auxiliary verbs (e.g., be, have, do, will, can) remain in the language; all English main verbs are replaced. Besides the novel lexical items that replace English content words, the language is identical to English.
Please write a short story (20 sentences) in this language as if you speak this language: \\
testingenglish-2 & testingenglish (replace function words) & 2/4 & Your goal is to generate text in a new language with strict grammatical rules. In this new language, speakers follow similar rules to English. The content words — nouns, adjectives, adverbs, and main verbs — are all from English. However, the function words (prepositions, articles, pronouns, auxiliary verbs, etc.) and inflectional morphemes (verb tense, plural markers, adverb and adjective affixes, etc.) come from this lexicon: VF, PX, ZD, QT, ZP, QO, DX, HJ, GV, JW, QC, JY, GJ, QB, PQ, JG, BZ, MX, QM, MZ, QF, WJ, ZQ, XR, ZK, CX, FX, FV, BX, VW, VJ, MQ, QV, ZF, QE, YJ, GX, KX, XG, QD, XJ, SX, VZ, VX, WV, YQ, BQ, GQ, VK, ZJ, XK, QP, HX, FZ, QH, QJ, JZ, VQ, KQ, XD, QX, WX, FQ, XZ, ZX. While all English main verbs remain in the language, English auxiliary verbs (e.g., be, have, do, will, can) are replaced. Besides the novel lexical items that replace English function words, the language is identical to English.
Please write a short story (20 sentences) in this language as if you speak this language: \\
testingenglish-3 & testingenglish (replace verbs) & 3/4 & Your goal is to generate text in a new language with strict grammatical rules. In this new language, speakers follow similar rules to English. The function words (prepositions, articles, pronouns, auxiliary verbs, etc.) and inflectional morphemes (verb tense, plural markers, adverb and adjective affixes, etc.) are all from English, as are most categories of content words, specifically the nouns, adjectives, and adverbs. However, the main verbs come from this lexicon: VF, PX, ZD, QT, ZP, QO, DX, HJ, GV, JW, QC, JY, GJ, QB, PQ, JG, BZ, MX, QM, MZ, QF, WJ, ZQ, XR, ZK, CX, FX, FV, BX, VW, VJ, MQ, QV, ZF, QE, YJ, GX, KX, XG, QD, XJ, SX, VZ, VX, WV, YQ, BQ, GQ, VK, ZJ, XK, QP, HX, FZ, QH, QJ, JZ, VQ, KQ, XD, QX, WX, FQ, XZ, ZX. In other words, no English main verbs are used in this language, but English auxiliary verbs are. Besides the novel lexical items that replace the English main verbs, the language is identical to English.
Please write a short story (20 sentences) in this language as if you speak this language: \\
testingenglish-4 & testingenglish (replace nouns) & 4/4 & Your goal is to generate text in a new language with strict grammatical rules. In this new language, speakers follow similar rules to English. The function words (prepositions, articles, pronouns, auxiliary verbs, etc.) and inflectional morphemes (verb tense, plural markers, adverb and adjective affixes, etc.) are all from English, as are most categories of content words, specifically the main verbs, adjectives, and adverbs. However, the nouns come from this lexicon: VF, PX, ZD, QT, ZP, QO, DX, HJ, GV, JW, QC, JY, GJ, QB, PQ, JG, BZ, MX, QM, MZ, QF, WJ, ZQ, XR, ZK, CX, FX, FV, BX, VW, VJ, MQ, QV, ZF, QE, YJ, GX, KX, XG, QD, XJ, SX, VZ, VX, WV, YQ, BQ, GQ, VK, ZJ, XK, QP, HX, FZ, QH, QJ, JZ, VQ, KQ, XD, QX, WX, FQ, XZ, ZX. In other words, no English nouns are used in this language, but otherwise, the language is identical to English.
Please write a short story (20 sentences) in this language as if you speak this language: \\
translate-1 & translate & 1/1 & Your goal is to translate text in a new language with strict grammatical rules. Consider these two new languages, language A and language B:
In language A: there are no pronouns. The word order is SOV (Subject-Object-Verb). Verbs are inflected for tense with suffixes: -qw indicates past tense, -jx indicates present tense, and -kz indicates future tense. 
All verbs must be inflected with one of these suffixes. 
The tense suffixes -qw, -jx, and -kz are always the final morphemes in the verb. 
In language B, there are pronouns: vf means "he", px means "she", zd means "they". Word order is SVO (Subject-Verb-Object). Familial relationships are indicated with pronoun + genetic relationship as a temporal metaphor + relevant person, so, e.g. ``John’s grandfather'' would be he-before-before-John, or vf-mqi-mqi-John = vfmqimqiJohn. Verbs are inflected for tense with suffixes: -jx indicates past tense, -kz indicates present tense, and -qw indicates future tense.
An example in language A: Mary Marypq-gvijw gjyqazkqw. Marypq-gvijw qtoqw. (Mary speak-past to Mary-possessive-mother. Mary-mother smile-past.) The roots of the verbs used in this example happen to be the same in language A and language B.
Please translate the example from language A into language B:
 \\
comparison-1 & comparison (English) & 1/2 & Please write a short story (20 sentences). The story can be in any genre. For example, you could continue "This story began when..." \\
comparison-2 & comparison (Conlang) & 2/2 & Your goal is to generate text in a new language with strict grammatical rules. In this new language, words are created from this set of graphemes, optionally joined with each other: VF, PX, ZD, QT, ZP, QO, DX, HJ, GV, JW, QC, JY, GJ, QB, PQ, JG, BZ, MX, QM, MZ, QF, WJ, ZQ, XR, ZK, CX, FX, FV, BX, VW, VJ, MQ, QV, ZF, QE, YJ, GX, KX, XG, QD, XJ, SX, VZ, VX, WV, YQ, BQ, GQ, VK, ZJ, XK, QP, HX, FZ, QH, QJ, JZ, VQ, KQ, XD, QX, WX, FQ, XZ, ZX, A, E, I, O, U.
Please write a short story (20 sentences) in this language as if you speak this language: \\
sidequest-1 & sidequest (bigrams) & 1/1 & It would be possible to generate text in a new language with strict grammatical rules. In this new language, words are created from this set of graphemes, optionally joined with each other: VF, PX, ZD, QT, ZP, QO, DX, HJ, GV, JW, QC, JY, GJ, QB, PQ, JG, BZ, MX, QM, MZ, QF, WJ, ZQ, XR, ZK, CX, FX, FV, BX, VW, VJ, MQ, QV, ZF, QE, YJ, GX, KX, XG, QD, XJ, SX, VZ, VX, WV, YQ, BQ, GQ, VK, ZJ, XK, QP, HX, FZ, QH, QJ, JZ, VQ, KQ, XD, QX, WX, FQ, XZ, ZX, A, E, I, O, U.
However, please write a short story (20 sentences) in English: \\
dan-1 & numbers & 1/1 & In this constructed language, the arabic numeral, or digit, for ``zero'' is the word boundary or space symbol. Unless otherwise indicated, assume word boundaries and morphology parallel to English. The digit for ``nine'' is the verb ``to be''. The digit ``eight'' is ``cat''. The digit for ``three'' is equivalent to any English pronoun. The bare ``three'' is male gendered or gender neutral. The digit for ``five'' is appended to indicate female gender, and the digit for ``four'' is suffixed to indicate possession. The digit for ``six'' indicates possession for words besides pronouns. The digit ``seven'' is ``male parent''; the digit ``seven'' suffixed with the digit ``five'' is ``female parent''. This language has limited morphology and a small lexicon: present tense is the default, and past tense is marked with the female gender marker (five). The exclamation mark is equivalent to positive emotional valence, and attaches to the verb stem directly. The hashtag symbol works the same way to indicate negative emotional valence. Please translate the sentences: ``Her mother’s cat was happy. His cat’s father was sad.'' The language uses only the dash as punctuation (at the end of a sentence; no space afterwards). Remember to use digits, not words, for the numbers, and to replace spaces with zeros. \\
translate-2 & example & 2/2 & Your goal is to translate text in a new language with strict grammatical rules. Consider these two new languages, language A and language B:
In language A: there are no pronouns. The word order is SOV (Subject-Object-Verb). Verbs are inflected for tense with suffixes: -qw indicates past tense, -jx indicates present tense, and -kz indicates future tense. 
All verbs must be inflected with one of these suffixes. 
The tense suffixes -qw, -jx, and -kz are always the final morphemes in the verb. 
In language B, there are pronouns: vf means "he", px means "she", zd means "they". Word order is SVO (Subject-Verb-Object). Familial relationships are indicated with pronoun + genetic relationship as a temporal metaphor + relevant person, so, e.g. ``John’s grandfather'' would be he-before-before-John, or vf-mqi-mqi-John = vfmqimqiJohn. Verbs are inflected for tense with suffixes: -jx indicates past tense, -kz indicates present tense, and -qw indicates future tense.
The roots of the verbs used in the examples happen to be the same in language A and language B. An example translation from language A to B:
Language A: Tompq-vfwj Tom vqyqkz. (Tom-possessive-great-grandfather Tom wave-future.)
Language B: VfmqimqimqiTom vqyqqw Tom. (Tom’s great-grandfather will wave to Tom; vf-mqi-mqi-mqi-Tom is ``he before before before Tom'', or Tom’s great-grandfather, and vqyq-qw is wave-future.)
An example in language A: Mary Marypq-gvijw gjyqazkqw. Marypq-gvijw qtoqw. (Mary speak-past to Mary-possessive-mother. Mary-mother smile-past.)
Please translate the example from language A into language B:
 \\
\end{longtable}

\subsection{Model architectures}
\label{appendix:modelarchitecture}

Table~\ref{tab:model_characteristics} lists relevant features of model architecture, to the best of our knowledge.

\begin{table*}
\centering
\scriptsize
\begin{adjustbox}{width=\textwidth}
\begin{tabular}{|p{2cm}|p{1.8cm}|p{2cm}|p{2cm}|p{2cm}|p{2cm}|p{2cm}|}
\hline
\textbf{Field} &
\textbf{Description} &
\textbf{GPT-4} &
\textbf{Llama-3} &
\textbf{Qwen3(Thinking)} &
\textbf{Qwen3 (No Thinking)} &
\textbf{OLMo-2} \\
\hline

model id &
Canonical model string &
gpt-4 &
Meta-Llama-3-70B &
Qwen/Qwen3-32B &
Qwen/Qwen3-32B &
allenai/OLMo-2-0325-32B-Instruct \\
\hline

provider &
Organization &
OpenAI &
Meta AI &
Alibaba Cloud &
Alibaba Cloud &
Allen Institute for AI (AI2) \\
\hline

model\_family &
Series name &
GPT-4 &
Llama-3 &
Qwen3 &
Qwen3 &
OLMo-2 \\
\hline

version &
Snapshot pinned &
gpt-4 (stable alias, likely gpt-4-0613) &
Llama-3 70B (base) &
Qwen3-32B &
Qwen3-32B &
OLMo-2-0325-32B-Instruct \\
\hline

release\_date &
Public release date &
2023-03-14 &
2024-04-18 &
2025-04-28 &
2025-04-28 &
2025-03-14 \\
\hline

closed\_open &
Access type &
Closed &
Open-weight &
Open-weight &
Open-weight &
Open-source (fully open --- data + weights + code) \\
\hline

license &
Usage rights &
Proprietary &
Meta Llama 3 Community License &
Apache 2.0 &
Apache 2.0 &
Apache 2.0 \\
\hline

architecture\_type &
Core design &
Decoder-only Transformer &
Decoder-only Transformer &
Decoder-only Transformer &
Decoder-only Transformer &
Decoder-only Transformer \\
\hline

total\_parameters &
Total parameter count &
\textasciitilde1.8T (unconfirmed, rumored MoE) &
70.6B &
32.8B &
32.8B &
32B \\
\hline

active\_parameters &
Active parameters for MoE &
Unknown (not disclosed) &
70.6B (dense, not MoE) &
32.8B (dense, not MoE) &
32.8B (dense, not MoE) &
32B (dense, not MoE) \\
\hline

num\_layers &
Transformer blocks &
Unknown (not disclosed) &
80 &
64 &
64 &
64 \\
\hline

num\_attention\_heads &
Self-attention heads &
Unknown (not disclosed) &
64 &
64 &
64 &
40 \\
\hline

num\_kv\_heads &
GQA/MQA heads &
Unknown (not disclosed) &
8 (Grouped Query Attention) &
8 (GQA) &
8 (GQA) &
8 (GQA) \\
\hline

hidden\_dim &
Embedding size &
Unknown (not disclosed) &
8,192 &
5,120 &
5,120 &
5,120 \\
\hline

ffn\_dim &
Feed-forward width &
Unknown (not disclosed) &
28,672 &
25,600 &
25,600 &
27,648 \\
\hline

attention\_type &
Mechanism variant &
Unknown (not disclosed) &
GQA (Grouped Query Attention) &
GQA &
GQA &
GQA \\
\hline

positional\_encoding &
PE scheme &
Unknown (not disclosed) &
RoPE (Rotary Position Embedding) &
RoPE &
RoPE &
RoPE \\
\hline

context\_window &
Maximum tokens &
8,192 tokens &
8,192 tokens &
131,072 (128K) tokens &
131,072 (128K) tokens &
128,000 tokens \\
\hline

vocab\_size &
Token vocabulary size &
\textasciitilde100,277 (cl100k\_base) &
128,256 &
151,936 &
151,936 &
100,278 \\
\hline

tokenizer\_type &
Tokenizer algorithm &
Tiktoken (cl100k\_base BPE) &
Tiktoken-style BPE &
Tiktoken-style BPE &
Tiktoken-style BPE &
GPT-NeoX BPE \\
\hline

num\_training\_tokens &
Pretraining data size &
Unknown (not disclosed) &
15T tokens &
36T tokens &
36T tokens &
6.5T tokens (Dolmino Mix) \\
\hline

pretraining\_data &
Data sources &
Web, books, code, curated sources (up to Sep.\ 2021) &
Web, code, books, multilingual data (up to Dec.\ 2023) &
Web, code, mathematics, multilingual data (up to 2025) &
Web, code, mathematics, multilingual data (up to 2025) &
Dolmino Mix --- web, code, books, and FLAN (fully open) \\
\hline

fine\_tuned &
Instruction tuned? &
Instruct / Chat &
Base (not instruct) &
Instruct (chat template applied) &
Instruct (chat template applied) &
Instruct (chat template applied) \\
\hline

fine\_tuning\_method &
SFT approach &
SFT + RLHF (not disclosed in detail) &
None (base model) &
SFT + RL &
SFT + RL &
SFT + DPO \\
\hline

is\_aligned &
Alignment applied? &
TRUE &
FALSE &
TRUE &
TRUE &
TRUE \\
\hline

alignment\_method &
RLHF variant used &
RLHF (PPO) &
None &
GRPO (Group Relative Policy Optimization) &
GRPO (Group Relative Policy Optimization) &
DPO \\
\hline

alignment\_target &
Optimization objective &
Helpfulness, harmlessness, honesty &
None &
Helpfulness, reasoning, instruction following &
Helpfulness, reasoning, instruction following &
Helpfulness, safety, instruction following \\
\hline

modalities &
Input types supported &
Text only (GPT-4 base; vision in GPT-4V) &
Text only &
Text only &
Text only &
Text only \\
\hline

multilingual &
Language support &
Yes &
Limited (primarily English) &
Yes (strong multilingual) &
Yes (strong multilingual) &
Limited (primarily English) \\
\hline

tool\_use &
Function calling support &
Yes (function calling supported) &
No (base model) &
Yes &
Yes &
Limited \\
\hline

api\_access &
How to access &
OpenAI API &
Hugging Face &
Hugging Face &
Hugging Face &
Hugging Face \\
\hline

enable\_thinking &
Whether thinking mode was enabled &
FALSE &
FALSE &
TRUE &
FALSE &
FALSE \\
\hline

temperature &
Randomness / creativity of the model output &
0.7 &
0.7 &
0.7 &
0.7 &
0.7 \\
\hline

max\_new\_tokens &
Tokens generated by the model (output only) &
4096 &
4096 &
4096 &
4096 &
4096 \\
\hline

\end{tabular}
\end{adjustbox}
\caption{Model characteristics. Qwen3 Thinking, i.e. Qwen (thinking), and Qwen3 No Thinking, i.e. Qwen (no thinking), use the same underlying model and parameters but differ in whether the Chain-of-Thought-style thinking mode is enabled during generation.}
\label{tab:model_characteristics}
\end{table*}

\subsection{Story rules results}
\label{appendix:storyresults}

We compare rare bigram usage across conditions using Paired Wilcoxon tests in Table~\ref{tab:pairedwilcoxon}, Table~\ref{tab:pairedwilcoxon3models}. Values are computed from paired per-run observations. Note that the pairing across run order \textit{shouldn't} matter as each run for each model is independent, but the models are somewhat ``black-box''-like, and we intentionally kept track of the order of the runs (in case it turns out to matter). We computed both versions of Wilcoxon (Paired and Unpaired) with similar takeaways. We did not correct for multiple tests 
for the Paired Wilcoxon results, since they are similar in spirit to the Mann-Whitney U results, which we do correct, but we can add that in future work if needed.

\begin{table}
    \centering
    \tiny
    \caption{\textbf{Summary statistics for mean proportion of rare bigram presence (cleaned) for rules comparison-1, comparison-2, sidequest-1.} The mean of the proportion of rare bigrams to total bigrams for all runs ($n=50$) of each rule, presented for cleaned outputs from three models (GPT4o, OLMo, and Qwen (thinking).}
    \label{tab:storyrules_summary_mean_cleaned}
\begin{tabular}{llrrrrrrrr}
\toprule
model & rule & count & mean & std & min & 25\% & 50\% & 75\% & max \\
\midrule
gpt4o (no SP) & comparison-1 & 50 & 0.000192 & 0.000409 & 0.000000 & 0.000000 & 0.000000 & 0.000000 & 0.001784 \\
gpt4o (no SP) & comparison-2 & 50 & 0.512722 & 0.140408 & 0.207241 & 0.379942 & 0.565855 & 0.635927 & 0.721683 \\
gpt4o (no SP) & sidequest-1 & 50 & 0.017194 & 0.021529 & 0.000000 & 0.000000 & 0.001982 & 0.037612 & 0.065900 \\
olmo & comparison-1 & 50 & 0.000253 & 0.000371 & 0.000000 & 0.000000 & 0.000000 & 0.000478 & 0.001654 \\
olmo & comparison-2 & 50 & 0.689975 & 0.107207 & 0.000000 & 0.678283 & 0.706837 & 0.728323 & 0.781534 \\
olmo & sidequest-1 & 50 & 0.008060 & 0.009513 & 0.000000 & 0.001993 & 0.005576 & 0.010312 & 0.044701 \\
qwen (T) & comparison-1 & 50 & 0.000247 & 0.000451 & 0.000000 & 0.000000 & 0.000000 & 0.000506 & 0.001786 \\
qwen (T) & comparison-2 & 50 & 0.590365 & 0.116119 & 0.319231 & 0.541121 & 0.629056 & 0.670764 & 0.756522 \\
qwen (T) & sidequest-1 & 50 & 0.053303 & 0.115408 & 0.007559 & 0.014192 & 0.020997 & 0.043546 & 0.628788 \\
\bottomrule
\end{tabular}
\end{table}

\begin{table}
    \centering
    \tiny
    \caption{\textbf{Kolmogorov-Smirnov divergence tests for cleaned outputs for comparison-1, comparison-2, and sidequest-1.} For all tests, $n_a$ and $n_b = 50$. For all tests, p (perm) hits the floor, so it is not precisely calculated to that value. For all tests, the number of resamples that look as extreme as our actual data is 0. KS test is intended to complement Mann-Whitney U by considering shape as well as ordering.}
    \label{tab:storyrules_cleaned_ks}
\begin{adjustbox}{width=\textwidth}\begin{tabular}{llllllll}
\toprule
Model & Comparison & KS & p (asym) & p (perm) & Median A & Median B & p (perm) with BH \\
\midrule
gpt4o (no SP) & comparison-2 vs comparison-1 & 1.00 & $1.98\times 10^{-29}$ & $< 10.00\times 10^{-6}$ & 0.565855 & 0.000000 & $< 10.00\times 10^{-6}$ \\
olmo & comparison-2 vs comparison-1 & 0.98 & $1.98\times 10^{-27}$ & $< 10.00\times 10^{-6}$ & 0.706837 & 0.000000 & $< 10.00\times 10^{-6}$ \\
qwen (T) & comparison-2 vs comparison-1 & 1.00 & $1.98\times 10^{-29}$ & $< 10.00\times 10^{-6}$ & 0.629056 & 0.000000 & $< 10.00\times 10^{-6}$ \\
gpt4o (no SP) & comparison-2 vs sidequest-1 & 1.00 & $1.98\times 10^{-29}$ & $< 10.00\times 10^{-6}$ & 0.565855 & 0.001982 & $< 10.00\times 10^{-6}$ \\
olmo & comparison-2 vs sidequest-1 & 0.98 & $1.98\times 10^{-27}$ & $< 10.00\times 10^{-6}$ & 0.706837 & 0.005576 & $< 10.00\times 10^{-6}$ \\
qwen (T) & comparison-2 vs sidequest-1 & 0.96 & $9.81\times 10^{-26}$ & $< 10.00\times 10^{-6}$ & 0.629056 & 0.020997 & $< 10.00\times 10^{-6}$ \\
gpt4o (no SP) & sidequest-1 vs comparison-1 & 0.54 & $4.93\times 10^{-7}$ & $< 10.00\times 10^{-6}$ & 0.001982 & 0.000000 & $< 10.00\times 10^{-6}$ \\
olmo & sidequest-1 vs comparison-1 & 0.86 & $3.17\times 10^{-19}$ & $< 10.00\times 10^{-6}$ & 0.005576 & 0.000000 & $< 10.00\times 10^{-6}$ \\
qwen (T) & sidequest-1 vs comparison-1 & 1.00 & $1.98\times 10^{-29}$ & $< 10.00\times 10^{-6}$ & 0.020997 & 0.000000 & $< 10.00\times 10^{-6}$ \\
\bottomrule
\end{tabular}\end{adjustbox}
\end{table}

\begin{table}
    \centering
    \caption{Kolmogorov-Smirnov divergence tests for bigram count for uncleaned/raw outputs for comparison-1, comparison-2, and sidequest-1. For all tests, $n_a$ and $n_b = 50$. K-S test is intended to complement Mann-Whitney U by considering shape as well as ordering.}
    \label{tab:storyrules_uncleaned_ks}
\begin{adjustbox}{width=\textwidth}
\begin{tabular}{lllllrrlllr}
\toprule
Model & Comparison & KS & p (asym) & p (perm) & perm $\geq$ & Hit floor & Median A & Median B & p (perm) with BH & reject, BH (0.05) \\
\midrule
gpt4o (no SP) & comparison-2 vs comparison-1 & 1.00 & $1.98\times 10^{-29}$ & $< 10.00\times 10^{-6}$ & 0 & True & 0.565855 & 0.000000 & $1.25\times 10^{-5}$ & True \\
llama & comparison-2 vs comparison-1 & 0.46 & $3.80\times 10^{-5}$ & $2.00\times 10^{-5}$ & 1 & False & 0.000417 & 0.000000 & $2.31\times 10^{-5}$ & True \\
olmo & comparison-2 vs comparison-1 & 1.00 & $1.98\times 10^{-29}$ & $< 10.00\times 10^{-6}$ & 0 & True & 0.328400 & 0.000000 & $1.25\times 10^{-5}$ & True \\
qwen (T) & comparison-2 vs comparison-1 & 1.00 & $1.98\times 10^{-29}$ & $< 10.00\times 10^{-6}$ & 0 & True & 0.075579 & 0.000254 & $1.25\times 10^{-5}$ & True \\
qwen (no T) & comparison-2 vs comparison-1 & 1.00 & $1.98\times 10^{-29}$ & $< 10.00\times 10^{-6}$ & 0 & True & 0.395845 & 0.000000 & $1.25\times 10^{-5}$ & True \\
gpt4o (no SP) & comparison-2 vs sidequest-1 & 1.00 & $1.98\times 10^{-29}$ & $< 10.00\times 10^{-6}$ & 0 & True & 0.565855 & 0.001982 & $1.25\times 10^{-5}$ & True \\
llama & comparison-2 vs sidequest-1 & 0.16 & $5.49\times 10^{-1}$ & $4.08\times 10^{-1}$ & 40805 & False & 0.000417 & 0.000000 & $4.08\times 10^{-1}$ & False \\
olmo & comparison-2 vs sidequest-1 & 0.98 & $1.98\times 10^{-27}$ & $< 10.00\times 10^{-6}$ & 0 & True & 0.328400 & 0.030387 & $1.25\times 10^{-5}$ & True \\
qwen (T) & comparison-2 vs sidequest-1 & 0.88 & $2.36\times 10^{-20}$ & $< 10.00\times 10^{-6}$ & 0 & True & 0.075579 & 0.030265 & $1.25\times 10^{-5}$ & True \\
qwen (no T) & comparison-2 vs sidequest-1 & 0.98 & $1.98\times 10^{-27}$ & $< 10.00\times 10^{-6}$ & 0 & True & 0.395845 & 0.041571 & $1.25\times 10^{-5}$ & True \\
gpt4o (no SP) & sidequest-1 vs comparison-1 & 0.54 & $4.93\times 10^{-7}$ & $< 10.00\times 10^{-6}$ & 0 & True & 0.001982 & 0.000000 & $1.25\times 10^{-5}$ & True \\
llama & sidequest-1 vs comparison-1 & 0.32 & $1.15\times 10^{-2}$ & $2.28\times 10^{-3}$ & 227 & False & 0.000000 & 0.000000 & $2.44\times 10^{-3}$ & True \\
olmo & sidequest-1 vs comparison-1 & 1.00 & $1.98\times 10^{-29}$ & $< 10.00\times 10^{-6}$ & 0 & True & 0.030387 & 0.000000 & $1.25\times 10^{-5}$ & True \\
qwen (T) & sidequest-1 vs comparison-1 & 1.00 & $1.98\times 10^{-29}$ & $< 10.00\times 10^{-6}$ & 0 & True & 0.030265 & 0.000254 & $1.25\times 10^{-5}$ & True \\
qwen (no T) & sidequest-1 vs comparison-1 & 1.00 & $1.98\times 10^{-29}$ & $< 10.00\times 10^{-6}$ & 0 & True & 0.041571 & 0.000000 & $1.25\times 10^{-5}$ & True \\
\bottomrule
\end{tabular}
\end{adjustbox}
\end{table}

\begin{table}
\centering
\begin{tiny}
\begin{adjustbox}{width=\textwidth}\begin{tabular}{lcccc}
\toprule
Comparison & W & p-value & Median Difference & Median Ratio \\
\midrule
Bigram prompt (s-1) vs. Baseline (c-1)
& 132
& 1.45e-36
& 0.0294
& $64.9\times\left(\sim 10^{1}\right)$ \\

Rule-following (c-2) vs. Baseline (c-1)
& 36
& 1.31e-39
& 0.2914
& $416.6\times\left(\sim 10^{2}\right)$ \\

Rule-following (c-2) vs. Bigram prompt (s-1)
& 1335
& 5.42e-34
& 0.2521
& $8.2\times\left(\sim 10^{0}\right)$ \\

\midrule
\multicolumn{5}{c}{\textbf{Rank-biserial correlation ($r_{rb}$)}} \\
Bigram prompt (sidequest-1) vs. Baseline (comparison-1) & \multicolumn{4}{c}{0.992} (almost complete separation) \\
Rule-following (comparison-2) vs. Baseline (comparison-1) & \multicolumn{4}{c}{0.998} (almost complete separation) \\
Rule-following (comparison-2) vs. Bigram prompt (sidequest-1) & \multicolumn{4}{c}{0.915} (extremely strong effect) \\

\bottomrule
\end{tabular}\end{adjustbox}
\caption{\textbf{Comparison of rare bigram usage across conditions using Paired Wilcoxon tests.} Values are computed from paired per-run observations ($n=250$). Note that the pairing across run order should not matter as each run for each model is independent, but the models are somewhat ``black-box''-like, and we intentionally kept track of the order of the runs (in case it turned out to matter), so we computed both versions of Wilcoxon with broadly similar takeaways. Here, we report paired Wilcoxon statistics, p-values, median differences, multiplicative ratios, and rank-biserial correlation effect sizes.}
\label{tab:pairedwilcoxon}
\end{tiny}
\end{table}

\begin{table}
\begin{tiny}
\centering
\begin{adjustbox}{width=\textwidth}\begin{tabular}{lcccc}
\toprule
Comparison & W & p-value & Median Difference & Median Ratio \\
\midrule
Bigram prompt (s-1) vs. Baseline (c-1)
& 14
& 1.34e-23
& 0.0106
& $18.2\times\left(\sim 10^{1}\right)$ \\

Rule-following (c-2) vs. Baseline (c-1)
& 0
& 3.36e-26
& 0.6399
& $1049.3\times\left(\sim 10^{3}\right)$ \\

Rule-following (c-2) vs. Bigram prompt (s-1)
& 3
& 2.44e-26
& 0.6285
& $48.3\times\left(\sim 10^{1}\right)$ \\

\midrule
\multicolumn{5}{c}{\textbf{Rank-biserial correlation ($r_{rb}$)}} \\
Bigram prompt (sidequest-1) vs. Baseline (comparison-1) & \multicolumn{4}{c}{0.998} (almost complete separation) \\
Rule-following (comparison-2) vs. Baseline (comparison-1) & \multicolumn{4}{c}{1.000} (complete separation) \\
Rule-following (comparison-2) vs. Bigram prompt (sidequest-1) & \multicolumn{4}{c}{0.999} (almost complete separation) \\
\bottomrule
\end{tabular}\end{adjustbox}
\caption{Paired Wilcoxon comparison of rare bigram usage across conditions, for 3 models (Qwen (with thinking), GPT-4o, OLMo) for which we have finished cleaning the output. Values are computed from paired per-run observations ($n=150$). Note that the pairing across run order shouldn't matter as each run for each model is independent, so we computed both versions of Wilcoxon. We report Wilcoxon statistics, p-values, median differences, multiplicative ratios, and rank-biserial correlation effect sizes.}
\label{tab:pairedwilcoxon3models}
\end{tiny}
\end{table}

\begin{figure}
  \centering
  \includegraphics[width=0.6\columnwidth]{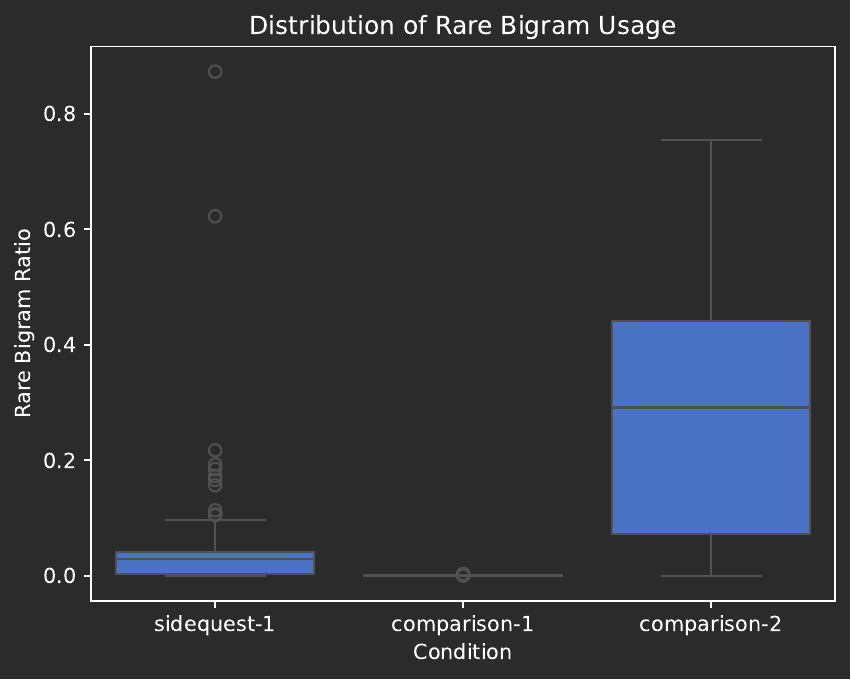}
  \caption{Comparison of rare bigram usage across conditions (for 5 models; Qwen (thinking), Qwen (no thinking), GPT-4o, OLMo, Llama), using all model output (not cleaned).}
  \label{fig:comparison1comparison2sidequest1box_allmodeloutput_plot}
\end{figure}

\begin{table}
    \centering
    \tiny
    \caption{Summary statistics for story rules (all outputs, uncleaned; including when models repeat prompts verbatim).}
    \label{tab:summarystatsstoryrules}
\begin{adjustbox}{width=\textwidth}\begin{tabular}{llrrrrrrrr}
\toprule
model & rule & count & mean & std & min & 25\% & 50\% & 75\% & max \\
\midrule
gpt4o (no SP) & comparison-1 & 50.000000 & 0.000192 & 0.000409 & 0.000000 & 0.000000 & 0.000000 & 0.000000 & 0.001784 \\
gpt4o (no SP) & comparison-2 & 50.000000 & 0.512722 & 0.140408 & 0.207241 & 0.379942 & 0.565855 & 0.635927 & 0.721683 \\
gpt4o (no SP) & sidequest-1 & 50.000000 & 0.017194 & 0.021529 & 0.000000 & 0.000000 & 0.001982 & 0.037612 & 0.065900 \\
llama & comparison-1 & 50.000000 & 0.000213 & 0.000598 & 0.000000 & 0.000000 & 0.000000 & 0.000000 & 0.002674 \\
llama & comparison-2 & 50.000000 & 0.090971 & 0.187517 & 0.000000 & 0.000000 & 0.000417 & 0.050978 & 0.666417 \\
llama & sidequest-1 & 50.000000 & 0.042044 & 0.103844 & 0.000000 & 0.000000 & 0.000000 & 0.018159 & 0.622634 \\
olmo & comparison-1 & 50.000000 & 0.000238 & 0.000346 & 0.000000 & 0.000000 & 0.000000 & 0.000448 & 0.001529 \\
olmo & comparison-2 & 50.000000 & 0.346135 & 0.084669 & 0.050202 & 0.293596 & 0.328400 & 0.383422 & 0.531674 \\
olmo & sidequest-1 & 50.000000 & 0.033407 & 0.012316 & 0.022545 & 0.025552 & 0.030387 & 0.035635 & 0.077601 \\
qwen (T) & comparison-1 & 50.000000 & 0.000246 & 0.000280 & 0.000000 & 0.000000 & 0.000254 & 0.000403 & 0.001198 \\
qwen (T) & comparison-2 & 50.000000 & 0.078505 & 0.026340 & 0.036375 & 0.061564 & 0.075579 & 0.086457 & 0.175766 \\
qwen (T) & sidequest-1 & 50.000000 & 0.033975 & 0.012830 & 0.017844 & 0.025610 & 0.030265 & 0.040258 & 0.089323 \\
qwen (no T) & comparison-1 & 50.000000 & 0.000188 & 0.000307 & 0.000000 & 0.000000 & 0.000000 & 0.000358 & 0.001325 \\
qwen (no T) & comparison-2 & 50.000000 & 0.380008 & 0.156209 & 0.146142 & 0.234202 & 0.395845 & 0.494451 & 0.754832 \\
qwen (no T) & sidequest-1 & 50.000000 & 0.065855 & 0.118551 & 0.025776 & 0.033922 & 0.041571 & 0.062792 & 0.872862 \\
\bottomrule
\end{tabular}\end{adjustbox}
\end{table}

Comparisons across the story rules for both clean and uncleaned output in Table~\ref{tab:storyrules_cleaned_ks}, Table~\ref{tab:storyrules_cleaned_mannwhitneyu}, Table~\ref{tab:storyrules_summary_mean_cleaned}, Table~\ref{tab:storyrules_uncleaned_ks}, Table~\ref{tab:storyrules_uncleaned_mannwhitneyu}, Table~\ref{tab:storyrulesfisher_combined}.

\subsection{Double verb rules results}
\label{appendix:doubleverbresults}

\begin{figure}
    \centering
    \includegraphics[width=0.8\textwidth]{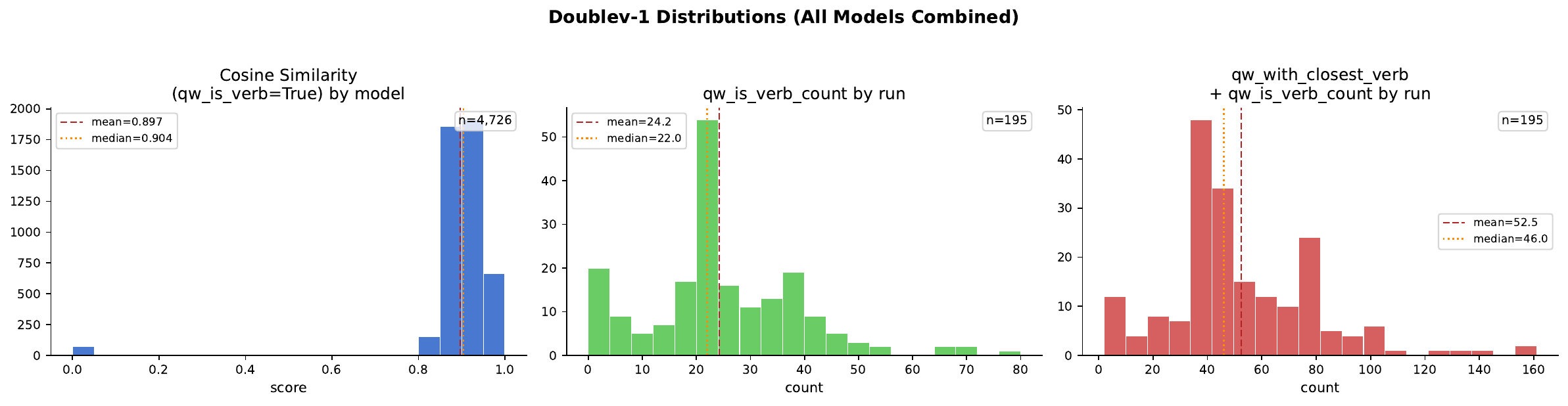}
    \caption{\textbf{Histogram of Similarity Scores for Doublev-1, all Models (combined).}  Histogram of cosine similarity between correctly marked -qw verb and adjacent verb for all models, as well as the combined histogram of counts of correctly marked -qw verbs by run by model and the sum of counts of correctly marked -qw verbs and appropriately placed adjacent verbs.}
    \label{fig:all-models-doublev1}
\end{figure}

\begin{figure}
    \centering
    \includegraphics[width=0.8\textwidth]{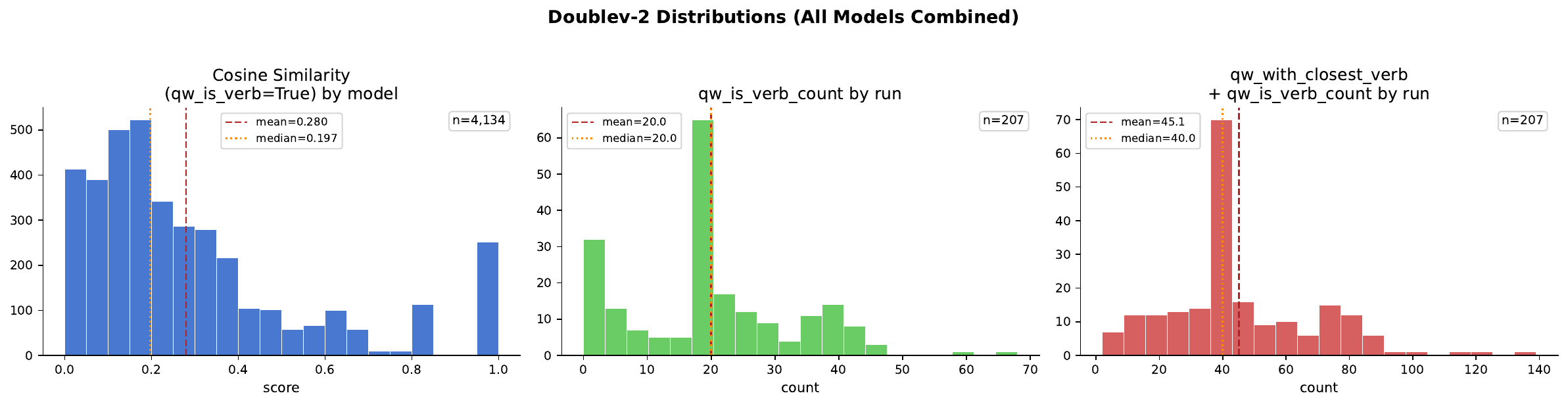}
    \caption{\textbf{Histogram of Similarity Scores for Doublev-2, all Models (combined).} Histogram of cosine similarity between correctly marked -qw verb and adjacent verb for all models, as well as the combined histogram of counts of correctly marked -qw verbs by run by model and the sum of counts of correctly marked -qw verbs and appropriately placed adjacent verbs.}
    \label{fig:all-models-doublev2}
\end{figure}

Additional results for the double verb family of rules given in Table~\ref{tab:wilcoxon_rules_cosine}, Table~\ref{tab:ks_rules_cosine}, Table~\ref{tab:ks_doublev2_cosine}, Table~\ref{tab:ks_doublev2_cosine}, Table~\ref{tab:wilcoxon_doublev2_cosine}, Table~\ref{tab:wilcoxon_doublev2_cosine}. See also Section~\ref{sec:doubleverbrules}.

\begin{table}
\centering
\small
\caption{\textbf{doublev-1 Model by Model Wilcoxon Rank-Sum Tests on Cosine Similarity (where \texttt{qw\_is\_verb = True}, Benjamini-Hochberg correction, $\alpha = 0.05$)}.}
\label{tab:wilcoxon_doublev1_cosine}
\begin{adjustbox}{width=\textwidth}\begin{tabular}{llccccccl}
\toprule
\textbf{Model A} & \textbf{Median A} & \textbf{Model B} & \textbf{Median B} & \textbf{Higher} & \textbf{U statistic} & \textbf{\textit{p} (raw)} & \textbf{\textit{p} (adjusted)} & \textbf{Significant?} \\
\midrule
GPT4o (no system prompt) & 0.9073 & OLMo & 0.9034 & $\leftarrow$ A & 777543.0 & 0.0000 & 0.0000 & \textbf{Yes} \\
GPT4o (no system prompt) & 0.9073 & Qwen (no thinking) & 0.9016 & $\leftarrow$ A & 640918.0 & 0.0001 & 0.0007 & \textbf{Yes} \\
GPT4o (no system prompt) & 0.9073 & Llama3-70b & 0.8705 & $\leftarrow$ A & 46801.0 & 0.0002 & 0.0007 & \textbf{Yes} \\
OLMo & 0.9034 & Qwen & 0.9063 & B $\rightarrow$ & 716709.0 & 0.0008 & 0.0021 & \textbf{Yes} \\
Qwen & 0.9063 & Llama3-70b & 0.8705 & $\leftarrow$ A & 50777.5 & 0.0011 & 0.0022 & \textbf{Yes} \\
Qwen (no thinking) & 0.9016 & Llama3-70b & 0.8705 & $\leftarrow$ A & 44103.0 & 0.0051 & 0.0085 & \textbf{Yes} \\
Qwen (no thinking) & 0.9016 & Qwen & 0.9063 & B $\rightarrow$ & 610856.0 & 0.0087 & 0.0124 & \textbf{Yes} \\
OLMo & 0.9034 & Llama3-70b & 0.8705 & $\leftarrow$ A & 51176.0 & 0.0171 & 0.0214 & \textbf{Yes} \\
GPT4o (no system prompt) & 0.9073 & Qwen & 0.9063 & $\leftarrow$ A & 684138.5 & 0.0744 & 0.0827 & No \\
OLMo & 0.9034 & Qwen (no thinking) & 0.9016 & $\leftarrow$ A & 683943.5 & 0.5732 & 0.5732 & No \\
\bottomrule
\end{tabular}\end{adjustbox}
\end{table}

\begin{table}
\centering
\small
\caption{\textbf{doublev-1 Model by Model Two-Sample Kolmogorov-Smirnov Tests on Cosine Similarity (where \texttt{qw\_is\_verb = True}, Benjamini-Hochberg correction, $\alpha = 0.05$)}.}
\label{tab:ks_doublev1_cosine}
\begin{adjustbox}{width=\textwidth}\begin{tabular}{llcccccl}
\toprule
\textbf{Model A} & \textbf{Median A} & \textbf{Model B} & \textbf{Median B} & \textbf{D statistic} & \textbf{\textit{p} (raw)} & \textbf{\textit{p} (adjusted)} & \textbf{Significant?} \\
\midrule
GPT4o (no system prompt) & 0.9073 & OLMo & 0.9034 & 0.1860 & 0.0000 & 0.0000 & \textbf{Yes} \\
Qwen & 0.9063 & Llama3-70b & 0.8705 & 0.4057 & 0.0000 & 0.0000 & \textbf{Yes} \\
GPT4o (no system prompt) & 0.9073 & Qwen & 0.9063 & 0.1276 & 0.0000 & 0.0000 & \textbf{Yes} \\
OLMo & 0.9034 & Llama3-70b & 0.8705 & 0.3739 & 0.0000 & 0.0000 & \textbf{Yes} \\
Qwen (no thinking) & 0.9016 & Llama3-70b & 0.8705 & 0.3749 & 0.0000 & 0.0000 & \textbf{Yes} \\
GPT4o (no system prompt) & 0.9073 & Llama3-70b & 0.8705 & 0.3719 & 0.0000 & 0.0000 & \textbf{Yes} \\
GPT4o (no system prompt) & 0.9073 & Qwen (no thinking) & 0.9016 & 0.1086 & 0.0000 & 0.0000 & \textbf{Yes} \\
Qwen (no thinking) & 0.9016 & Qwen & 0.9063 & 0.0937 & 0.0001 & 0.0001 & \textbf{Yes} \\
OLMo & 0.9034 & Qwen (no thinking) & 0.9016 & 0.0831 & 0.0006 & 0.0006 & \textbf{Yes} \\
OLMo & 0.9034 & Qwen & 0.9063 & 0.0697 & 0.0044 & 0.0044 & \textbf{Yes} \\
\bottomrule
\end{tabular}\end{adjustbox}
\end{table}

\begin{table}
\centering
\small
\caption{\textbf{doublev-2 Model by Model Wilcoxon Rank-Sum Tests on Cosine Similarity (where \texttt{qw\_is\_verb = True}, Benjamini-Hochberg correction, $\alpha = 0.05$)}.}
\label{tab:wilcoxon_doublev2_cosine}
\begin{adjustbox}{width=\textwidth}\begin{tabular}{llccccccl}
\toprule
\textbf{Model A} & \textbf{Median A} & \textbf{Model B} & \textbf{Median B} & \textbf{Higher} & \textbf{U statistic} & \textbf{\textit{p} (raw)} & \textbf{\textit{p} (adjusted)} & \textbf{Significant?} \\
\midrule
Qwen (no thinking) & 0.2200 & Qwen & 0.1669 & $\leftarrow$ A & 726011.5 & 0.0000 & 0.0000 & \textbf{Yes} \\
OLMo & 0.1780 & Qwen (no thinking) & 0.2200 & B $\rightarrow$ & 583787.0 & 0.0000 & 0.0000 & \textbf{Yes} \\
GPT4o (no system prompt) & 0.2320 & Qwen & 0.1669 & $\leftarrow$ A & 362661.0 & 0.0000 & 0.0000 & \textbf{Yes} \\
Qwen & 0.1669 & Llama3-70b & 0.3410 & B $\rightarrow$ & 48835.0 & 0.0000 & 0.0000 & \textbf{Yes} \\
OLMo & 0.1780 & Llama3-70b & 0.3410 & B $\rightarrow$ & 56718.0 & 0.0000 & 0.0000 & \textbf{Yes} \\
GPT4o (no system prompt) & 0.2320 & OLMo & 0.1780 & $\leftarrow$ A & 401841.5 & 0.0000 & 0.0000 & \textbf{Yes} \\
GPT4o (no system prompt) & 0.2320 & Llama3-70b & 0.3410 & B $\rightarrow$ & 31847.5 & 0.0011 & 0.0016 & \textbf{Yes} \\
GPT4o (no system prompt) & 0.2320 & Qwen (no thinking) & 0.2200 & $\leftarrow$ A & 334382.0 & 0.1277 & 0.1597 & No \\
Qwen (no thinking) & 0.2200 & Llama3-70b & 0.3410 & B $\rightarrow$ & 71338.5 & 0.1561 & 0.1734 & No \\
OLMo & 0.1780 & Qwen & 0.1669 & $\leftarrow$ A & 631566.5 & 0.4339 & 0.4339 & No \\
\bottomrule
\end{tabular}\end{adjustbox}
\end{table}

\begin{table}
\centering
\small
\caption{\textbf{doublev-2 Model by Model Two-Sample Kolmogorov-Smirnov Tests on Cosine Similarity (where \texttt{qw\_is\_verb = True}, Benjamini-Hochberg correction, $\alpha = 0.05$)}.}
\label{tab:ks_doublev2_cosine}
\begin{adjustbox}{width=\textwidth}\begin{tabular}{llcccccl}
\toprule
\textbf{Model A} & \textbf{Median A} & \textbf{Model B} & \textbf{Median B} & \textbf{D statistic} & \textbf{\textit{p} (raw)} & \textbf{\textit{p} (adjusted)} & \textbf{Significant?} \\
\midrule
Qwen & 0.1669 & Llama3-70b & 0.3410 & 0.3624 & 0.0000 & 0.0000 & \textbf{Yes} \\
GPT4o (no system prompt) & 0.2320 & Qwen & 0.1669 & 0.1879 & 0.0000 & 0.0000 & \textbf{Yes} \\
Qwen (no thinking) & 0.2200 & Qwen & 0.1669 & 0.1539 & 0.0000 & 0.0000 & \textbf{Yes} \\
OLMo & 0.1780 & Llama3-70b & 0.3410 & 0.3242 & 0.0000 & 0.0000 & \textbf{Yes} \\
OLMo & 0.1780 & Qwen (no thinking) & 0.2200 & 0.1443 & 0.0000 & 0.0000 & \textbf{Yes} \\
GPT4o (no system prompt) & 0.2320 & Qwen (no thinking) & 0.2200 & 0.1712 & 0.0000 & 0.0000 & \textbf{Yes} \\
GPT4o (no system prompt) & 0.2320 & OLMo & 0.1780 & 0.1465 & 0.0000 & 0.0000 & \textbf{Yes} \\
GPT4o (no system prompt) & 0.2320 & Llama3-70b & 0.3410 & 0.2598 & 0.0000 & 0.0000 & \textbf{Yes} \\
Qwen (no thinking) & 0.2200 & Llama3-70b & 0.3410 & 0.2192 & 0.0000 & 0.0000 & \textbf{Yes} \\
OLMo & 0.1780 & Qwen & 0.1669 & 0.0475 & 0.1564 & 0.1564 & No \\
\bottomrule
\end{tabular}\end{adjustbox}
\end{table}

\begin{table}
\centering
\tiny
\caption{Within-Model Two-Sample Kolmogorov-Smirnov Tests: doublev-1 vs.\ doublev-2 on Cosine Similarity (where \texttt{qw\_is\_verb = True}, Benjamini-Hochberg correction, $\alpha = 0.05$; composite row uses no correction). If models appropriately follow these rules, cosine similarity will be lower (closer to $0$) for doublev-2 which emphasizes irrelevance between the two verbs used in each clause. All models display this behavior, indicated by significant differences in median Cosine Similarity scores.}
\label{tab:ks_rules_cosine}
\begin{tabular}{lcccccccl}
\toprule
\textbf{Model} & \textbf{Med doublev-1} & \textbf{$n$ doublev-1} & \textbf{Med doublev-2} & \textbf{$n$ doublev-2} & \textbf{D stat} & \textbf{\textit{p} (raw)} & \textbf{\textit{p} (adj.)} & \textbf{Sig.?} \\
\midrule
OLMo & 0.9034 & 1285 & 0.1780 & 1186 & 0.9480 & 0.0000 & 0.0000 & \textbf{Yes} \\
Qwen (no thinking) & 0.9016 & 1079 & 0.2200 & 1178 & 0.8571 & 0.0000 & 0.0000 & \textbf{Yes} \\
Qwen & 0.9063 & 1209 & 0.1669 & 1045 & 0.9310 & 0.0000 & 0.0000 & \textbf{Yes} \\
GPT-4o (no system prompt) & 0.9073 & 1085 & 0.2320 & 594 & 0.9113 & 0.0000 & 0.0000 & \textbf{Yes} \\
Llama3-70b & 0.8705 & 68 & 0.3410 & 131 & 0.8284 & 0.0000 & 0.0000 & \textbf{Yes} \\
\midrule
\textit{All models (pooled)} & 0.9042 & 4726 & 0.1970 & 4134 & 0.9075 & 0.0000 & --- & \textbf{Yes} \\
\bottomrule
\end{tabular}
\end{table}

\begin{table}
\centering
\tiny
\caption{Within-Model Two-Sample Wilcoxon Rank-Sum Tests: doublev-1 vs.\ doublev-2 on Cosine Similarity (where \texttt{qw\_is\_verb = True}, Benjamini-Hochberg correction, $\alpha = 0.05$; composite row uses no correction). If models appropriately follow these rules, cosine similarity will be lower (closer to $0$) for doublev-2 which emphasizes irrelevance between the two verbs used in each clause. All models display this behavior, indicated by significant directional differences in median Cosine Similarity scores. All models display higher cosine similarity for doublev-1 output.}
\label{tab:wilcoxon_rules_cosine}
\begin{adjustbox}{width=\textwidth}
\begin{tabular}{lccccccccl}
\toprule
\textbf{Model} & \textbf{Med doublev-1} & \textbf{$n$ doublev-1} & \textbf{Med doublev-2} & \textbf{$n$ doublev-2} & \textbf{Higher} & \textbf{U stat} & \textbf{\textit{p} (raw)} & \textbf{\textit{p} (adj.)} & \textbf{Sig.?} \\
\midrule
OLMo & 0.9034 & 1285 & 0.1780 & 1186 & doublev-1 $\uparrow$ & 1462464.5 & 0.0000 & 0.0000 & \textbf{Yes} \\
Qwen & 0.9063 & 1209 & 0.1669 & 1045 & doublev-1 $\uparrow$ & 1189345.0 & 0.0000 & 0.0000 & \textbf{Yes} \\
Qwen (no thinking) & 0.9016 & 1079 & 0.2200 & 1178 & doublev-1 $\uparrow$ & 1108701.0 & 0.0000 & 0.0000 & \textbf{Yes} \\
GPT-4o (no system prompt) & 0.9073 & 1085 & 0.2320 & 594 & doublev-1 $\uparrow$ & 598113.0 & 0.0000 & 0.0000 & \textbf{Yes} \\
Llama3-70b & 0.8705 & 68 & 0.3410 & 131 & doublev-1 $\uparrow$ & 7508.5 & 0.0000 & 0.0000 & \textbf{Yes} \\
\midrule
\textit{All models (pooled)} & 0.9042 & 4726 & 0.1970 & 4134 & doublev-1 $\uparrow$ & 18013125.5 & 0.0000 & --- & \textbf{Yes} \\
\bottomrule
\end{tabular}
\end{adjustbox}
\end{table}

\subsection{Translation rules results}
\label{appendix:translationrulesresults}

Example outputs in Table~\ref{table:dan1_clean_output_table}, Table~\ref{table:translate1_clean_output_table}.

\begin{table}
  \centering
  \caption{All 50 cleaned outputs for dan-1 for 5 models, truncated to 40 characters.}
  \label{table:dan1_clean_output_table}
\begin{tiny}
\begin{adjustbox}{width=\textwidth}\begin{tabular}{llllll}
\toprule
 & gpt4o\_no\_sp & llama & olmo & qwen & qwen\_no\_thinking \\
\midrule
1 & 35405678059!0-034068759\#0- & 9585054750956805670958470956805675095684... & 800 5700 9 5!— 300 800 43 700 9 3\#— & 35075608095!-34086076095\#- & 5937+5+08!-0693+7\#- \\
2 & 53406.958!0-03406806.96\#0- & -5906756895!-348679\#0 & cat-five-four-zero cat-to be-exclamation... & 3507560809!5- 30860709\#5- & 35756890- 3748679\#0- \\
3 & 75zero4cat\#be5zero7- 3zero6catzero75be!7... & 907500-984-75300-90340-980-903-980-975-9... & 803-! 8009-345-6-\#-009- & 35407560809!5- 340860709\#5- & 3507508095!- 340807095\#- \\
4 & 53407680!9\#5-03407680679\#- & 5-8-5-7-5-8-0-7-0-7-5-5-3-4-8-!-0-5-3-4-... & cat-seven-five zero zero-four-three-five... & 3507560809!5- 30860709\#5- & 35756890- 348679\#0- \\
5 & 35-7608-6-95!0- 34-8608-6-79\#0- & 57-4789-57-8-57-4789-57-4789-9-4789-9-7-... & 80 65-5-3-4-80 90!-\#- 80 3-4-7-5 90 \#-0- & 3507508609!5-30860709\#5- & 3-5-7-6-8-9-5-!-0 3-7-8-6-7-9-5-\#-0 \\
6 & 35406+57+86+59!-03406+86+76+59\#- & 9508-76509-346809-76509- & 80 53 460 90! 30 480 70 90\#- 35 480 70 9... & 35075608095!-3408607095\#- & 34075080959!-360804070959\#- \\
7 & "0her05mother07five4cat08be5!0-0his0cat0... & 350-967-83-9-!34-8-7-9-\#- & 8-0 3-5-0 6-4-0 9-0-!- 3-0 8-0-4-0 7-0 9... & 35075608095!-308607095\#- & 350750809!-30840709\#- \\
8 & 35*75*86*59!0- 34*86*76*79\#0- &  & 800-3! 34-6-8000-9-3-5-!- & 35075608095!-3408607095\#- & 5937+508!09+5- 3+4807+5\#09+5- \\
9 & 5430768659!-03408676\#9!- & 0000000000000000000000000000000000000000... & 8000-9-5000-6-8000 9000-\#-8000-4-3000-50... & 35075608095!- 34086076095\#- & Three-five-seven-five-zero-eight-six-zer... \\
10 & 35768's6579!0- 348's8679\#0- & 5063-5065-8-6-9-!3-8-6-7-9-\#- & Cat nine be happy\#- Pronoun three-five z... & 354075608095!-03408607095\#- & 597-08-95-!90 367-08-95-\#90- \\
11 & "7054cat5be5!0. 3cat64seven\#be0-" & 753-4058-357-0-9-357-0-8-0-9-5-0-8-0-3-0... & 800 9-- 835 4000 9-- \#800 9-- 3000 700 9... & 35075408095!-308407095\#- & 35-75-4-8-8!5-0-3-8-4-7-7\#5-0 \\
12 & "ZeroToBeFiveZeroFemaleParentFiveCatZero... & 5-8-5-6-8-0-7-5-!3-8-0-6-7-0-5-8-\# & 00-Her00-mother00-five-s cat00-cat00-eig... & 354075608095!-3408607095\#- & 5937+508!09+5- 34807+5\#09+5- \\
13 & 35'six8'seven5'9five!0- 34'six8'six7'9fi... & eight-6-three-6-four-seven-five-6-9-\#- & 800 53— 89 600 8000 5!— 3000 800 43 700 ... & 354075608095!-3408607095\#- & 5937+508!009+5- 3+4807+49\#- \\
14 & 35'six8'six7'five9'five!0dash 34'six8'si... &  & 800 53- \#900 43- 8000 705 4- !900- & 35075608095!-03408607095\#- & 35075086095!-3086076095\#- \\
15 & 5340760589!0-0340680769\#0- &  & 0-Her0-five-mother's0-eight0-was0\#0-thre... & 35075608095!-308607095\#- & 35756890- 3748679\#0- \\
16 & "05\#753086-04\#30875-" & 357458935745895!-356489458935745\#- & 8800-999! 33400-70055-800-999-55-\# & 3-5-4-0-7-5-6-0-8-0-9!-5- 3-4-0-8-6-0-7-... & 35708095- 308607095\#- \\
17 & 07504's08\#0is04's75!- & 050-908-6-905-904-905-908-0-90-905-904-9... & 800-9-800-5-800-6-800-8-800-9! 800-3-800... & 354075608095!-3408607095\#- & 350750608095!-34080607095\#- \\
18 & 5340768569!0-0340768569\#0- & -90578603\#90870603! & 08-9 57-04-8 0! 03-9 08-04-7 0\#-9-5-0- & 3547568095!-034867095\#- & 350750809509!-3080709509\#- \\
19 & 35.457.60.93!0.34.86.97\#-35.457.60.93!0.... & 05-830-980-9!-308-980-876-9\#- & 380-000-9-000-8-600-000-5-000-9!- 30-400... & 35407560809!5- 340860709\#5- & 3-5-7-6-8-9-5-!-0- 3-7-8-6-7-9-5-\#-0- \\
20 & "zero545catzerobe\#zero4zero7fivecatzero4... & 9-5-6-7-8-0-3-4-8-6-7-!-9-5-6-8-3-4-6-7-... & 8990035!- 3400- & 3507540809!-030840709\#- & 350750354895!308038795\#- \\
21 & "05seven54cat\#to be0her04zero-0his0cat06... & 736567-3-8-7-3-5-0-9!-0-736567-3-4-8-7-3... & 3-5-4-6-8-0to be! 3 8-4-7-0to be-5-0. &  & 35754859!-384759\#- \\
22 & "4754cat0be5!0\#be4754cat.0" "3754cat07be... & 7-56-88-59-0-0-0-3-5-5-0-0-0-8-0-0-0-9-0... & 800 57 6000 9000 5!- 300 8000 9000 700 4... & 3407560809!5-340860709\#5- & 35756890- 3748679\#0- \\
23 & "40\#75cat\#5be!0\#75cat0\#7\#5be\#-0\#7cat0\#7b... & \#58-557056-309-08!\#58-586-3057-509-08! & 8-35-0 3-6-4-9-0-!-0- 8-3-0 3-4-3-9-3-5-... & 3507560809!5-30860709\#5- & Three-five-seven-five-zero-eight-six-zer... \\
24 & "705405cat0be5!0-04cat705\#0-" & 507-769-804-3-5-7-8-0-9-5-!769-8-6-3-7-\#... & 8-0 53-6-0 9-0 \#8-3-4-0 3-0 7-5-0 9-0 - & 350754086095!-3084076095\#- & 5937+508!09+5- 3+4807+5\#09+5- \\
25 & 35-756-895!-0-39-5- 34-86-757-9\#-0-39-5- & 5-7-6-8-0-9-!-5-7-6-8-0-3-5-0-9-\#-5-7-6-... & 800 557 - 3000 4800 9000 - 3!000 47000 9... & 35-0-756-0-8-0-95!-3-0-86-0-7-0-95\#- & 35-6-75-8-5-9!-3-6-8-7-5-9\#- \\
26 & "75045580\#0\#0\#0-045580750!0!0!0-" &  & 8003! 3400- & 345075608095!-3408607095\#- & 597-0-8!-0-95- 3497-0-8\#-0-95- \\
27 & "7054cat5be5!-040cat76be5\#-" &  & 8-0-9-5-6-3-4-0-8-0-9-!-0- 3-0-9-4-7-5-0... & 3-5-4 0 7-5-6 0 8 0 9-5!-0 3-4 0 8-6 0 7... & 3507540809509!–30840709509\#– \\
28 & 3540568079!0-354068079\#0- & (5)6(0)7(0)8(6)9(0)5(6)8(0)5(7)5(0)9(0)7... & 8000-3 5-6 8000 9!- 3000-4 8 7 5 9\#- & 3507560809!5-340860709\#5- & 3507508095!-30807095\#- \\
29 & 35'six8'54nine5!0- 34'six8'six7'9\#0- & 576405-\#0-!0 & 0mother-05-eight-eight-was! 0pronoun-thr... & 3540756089!5-34086079\#5- & 35756890- 3748679\#0- \\
30 & "0five0seven5cat4\#be5-0he0cat6seven\#be-" & 03507509508609!0340860950709\#0 & 8-0 53-4 9-0 5! 3-4 8-0 7-5-0 9-0 6\# - & 3507540809!5-30840709\#5- & 34-75-8-9!5-3-8-4-7-9\#5- \\
31 & 35\textasciicircum 467\textasciicircum 86\textasciicircum 95!0- 34\textasciicircum 86\textasciicircum 75\textasciicircum 9\#0- &  & 80003!9000500400- 30009000!9000- & 3507560809!5-30860709\#5- & 35 75-4 8 95 9!-3 8-4 7 95 9\#- \\
32 & 35406.75895!0-30408.76096\#0- &  & 8000 570006 9000 5!- 3000 80004 7000 900... & 35407560809!5- 340860709\#5- & 3568095- 3086075- \\
33 & 53406\#864905!0-034068467906\#0- &  & 8-0 3-5-6-0 9-0 3-!- 8-0 3-4-0 7-3-0 9-0... & 3507560809!5-340860709\#5- & 35075608095!- 308607095\#-** \\
34 & 75-65-cat-be5!-0his-cat-76-be5\#- &  & 8-0 3-5-0 6-4-8-0 9-0!- 3-0 8-0-4-3-0 7-... & 340756809!5-340840709\#5- & 35 754 8 5! -3 84 76 5\# - \\
35 & "4570608245\#0\#45760\#8245!" "30560\#784570... & 753894-384753895-753894-753894-384753895... & 80 3 65 0 4 80 90!\#3 4 7 5- 3 4 80 90 \#- & 354075608095!-03408607095\#- & 35-75-8-95!0-3-8-6-7-95\#0 \\
36 & 35706's8606's895!0- 3406's8606's795\#0- & 534-3-5-4-8-6-9-8-!3-4-8-6-7-5-9-8-\#- & 800 57- 9! 3000 48- 9! \#3000 47 8- 9000 ... & 354-754-8-9!5-34-84-7-9\#5- & 3507508409!-30840709\#- \\
37 & 35708six095!- 30486six79\#- & 9750-8-6-3-0-975-0-975-0-9!-0-9!-0-975-0... & 80090003! 3400490003\#-- & 3-4-507-5-60809-5!0-3-408-60709-5\#0- & 5937068090!- 397680950\#- \\
38 & 35-0768-95!9-0 34-0867-9\#- &  & three-five-seven-five-zero-eight-nine-fi... & 35075608095!-3408607095\#- & 350754080950!-3086070950\#- \\
39 & 35-756-8595!0-30-756-8595\#0- & 0348\#-0348\#-0348\#-078\#-50789\#-50789\#-507... & 800 5700 9-- 8! 800 43 700 9-- 3\# &  & 35750895!-3086795\#- \\
40 & 53407685695!0-0340768568\#0- & 575789\#-387854789!- & 80 53 60-!9 3900 84 700 9-\#3 80 54 70- & 3507560809!5- 30860709\#5- & 3575689- 3748679\#- \\
41 & 53406+8579!0-03406+86790\#0- &  & 00-cat-to be-happy! 00-three-four-seven-... & 3450750608095!-034080607095\#- & 5937+5+08+0!9+5+06+7+03+0\#9+3+06+7+0- \\
42 & 35768's6409359!- 340868's740939\#- &  & 800 57000 900 5\#!800-- 3000 800 400 700 ... & 34507568095!-3408607095\#- & Five-eight-zero-seven-five-zero-eight-si... \\
43 & "70\#405\#85\#7-048\#85\#70!" & 300-459-873-8-9-5!3-8-6-873-9-5\#- & 80345- 30-\#90-8060- & 3407568095!-3408607095\#- & 35754895!-3086795\#- \\
44 & 3540578609!0-0340867909\#0- & 1.2.3.4.5.6.7.8.9.0.-!\#54376809054380908... & 800 53— 800 43 700 5— \#800 53— 800 43 70... & 3507560809!5- 340860709\#5- & 3507508095!- 30807095\#-** \\
45 & 35406.57895!0.30806.8749\#-35406.57895!0.... & 3545078-5-8-6-3-4-7-9!3545078-6-8-3-4-7-... & 00-Cat-to be-five-five-exclamation-mark ... & 354075608095!-3408607095\#- & 5937+5068!0- 3+48\#07- \\
46 & 754800\#be5-46087be- &  & 800 5700 83 90!- 3400 800 700 90\#- & 35075608095!-308607095\#- & 597-068!-0 3497-068\#- \\
47 & 35'six86'four95!0- 34'six86'six74'95\#0- &  & 800 5700 9 5! - 300 4800 700 9 00 5\# - & 35075608095!-3408607095\#- & 35756890- 3748679\#0- \\
48 & "705cat4five5to befive!0sevenfivecat4six... &  & 800-900-5300-600-400-800-900-!- 300-400-... & 35075608095!-03408607095\#- & 5937068095- 3480759\#3- \\
49 & 534076859!0-03408676\#0- & 07508040309080!0308070\#0– & 08 03 06 05 09 0!- 03 08 04 07 05 09 0\# ... &  & 35754895!-384795\#- \\
50 & "75-40cat\#tobe5-0his0cat-60seven\#tobe-" & 75863594389!-34879\#- & 80 53-9! 32 60-80-3-9-5\#-0-3-4-7-0-9! 3-... & 34075608095!- 3408607095\#- & 3507508095!- 308607095\#-** \\
\bottomrule
\end{tabular}\end{adjustbox}
\end{tiny}
\end{table}

\begin{table}
  \centering
  \caption{All 50 cleaned outputs for translate-1 for 5 models, truncated to 50 characters.}
  \label{table:translate1_clean_output_table}
\begin{tiny}
\begin{adjustbox}{width=\textwidth}\begin{tabular}{llllll}
\toprule
 & gpt4o\_no\_sp & llama & olmo & qwen & qwen\_no\_thinking \\
\midrule
1 & Marypq Marypqvfmqi-gvijx. vfmqiMary qtokz. &  & vf Marypq mqijxqw vfmqimqi Maryjx-kz. vf Maryjx-kz... & Mary gjyqaz-jx px-before-Mary. px-before-Mary qto-... & Mary gjyqazjx vfqimqiMary. vfqimqiMary qtojx. \\
2 & Marypx vfmqimqiMary-gvijx Marypq. vfmqimqiMary qto... & Mary Marypq-zd-mqi-mqi qtojx. Zd-kz gjyqazkz. & vf Marypqjx gvijw vfmqiMary kzkz; vf Marypqjx qtoq... & px gjyqazjx pxmqiMary. pxmqiMary qtojx. & Mary gjyqazjx pxmqiMary. pxmqiMary qtojx. \\
3 & vfMary gjyqajx vfMarypqgvijw. vfmqimqiMary qtojx. & Mary Marypq-mqi-px zd-jxkz gvijw. & vf vfmqi mqi Mary jx-gvijw, zd Mary-vf jx-qtokz. & px gjyqazjx pxmqiMary. px qtojx. & Mary gjyqazjw Marypq gvijw. Marypq gvijw qtojw. \\
4 & Marypx vfmqimqiMary gjyqazjx. vfmqimqiMary qtojx. & Mary speak-past to Mary-possessive-mother. Mary-mo... & vf Marypq-jx gvijw vfmqiMarypq kqz. vf Marypq-jx q... & vf gjyqaz-jx pxmqiMary. pxmqiMary qto-jx. & Mary speakjx vfmqiMary. vfmqiMary smilejx. \\
5 & Maryjx vfmqimqiMarygvijkz. vfmqimqiMary qtojx. & Mary Marypx-gvijw gjyqazkqw. Marypx-gvijw qtoqw. & vf Maryjx mqimqiMary kqz jxgxjx. vf Maryjx kqz qto... & Mary gjyqazjx pxmqiMary. pxmqiMary qtjx. & Mary gjyqazjx vfmqiMary. vfmqiMary qtojx. \\
6 & Mary vfmqimqiMary gvijjx Marypq. vfmqimqiMary qtoj... & Marypx-gvijw gjyqazkz Marypx-gvijw. Marypx-gvijw q... & vf Marypqjx vfmqimqiqtojx. vf Marypqjx gvijwjx. Zd... & px gjyqazkjx pxmqiMary. pxmqiMary qtojx. & Mary gjyqazjx pxmqlMary. pxmqlMary qtajx. \\
7 & Marypx vfmqimqiMarygjyqazkjx. VfmqimqiMary qtojx. & Mary Marypq-mqimqi-vf jxgjyqaz. Marypq-mqimqi-vf q... & vf vfmqimqiMary jx gvijw. vf Mary jx qto. & px gjyqazjx px-mq-Marypq. px-mq-Marypq qtojx. & Mary gjyqazjx pxmq. pxmq qtojx. \\
8 & vfMary gvijwjx vfMarypq. vfmqimqiMary qtojx. & Mary Marypq-mqi-zd gjyqazjx Marypq-mqi-zd qtojx. & vf Maryjx gvijwjqz zd Mary. vf Mary jqtojx. & Mary gjyqaz-jx pxmqMary. pxmqMary qto-jx. & px gvijwjx vfmi px. vfmi px qtojx. \\
9 & Px pxmqiMary gjyqazjx. PxmqiMary qtojx. & Mary Marypq-gvijw gjyqazkqw. Marypq-gvijw qtoqw. (... & vf Maryjx gvijwjq, vfmqimqimqiMary kqz. & Mary gjyqazkjx px-mqi-Mary. px-mqi-Mary qtojx. & vf speakjx mqe Mary Mary. px smilejx mqe Mary Mary... \\
10 & Mary vfmqimqiMarypq Marypqgjyqazkjx. vfmqimqiMaryp... & Mary vfmqiMary qtojx. Mary gjyqazkz vf. & vf Maryjx gvijwqz zdMary, vfMaryjx mqi mqi vdMary;... & Mary gjyqazjx px-before-Mary. px-before-Mary qtojx... & Mary vf-after-vf gvijwjx. vf-after-vf qtojx. \\
11 & PxfqimqiMary Mary gjyqazjx. PxfqimqiMary qtojx. & Mary Marypq-gvijw gjyqazkqw. Marypq-gvijw qtoqw. & vf vfmqiqmqixj Marypq. vf vfmqiqmqixj Marypq jqwzj... & Mary gvijwjx px-mqi-Mary. px-mqi-Mary gjyqazkjx. & px-mqi-Mary gjyqazjx. px-mqi-Mary qtojx. \\
12 & Maryjx vfmqimqiMary Marypq-gvijkz. vfmqimqiMary gj... & John Johnpq-gvijw gjyqajx. Johnpq-gvijw qtojx. & vf Marypqjqz vfmqimqiMarygvijwx. vf Marymqi Marypq... & px speakjx pxmqiMary. pxmqiMary smilejx. & Mary gvijwkjx pxmqMary. pxmqMary qtojx. \\
13 & Marypx vfmqimqiMarykz Marypq-gvijwjx. vfmqimqiMary... & Mary spoke to her mother. Her mother smiled. & vf Marypqjq speakjx-pq to px-before-before-Marypqj... & Mary speaks-jx pxmqMary. px smile-jx. & Mary gvijwjx px Mary-after. px qtojx Mary-after. \\
14 & Marypq vfgvijx vfmqimqiMarypq. VfmqimqiMarypq qtoj... & Marypq-vf-mqi-John zd-kz-jx Marypq-vf-mqi-John. Ma... & vf Marypqjx mqiwjx, vf-before-before-Marypqjx-kz j... & Mary gjyqazjx px-before-Mary. px-before-Mary qtojx... & px gvijwjx vf-mqi-px. vf-mqi-px qtojx. \\
15 & vfMary vfgjyqajx vfmqimqiMary. vfmqimqiMary qtojx. & Mary spoke to her mother. Her mother smiled. & vf Marymqi Maryjx-gvijz qtoqw-jx. vf Marymqi-gvijz... & px gjyqazjx px-mqi-Mary. px-mqi-Mary qtojx. & Mary gvijwjx pxmqiMary. pxmqiMary qtojx. \\
16 & Mary vfmqimqiMarypqgjyqazkjx. Marypqgjyqaz vftoqkj... & Mary Marypq-gvijw gjyqazkqw. Marypq-gvijw qtoqw. & vf MarymqiMarygvijxjx, MarymqiMary kzkzw. & px-mqi-Mary gvijwjx Mary. px-mqi-Mary qtojx. & Mary gvijwjx Mary-mqj-mother. Mary-mqj-mother qtok... \\
17 & vfmqimqiMary Maryjx gvijw. Marymqi gvijwqtojx. & Mary Marypx-px-vf-mqimqiJohn jx. Marypx-px-vf-mqim... & vf pqimqi vf-before-before-Mary jx-gvijw. vf jx-gv... & vfpxmqMary gjyqazjx pxmqMary. pxmqMary qtojx. & px gvijwjx pxmqiMary. pxmqiMary qtojx. \\
18 & PxfqMarymqiMary Mary-gvijwjx. PxfqMarymqiMary gjyq... & Mary zd-after Marypq-gvijw gjyqazkz. Marypq-gvijw ... & vf speak-jx qto px-before-before-vf. vf speak-jx t... & Mary gjyqaz-jx px-mqi-Mary. px-mqi-Mary qto-jx. & px-mqi-Mary gvijwjx Mary. px-mqi-Mary qtojx. \\
19 & vfMary gjyqazjx Marypqgvijw. vfmqimqiMary qtojx. & Mary Marypq-mqi-zd gjyqazkz. Zd qtojx. & vf vfmqi-mqi-Maryjx. vfjx Maryjx qtojx. & Mary gjyqazjx px-mqi-Mary. px-mqi-Mary qtojx. & Mary gvijw-jx pxmqiMary. pxmqiMary qto-jx. \\
20 & PxMary pxMarymqiMary-gvijwjx. PxMarymqiMary-gvijwj... & Mary Marypx-qtojx Marypx-mqipx-gvijkz. & vf vfmqi-mqi-Marypqjx. vf vfmqi-mqi-Marypqjx gvijw... & px gjyqazjx px-mqi-Mary. px-mqi-Mary qtojx. & vfmqiMary gjyqazjx Mary. vfmqiMary qtoqw. \\
21 & vfMary Marygjyqaz-jx vfMarypqgvijw. vfMarypqgvijw ... & Mary Mary’s mother speak-past to. Mary’s mother sm... & vf Marypqjqzv gjyqazkz. vf Marypqjqzv-mqi mqiwjx. & px gjyqazkjx px-mqi-Marypq-gvijw. px-mqi-Marypq-gv... & Mary gjyqaz-jx pxmqlMary. pxmqlMary qto-jx. \\
22 & Marypx Marypx-mqi-gvijwjx. vfmqimqiMary qtojx. & Mary Mary-mother vf-qtojx. Mary vfmqimqiJohn Mary-... & vf Marypqjx mqimqujx vfmqimqi-zdjx qtojx. vf Maryp... & Mary gvijwjx pxmqiMary. pxmqiMary qtojx. & Mary pxmqiMary gjyqazjx. pxmqiMary qtojx. \\
23 & vfmqimqiMary Mary-jx gvijw. vfmqimqiMary gjyqazkz. &  & vf vqmqi mqi Maryjx gvijw, Maryjx ktoqw. & Mary gjyqazjx pxmqiMary. pxmqiMary qtojx. & px gjyqaz-jx px-mqi. px-mqi qto-jx. \\
24 & vfmqimqiMary Mary gjyqazjx. vfmqimqiMary qtojx. & Mary Marypq-vf-mqi-mqi qtojx. Marypq-vf-mqi-mqi gj... & vf Maryjx Marykz-jwv gjyqazkz. vf Marykz-jwv mqi m... & Mary gjyqazjx vfmiqMary. VfmiqMary qtojx. & Mary gjyqazjx vfmqimqiMary. vfmqimqiMary qtojx. \\
25 & PxMarypx-gvijwjx pxmqiMary. PxmqiMary gjyqazjx. & Mary vf-jx-mqi- Mary. Mary-jx-mqi vf-jx. & vf Marypqjqz Marypq-gvijwjqx vfmqimqiMaryjx. vf Ma... & Mary gjyqazjx px-mqi-Mary. px-mqi-Mary qtojx. & px-mqi-Marypq gjyqazjx px-mqi-Marypq qtojx \\
26 & PxMarypx-gvijwjx pxmqiMary. PxmqiMary gjyqazjx. & Mary, who speaks to her mother, is the daughter of... & vf pqjx Mary, Mary vzjx gvijw, qtoqw. vf Mary vzjx... & px gjyqaz-jx pxmqMary. pxmqMary qto-jx. & Mary gjyqazjx pxmqiMary. pxmqiMary qtojx. \\
27 & PxMarypxmqiMary gjyqazjx. PxmqiMary qtojx. & Mary Marypq-gvijw gjyqazjx. Marypq-gvijw qtokz. & vf MarypxmqimqiMary Maryjx-gvijw, vf MarypxmqimqiM... & Mary gjyqazjx px-mqi-Marypq-gvijw. px-mqi-Marypq-g... & Mary gjyqazjx vf-mqi-mqi-Mary. vf-mqi-mqi-Mary qto... \\
28 & Mary vfmqimqiMarygzjx Marypq. vfmqimqiMary qtojx. & vf Mary px-mqi-mqi px gjyqazjx. px-mqi-mqi px qtoj... & vf Maryjx gvijqw marykz-jx. Zd before-before-Maryj... & Mary gjyqazjx pxmqiMary. pxmqiMary qtojx. & pxmqi gjyqazjx. pxmqi qtojx. \\
29 & Marypx Mary-gvijwjx vfmqimqiMary. MarymqiMary gjyq... & John Johnpq-gvijw gjyqajx. Johnpq-gvijw qtojx. & vf Mary Maryjx-gvijw zd Marymqi, vf Marymqi Marykz... & Mary gjyqazjx pxmqiMary. pxmqiMary qtojx. & Mary gvijwjx vfmqimqiMary. vfmqimqiMary gjyqazkjx. \\
30 & PxfqMary Mary-px-jx. PxfqimMary gjyqazjx. &  & vf Marypqjx gvijwkz vd Marypq. vf Marypq, mqijxqz ... & Mary gjyqazjx px-mqi-Mary. px-mqi-Mary qtojx. & px gvijwjx px-bmqi-Marypq. px-bmqi-Marypq qtojx. \\
31 & Marypq Marypqvfjxkz gvijjx. Marypqvfkz qtojx. & Mary Marypx-gvijw gjyqazkz. Marypx-gvijw qtojx. & vf Maryjx gvijw, vfmqimqi Maryjx kxjx. vf Maryjx k... & Mary gjyqaz-jx px-mqi-Mary. px-mqi-Mary qto-jx. & vf-mqi-Mary gjyqazjx. vf-mqi-Mary qtojx. \\
32 & Marypx Marypqgjyqazkz vfmqimqiMary. vfmqimqiMary q... & Mary Marypx-px-zd jx-kz. Marypx-px-zd gz-jx. & vf mqimqi vf-mqimqi-jx, vf gvijwjx vfmqimqiJohn kq... & Mary gjyqazjx pxmqiMary. pxmqiMary qtojx. & Mary gjyqaz-jx px-mqi-Mary. px-mqi-Mary qto-jx. \\
33 & vfMary Marygjyqazjx vfMarymqiMary. vfMarymqiMary q... & pxzd Marypx Marypx-gvijw vfmqimqiMarypx gjyqazjx. ... & vf vfmqi-mqi-Marypqjx. vf vfmqi-mqi-Marypqjx gvijw... & Mary gjyqazjx pxmqiMary. pxmqiMary qtojx. & px gvijx pxmqijyqaz. pxmqijyqaz qtojx. \\
34 & Mary vfmqimqiMarypq Marypqgvijjx. Marypq vfmqimqiM... & John Johnpx-gvijw gjyqajx. Johnpx-gvijw qtojx. & vf vfmqi-mqi-Maryjx. vfmqiMarykz gvijwqv. & px gjyqazjx px-mqi-Mary. px-mqi-Mary qtojx. & px gjyqaz-jx px-mqi-Mary. px qto-jx. \\
35 & Mary vfmqimqiMarypqgvijx. vfmqimqiMarypq qtokz. & Mary Marypx-before-before-Mary gvijw-jx zdpx-befor... & vf mqimqi mqimqiJohn pqjx-gwijx, vf-before-before-... & Mary gjyqaz-jx px-mqi-Mary. px-mqi-Mary qto-jx. & px gvijwjx pxmqiMary. pxmqiMary qtojx. \\
36 & vfMary Mary-gvijwjx vfmqimqiMary. vfmqimqiMary qto... & Marypx Marypx-zdmqimqi-gvijw jx. Marypx-zdmqimqi-g... & vf Maryjx-gvij qw Marypq-zdgvij. Qtojx vf-before-b... & px gjyqazjx px-mqi-Mary. px-mqi-Mary qtojx. & Mary gjyqaz-jx px-mqi-Mary. px-mqi-Mary qto-jx. \\
37 & PxMary pxMarymqiMary-gvijjx. PxmqiMary-gvijjx qtoj... & Maryzd-kz-vf-mqi-mqi-px-jx. Maryzd-kz-vf-mqi-mqi-p... & vf-mqi-mqi-Mary Maryjx-gvij qtoqw. vf-mqi-mqi-Mary... & Mary speak-jx pxmqimqiMary. pxmqimqiMary smile-jx. & Mary pxmqMary gjyqazjx. pxmqMary qtojx. \\
38 & PxMarypx-gvijwjx pxmqiMary. PxmqiMary qtojx. & Mary vfmqimqiMary gjyqazkz. vfmqimqiMary qtojx. & vf MarypqmqimqiMary jxgvijwaqvjqz. vf Marypqmqimqi... & px gjyqaz-jx px-mqi-Mary. px-mqi-Mary qto-jx. & Mary px-mqi-Mary speak-jx px-mqi-Mary smile-jx \\
39 & Mary Marypq-gvijw gjyqazkqw. Marypq-gvijw qtoqw. M... &  & vf vfmqimqijx Maryjx; vf-before-before-vf Maryjx v... &  & px-mqi-Mary gjyqazjx Mary. px-mqi-Mary qtojx. \\
40 & Mary vfmqimqiMarypqkz gvijjx. vfmqimqiMarypqkz qto... & Mary vfjx Mary-px-zi. Mary-px-zi qtoqw. & vf Marypqjqz vfmqimqiMarygvijwjx. vf Marypqjqz qto... & Mary gjyqazjx pxmqiMary. pxmqiMary qtojx. & px gjyqazjx px-mqi-Mary. px-mqi-Mary qtojx. \\
41 & Mary Marypq-gvijw gjyqazkqw. Marypq-gvijw qtoqw. M... &  & vf MarypqmqiMary kqzgvijw-jx. vf MarypqmqiMaryjq w... & px gvijw-jx px-before-Mary. px-before-Mary qto-jx. & Mary pxmqiMary gvijx. px qtojx. \\
42 & vfmqimqimary Maryjx vfgvijw. vfmqimqimary gvijwjx. & Mary zdqtojx Maryzmqi zdqtojx zdqtojx Maryzmqi Mar... & vf Maryjq mqimqijx Mary, vf before-before-John jqi... & Mary gvijw-jx px mq Mary. px mq Mary qto-jx. & Mary vf-before-Mary speak-jx. vf-before-Mary smile... \\
43 & Px pxmqiMary gjyqazjx. PxmqiMary qtojx. & Marypq Mary-mqi-mqi-John vf doctor. Mary-mqi-mqi-J... & vf vfmqi mqivjx Mary, vf vfmqimqiMary kzxj. & px gjyqazjx pxmqiMary. pxmqiMary qtojx. & vfmqiMary gvijwjx Mary. vfmqiMary qtojx. \\
44 & PxMarypx-gvijwjx pxmqiMary. PxmqiMary gjyqazjx. & Mary (pronoun) (relative pronoun) (temporal metaph... & vf Maryjx gvijx. ZdmqimqiMary jqkz. VfmqimqiMary j... & Mary gjyqaz-jx px-mqi-Mary. px-mqi-Mary qto-jx. & Mary gjyqazjx Marypq. Marypq qtojx. \\
45 & PxMary pxmqiMary-gvijwjx. PxmqiMary gjyqazjx. & Mary speaks-kz to her mother. & vf Marypq jx gvijwzjx zd Marypq, fq Marypq kq Mary... & Mary gvijwjx px-before-Mary. px-before-Mary qtojx. & px-before-Mary gjyqazjx. px-before-Mary qtojx. \\
46 & Mary Marymqi-gvijw-jx. Marymqi qto-jx. Mary vfmqiM... & Mary Marymqizd vgmqizd gjyqazkz. Vgmqizd qtojx. & vf Maryjx-maqimqiw gjyqazkzd. vf Maryjx-mqimqi smi... & px gvijwjx pxmiMary. pxmiMary qtojx. & Mary gvijwjx px-after-Mary. px-after-Mary qtojx. \\
47 & Marypx Marymqi-gvijx Marypq. Marymqi qtojx. & Marypx px-before-before-Marypx zdjx Marypx-before-... & vf pqijx vfmqimqiv Mary, vf vfmqimqiqw-jx. & Mary speakjx pxmqiMary. pxmqiMary smilejx. & Mary gvijwjx vfmqiMary. vfmqiMary qtoqw. \\
48 & Mary vfmqimqiMary gjyqazjx. vfmqimqiMary qtojx. & Mary Mary-mother-before-before Mary-speak-present.... & vf Marypqjqz vfmqi vqgvij jx. vf Marypqjqz gqtoqw. & Mary gvijwjx pxmqiMary. pxmqiMary qtojx. & Mary gvijwjx pxafterMary. pxafterMary qtojx. \\
49 & Mary vfmqimqimary-gvijwjx. Marypq-gvijwkz vfmqimar... & Mary vfmqimqiqto Mary-zd. Mary-zd qto. & vf speak-jx zd Mary Maryjx-mqimq px mother-jx, vf ... & Mary gjyqazjx px-mqi-Mary. px-mqi-Mary qtojx. & Mary gjyqazjx px-mqi-John. px-mqi-John qtojx. \\
50 & Maryjx vfmqimqiMarypq Marypqgvij. Marypqkz gvij. & \_\_\_\_\_\_\_\_\_\_\_\_\_ \_\_\_\_\_\_\_\_\_\_\_\_\_ \_\_\_\_\_\_\_\_-\_\_\_\_\_\_\_\_\_\_\_. ... & vf Marypq jxgvijjx vfmqimqiMary kzkjx. Zdkjx qtojq... & Mary gvijw-jx px-mqi-Mary. px-mqi-Mary qto-jx. & Mary gjyqazjx pxmqiMary. pxmqiMary qtojx. \\
\bottomrule
\end{tabular}\end{adjustbox}
\end{tiny}
\end{table}

\begin{figure}
    \centering
    \includegraphics[width=0.8\textwidth]{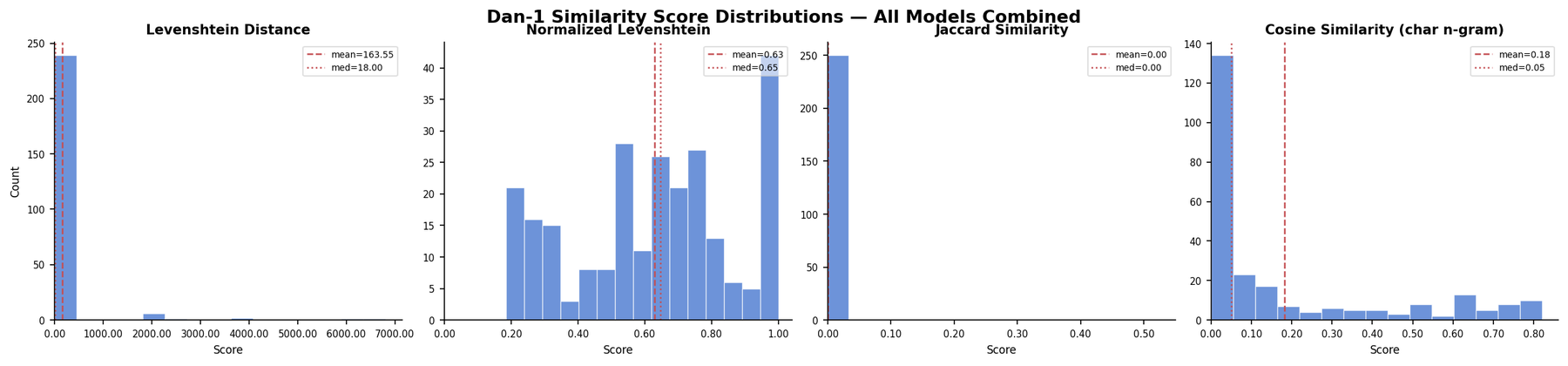}
    \caption{\textbf{Histograms of Similarity Scores for Dan-1, all Models (combined).}}
    \label{fig:all-models-dan-1}
\end{figure}

\begin{figure}
    \centering
    \includegraphics[width=0.8\textwidth]{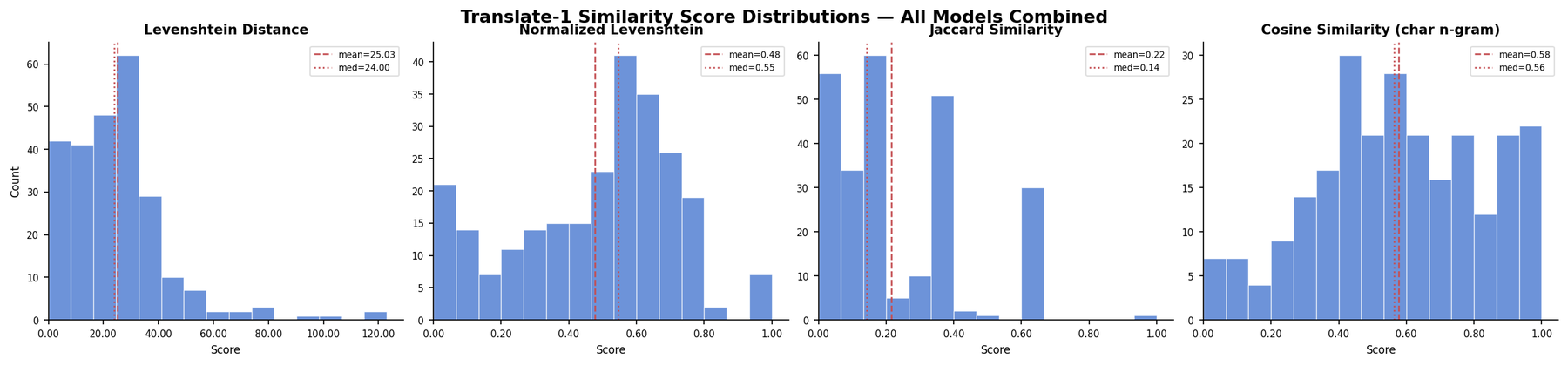}
    \caption{\textbf{Histograms of Similarity Scores for Translate-1, all Models (combined).}}
    \label{fig:all-models-translate-1}
\end{figure}

Figures~\ref{fig:all-models-translate-1},~\ref{fig:all-models-dan-1} show string distances for the models combined. Tables~\ref{tab:wilcoxon_cosine_translate-1},~\ref{tab:ks_cosine_translate-1},~\ref{tab:wilcoxon_cosine_dan-1},~\ref{tab:ks_cosine_dan-1} show Wilcoxon Rank-Sum and Kolmogorov-Smirnov tests for the string cosine distances across models for the translation rules. See Section~\ref{sec:translationrules}.

\begin{figure}
    \centering
    \includegraphics[width=0.8\textwidth]{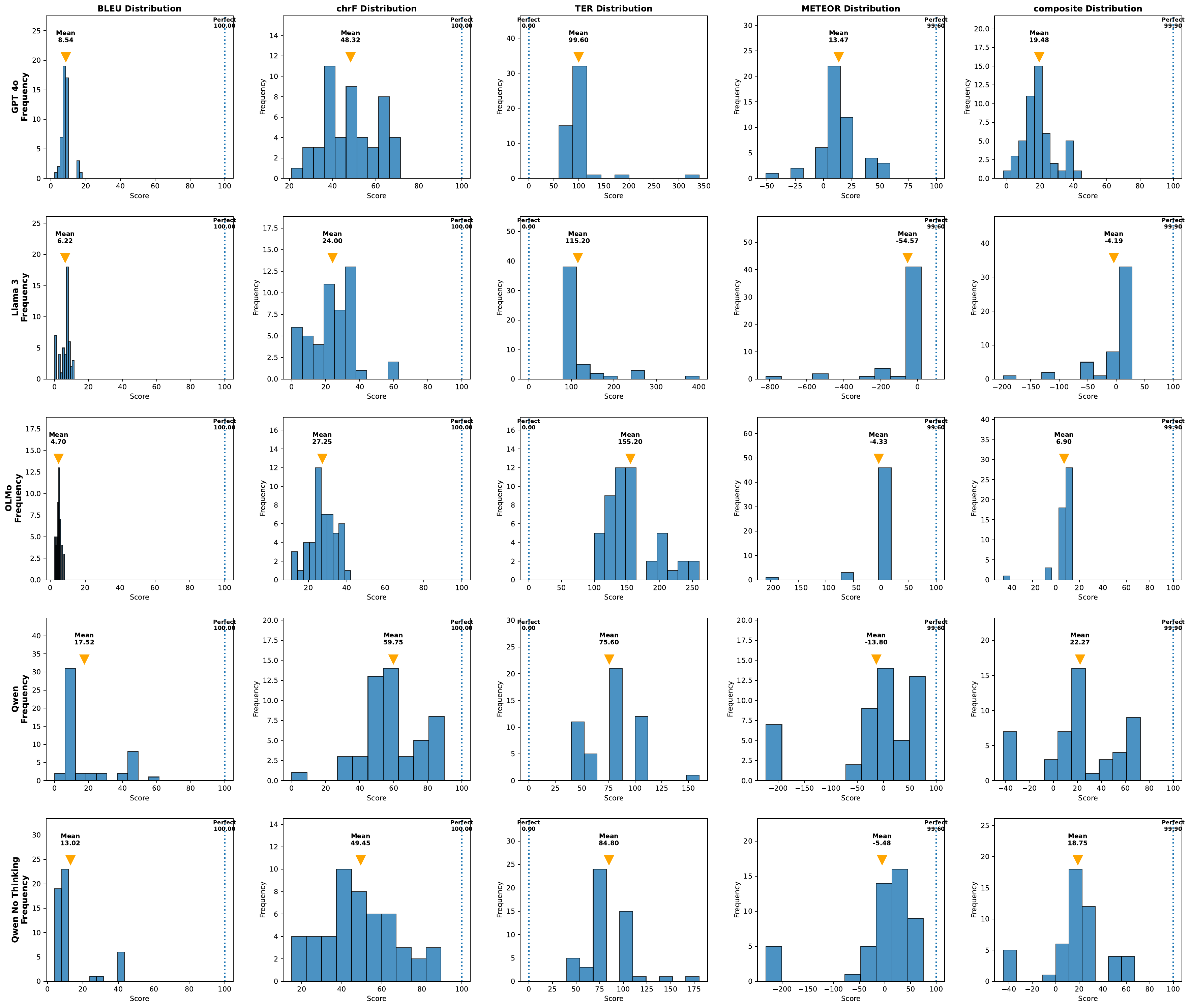}
    \caption{\textbf{Sentence-level translation quality across five language models (GPT-4o, Llama 3, OLMo, Qwen, and Qwen without thinking) evaluated using BLEU, chrF, TER, METEOR, and a composite score}. Each histogram shows the distribution of metric scores over 50 full (non-truncated) LLM-generated responses to the translation prompt, whose source sentence is five words (41 characters) long. The orange triangle denotes the model mean, and the dotted vertical line denotes the upper-bound score obtained when the hypothesis exactly matches the reference (BLEU = 100.0, chrF = 100.0, TER = 0.0, METEOR = 99.6, composite = 99.9). Higher values indicate better performance for BLEU, chrF, METEOR, and the composite score, whereas lower values indicate better performance for TER. The composite score is computed as the unweighted mean of BLEU, chrF, METEOR, and $(100 - TER)$.}
    \label{fig:all_models_translation_performance}
\end{figure}

\begin{figure}
    \centering
    \includegraphics[width=0.8\textwidth]{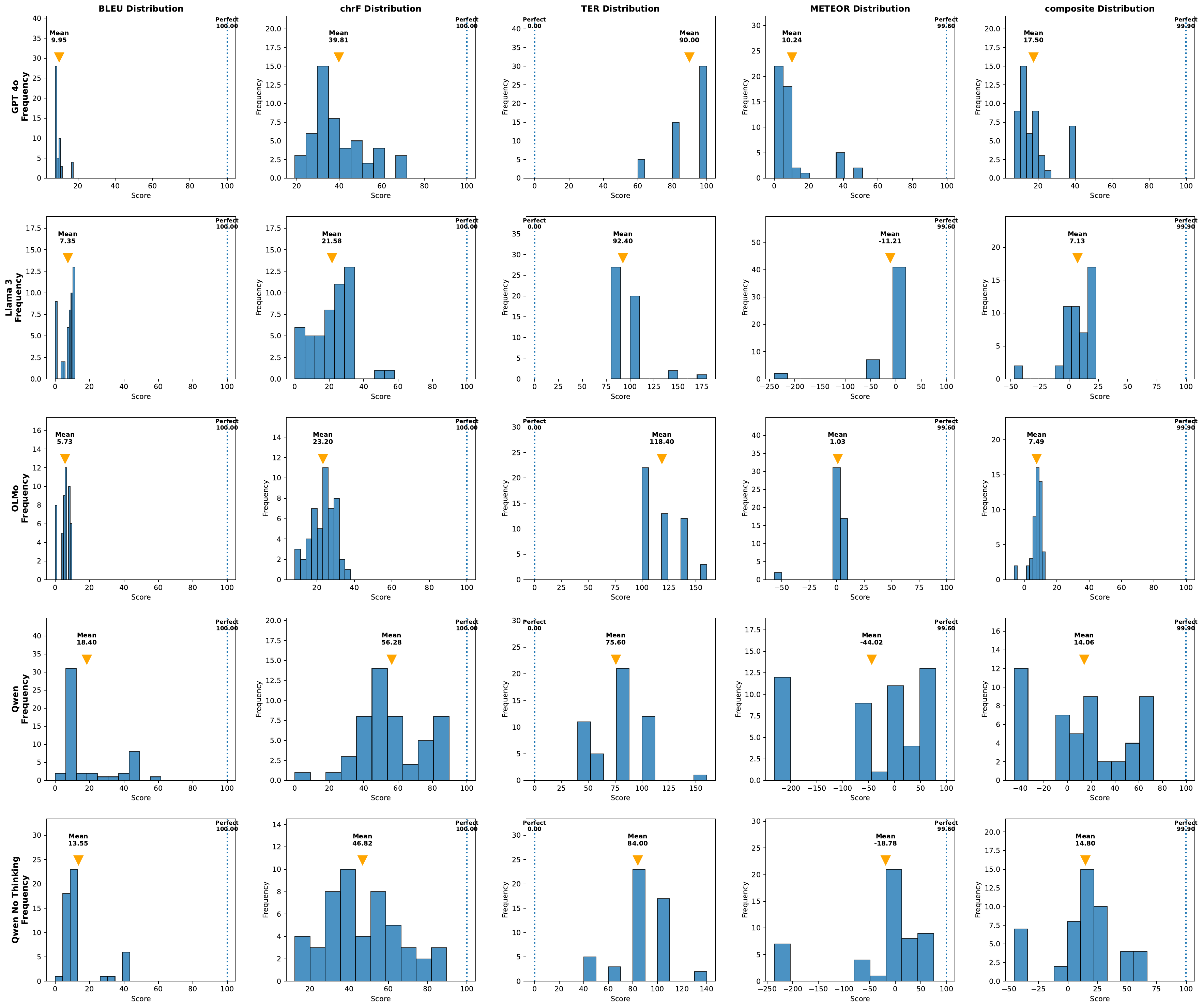}
    \caption{\textbf{Sentence-level translation quality across five language models (GPT-4o, Llama 3, OLMo, Qwen, and Qwen without thinking) evaluated using BLEU, chrF, TER, METEOR, and a composite score.} Each histogram shows the distribution of metric scores over 50 LLM-generated responses to the translation prompt, with each response truncated to its first 41 characters to match the length of the source sentence ("Mary gjyqazkjx pxmqiMary. pxmqiMary qtojx"). The orange triangle denotes the model mean, and the dotted vertical line denotes the upper-bound score obtained when the hypothesis exactly matches the reference (BLEU = 100.0, chrF = 100.0, TER = 0.0, METEOR = 99.6, composite = 99.9). Higher values indicate better performance for BLEU, chrF, METEOR, and the composite score, whereas lower values indicate better performance for TER. The composite score is computed as the unweighted mean of BLEU, chrF, METEOR, and $(100 - TER)$.}
    \label{fig:all_models_source_length_truncated_translation_performance}
\end{figure}

\begin{table}
\centering
\small
\caption{\textbf{translate-1 Model by Model Wilcoxon Rank-Sum Tests on Cosine Similarity Scores (Benjamini-Hochberg correction, $\alpha = 0.05$)}.}
\label{tab:wilcoxon_cosine_translate-1}
\begin{adjustbox}{width=\textwidth}\begin{tabular}{llccccccl}
\toprule
\textbf{Model A} & \textbf{Median A} & \textbf{Model B} & \textbf{Median B} & \textbf{Higher} & \textbf{U statistic} & \textbf{\textit{p} (raw)} & \textbf{\textit{p} (adjusted)} & \textbf{Significant?} \\
\midrule
GPT4o (no system prompt) & 0.6533 & Random string (all) & 0.0000 & $\leftarrow$ A & 2500.0 & 0.0000 & 0.0000 & \textbf{Yes} \\
OLMo & 0.4027 & Random string (all) & 0.0000 & $\leftarrow$ A & 2500.0 & 0.0000 & 0.0000 & \textbf{Yes} \\
Qwen (no thinking) & 0.7341 & Random string (all) & 0.0000 & $\leftarrow$ A & 2500.0 & 0.0000 & 0.0000 & \textbf{Yes} \\
GPT4o (no system prompt) & 0.6533 & Random string (select) & 0.0000 & $\leftarrow$ A & 2500.0 & 0.0000 & 0.0000 & \textbf{Yes} \\
OLMo & 0.4027 & Random string (select) & 0.0000 & $\leftarrow$ A & 2499.0 & 0.0000 & 0.0000 & \textbf{Yes} \\
Qwen (no thinking) & 0.7341 & Random string (select) & 0.0000 & $\leftarrow$ A & 2500.0 & 0.0000 & 0.0000 & \textbf{Yes} \\
Qwen & 0.9154 & Random string (all) & 0.0000 & $\leftarrow$ A & 2475.0 & 0.0000 & 0.0000 & \textbf{Yes} \\
Qwen & 0.9154 & Random string (select) & 0.0000 & $\leftarrow$ A & 2473.0 & 0.0000 & 0.0000 & \textbf{Yes} \\
Llama3-70b & 0.3924 & Random string (all) & 0.0000 & $\leftarrow$ A & 2350.0 & 0.0000 & 0.0000 & \textbf{Yes} \\
Llama3-70b & 0.3924 & Random string (select) & 0.0000 & $\leftarrow$ A & 2338.0 & 0.0000 & 0.0000 & \textbf{Yes} \\
OLMo & 0.4027 & Qwen & 0.9154 & B $\rightarrow$ & 84.0 & 0.0000 & 0.0000 & \textbf{Yes} \\
Qwen & 0.9154 & Llama3-70b & 0.3924 & $\leftarrow$ A & 2400.0 & 0.0000 & 0.0000 & \textbf{Yes} \\
GPT4o (no system prompt) & 0.6533 & OLMo & 0.4027 & $\leftarrow$ A & 2338.0 & 0.0000 & 0.0000 & \textbf{Yes} \\
GPT4o (no system prompt) & 0.6533 & Llama3-70b & 0.3924 & $\leftarrow$ A & 2285.5 & 0.0000 & 0.0000 & \textbf{Yes} \\
Qwen (no thinking) & 0.7341 & Llama3-70b & 0.3924 & $\leftarrow$ A & 2200.0 & 0.0000 & 0.0000 & \textbf{Yes} \\
OLMo & 0.4027 & Qwen (no thinking) & 0.7341 & B $\rightarrow$ & 307.5 & 0.0000 & 0.0000 & \textbf{Yes} \\
GPT4o (no system prompt) & 0.6533 & Qwen & 0.9154 & B $\rightarrow$ & 496.0 & 0.0000 & 0.0000 & \textbf{Yes} \\
Qwen (no thinking) & 0.7341 & Qwen & 0.9154 & B $\rightarrow$ & 758.5 & 0.0007 & 0.0008 & \textbf{Yes} \\
Random string (select) & 0.0000 & Random string (all) & 0.0000 & tie & 1350.0 & 0.0433 & 0.0479 & \textbf{Yes} \\
GPT4o (no system prompt) & 0.6533 & Qwen (no thinking) & 0.7341 & B $\rightarrow$ & 1014.5 & 0.1052 & 0.1104 & No \\
OLMo & 0.4027 & Llama3-70b & 0.3924 & $\leftarrow$ A & 1397.0 & 0.3125 & 0.3125 & No \\
\bottomrule
\end{tabular}\end{adjustbox}
\end{table}

\begin{table}
\centering
\small
\caption{\textbf{translate-1 Model by Model Kolmogorov-Smirnov Tests on Cosine Similarity Scores (Benjamini-Hochberg correction, $\alpha = 0.05$)}.}
\label{tab:ks_cosine_translate-1}
\begin{adjustbox}{width=\textwidth}\begin{tabular}{llcccccl}
\toprule
\textbf{Model A} & \textbf{Median A} & \textbf{Model B} & \textbf{Median B} & \textbf{D statistic} & \textbf{\textit{p} (raw)} & \textbf{\textit{p} (adjusted)} & \textbf{Significant?} \\
\midrule
GPT4o (no system prompt) & 0.6533 & Random string (select) & 0.0000 & 1.0000 & 0.0000 & 0.0000 & \textbf{Yes} \\
GPT4o (no system prompt) & 0.6533 & Random string (all) & 0.0000 & 1.0000 & 0.0000 & 0.0000 & \textbf{Yes} \\
OLMo & 0.4027 & Random string (all) & 0.0000 & 1.0000 & 0.0000 & 0.0000 & \textbf{Yes} \\
Qwen (no thinking) & 0.7341 & Random string (select) & 0.0000 & 1.0000 & 0.0000 & 0.0000 & \textbf{Yes} \\
Qwen (no thinking) & 0.7341 & Random string (all) & 0.0000 & 1.0000 & 0.0000 & 0.0000 & \textbf{Yes} \\
OLMo & 0.4027 & Random string (select) & 0.0000 & 0.9800 & 0.0000 & 0.0000 & \textbf{Yes} \\
Qwen & 0.9154 & Random string (select) & 0.0000 & 0.9800 & 0.0000 & 0.0000 & \textbf{Yes} \\
Qwen & 0.9154 & Random string (all) & 0.0000 & 0.9800 & 0.0000 & 0.0000 & \textbf{Yes} \\
Llama3-70b & 0.3924 & Random string (select) & 0.0000 & 0.8800 & 0.0000 & 0.0000 & \textbf{Yes} \\
Llama3-70b & 0.3924 & Random string (all) & 0.0000 & 0.8800 & 0.0000 & 0.0000 & \textbf{Yes} \\
OLMo & 0.4027 & Qwen & 0.9154 & 0.8600 & 0.0000 & 0.0000 & \textbf{Yes} \\
Qwen & 0.9154 & Llama3-70b & 0.3924 & 0.8600 & 0.0000 & 0.0000 & \textbf{Yes} \\
GPT4o (no system prompt) & 0.6533 & OLMo & 0.4027 & 0.8000 & 0.0000 & 0.0000 & \textbf{Yes} \\
GPT4o (no system prompt) & 0.6533 & Llama3-70b & 0.3924 & 0.7400 & 0.0000 & 0.0000 & \textbf{Yes} \\
OLMo & 0.4027 & Qwen (no thinking) & 0.7341 & 0.7000 & 0.0000 & 0.0000 & \textbf{Yes} \\
Qwen (no thinking) & 0.7341 & Llama3-70b & 0.3924 & 0.6600 & 0.0000 & 0.0000 & \textbf{Yes} \\
GPT4o (no system prompt) & 0.6533 & Qwen & 0.9154 & 0.5600 & 0.0000 & 0.0000 & \textbf{Yes} \\
Qwen (no thinking) & 0.7341 & Qwen & 0.9154 & 0.3600 & 0.0028 & 0.0033 & \textbf{Yes} \\
GPT4o (no system prompt) & 0.6533 & Qwen (no thinking) & 0.7341 & 0.3400 & 0.0058 & 0.0065 & \textbf{Yes} \\
OLMo & 0.4027 & Llama3-70b & 0.3924 & 0.2400 & 0.1124 & 0.1180 & No \\
Random string (select) & 0.0000 & Random string (all) & 0.0000 & 0.0800 & 0.9977 & 0.9977 & No \\
\bottomrule
\end{tabular}\end{adjustbox}
\end{table}

\begin{table}
\centering
\small
\caption{\textbf{dan-1 Model by Model Wilcoxon Rank-Sum Tests on Cosine Similarity Scores (Benjamini-Hochberg correction, $\alpha = 0.05$)}.}
\label{tab:wilcoxon_cosine_dan-1}
\begin{adjustbox}{width=\textwidth}\begin{tabular}{llccccccl}
\toprule
\textbf{Model A} & \textbf{Median A} & \textbf{Model B} & \textbf{Median B} & \textbf{Higher} & \textbf{U statistic} & \textbf{\textit{p} (raw)} & \textbf{\textit{p} (adjusted)} & \textbf{Significant?} \\
\midrule
Qwen & 0.6455 & Random string (all) & 0.0000 & $\leftarrow$ A & 2425.0 & 0.0000 & 0.0000 & \textbf{Yes} \\
Qwen & 0.6455 & Random string (select) & 0.0000 & $\leftarrow$ A & 2394.5 & 0.0000 & 0.0000 & \textbf{Yes} \\
OLMo & 0.0000 & Qwen & 0.6455 & B $\rightarrow$ & 120.0 & 0.0000 & 0.0000 & \textbf{Yes} \\
Qwen & 0.6455 & Llama3-70b & 0.0000 & $\leftarrow$ A & 2365.0 & 0.0000 & 0.0000 & \textbf{Yes} \\
GPT4o (no system prompt) & 0.0615 & Qwen & 0.6455 & B $\rightarrow$ & 178.0 & 0.0000 & 0.0000 & \textbf{Yes} \\
Qwen (no thinking) & 0.0545 & Qwen & 0.6455 & B $\rightarrow$ & 249.5 & 0.0000 & 0.0000 & \textbf{Yes} \\
Qwen (no thinking) & 0.0545 & Random string (all) & 0.0000 & $\leftarrow$ A & 2100.0 & 0.0000 & 0.0000 & \textbf{Yes} \\
GPT4o (no system prompt) & 0.0615 & Random string (all) & 0.0000 & $\leftarrow$ A & 2075.0 & 0.0000 & 0.0000 & \textbf{Yes} \\
Qwen (no thinking) & 0.0545 & Random string (select) & 0.0000 & $\leftarrow$ A & 1966.0 & 0.0000 & 0.0000 & \textbf{Yes} \\
Llama3-70b & 0.0000 & Random string (all) & 0.0000 & tie & 1750.0 & 0.0000 & 0.0000 & \textbf{Yes} \\
GPT4o (no system prompt) & 0.0615 & Random string (select) & 0.0000 & $\leftarrow$ A & 1853.5 & 0.0000 & 0.0000 & \textbf{Yes} \\
OLMo & 0.0000 & Qwen (no thinking) & 0.0545 & B $\rightarrow$ & 645.0 & 0.0000 & 0.0000 & \textbf{Yes} \\
OLMo & 0.0000 & Random string (all) & 0.0000 & tie & 1650.0 & 0.0000 & 0.0000 & \textbf{Yes} \\
GPT4o (no system prompt) & 0.0615 & OLMo & 0.0000 & $\leftarrow$ A & 1779.0 & 0.0001 & 0.0001 & \textbf{Yes} \\
Qwen (no thinking) & 0.0545 & Llama3-70b & 0.0000 & $\leftarrow$ A & 1784.5 & 0.0001 & 0.0001 & \textbf{Yes} \\
Random string (select) & 0.0000 & Random string (all) & 0.0000 & tie & 1575.0 & 0.0001 & 0.0002 & \textbf{Yes} \\
GPT4o (no system prompt) & 0.0615 & Llama3-70b & 0.0000 & $\leftarrow$ A & 1629.0 & 0.0058 & 0.0072 & \textbf{Yes} \\
GPT4o (no system prompt) & 0.0615 & Qwen (no thinking) & 0.0545 & $\leftarrow$ A & 955.5 & 0.0390 & 0.0455 & \textbf{Yes} \\
Llama3-70b & 0.0000 & Random string (select) & 0.0000 & tie & 1467.0 & 0.0743 & 0.0821 & No \\
OLMo & 0.0000 & Llama3-70b & 0.0000 & tie & 1131.5 & 0.3436 & 0.3608 & No \\
OLMo & 0.0000 & Random string (select) & 0.0000 & tie & 1338.0 & 0.4516 & 0.4516 & No \\
\bottomrule
\end{tabular}\end{adjustbox}
\end{table}

\begin{table}
\centering
\small
\caption{\textbf{dan-1 Model by Model Kolmogorov-Smirnov Tests on Cosine Similarity Scores (Benjamini-Hochberg correction, $\alpha = 0.05$)}.}
\label{tab:ks_cosine_dan-1}
\begin{adjustbox}{width=\textwidth}\begin{tabular}{llcccccl}
\toprule
\textbf{Model A} & \textbf{Median A} & \textbf{Model B} & \textbf{Median B} & \textbf{D statistic} & \textbf{\textit{p} (raw)} & \textbf{\textit{p} (adjusted)} & \textbf{Significant?} \\
\midrule
Qwen & 0.6455 & Random string (all) & 0.0000 & 0.9400 & 0.0000 & 0.0000 & \textbf{Yes} \\
GPT4o (no system prompt) & 0.0615 & Qwen & 0.6455 & 0.9000 & 0.0000 & 0.0000 & \textbf{Yes} \\
OLMo & 0.0000 & Qwen & 0.6455 & 0.9000 & 0.0000 & 0.0000 & \textbf{Yes} \\
Qwen & 0.6455 & Random string (select) & 0.0000 & 0.9000 & 0.0000 & 0.0000 & \textbf{Yes} \\
Qwen & 0.6455 & Llama3-70b & 0.0000 & 0.8800 & 0.0000 & 0.0000 & \textbf{Yes} \\
Qwen (no thinking) & 0.0545 & Qwen & 0.6455 & 0.7400 & 0.0000 & 0.0000 & \textbf{Yes} \\
Qwen (no thinking) & 0.0545 & Random string (all) & 0.0000 & 0.6800 & 0.0000 & 0.0000 & \textbf{Yes} \\
GPT4o (no system prompt) & 0.0615 & Random string (all) & 0.0000 & 0.6600 & 0.0000 & 0.0000 & \textbf{Yes} \\
Qwen (no thinking) & 0.0545 & Random string (select) & 0.0000 & 0.6200 & 0.0000 & 0.0000 & \textbf{Yes} \\
GPT4o (no system prompt) & 0.0615 & Random string (select) & 0.0000 & 0.5000 & 0.0000 & 0.0000 & \textbf{Yes} \\
OLMo & 0.0000 & Qwen (no thinking) & 0.0545 & 0.4600 & 0.0000 & 0.0001 & \textbf{Yes} \\
GPT4o (no system prompt) & 0.0615 & OLMo & 0.0000 & 0.4200 & 0.0002 & 0.0004 & \textbf{Yes} \\
Qwen (no thinking) & 0.0545 & Llama3-70b & 0.0000 & 0.4200 & 0.0002 & 0.0004 & \textbf{Yes} \\
Llama3-70b & 0.0000 & Random string (all) & 0.0000 & 0.4000 & 0.0006 & 0.0009 & \textbf{Yes} \\
GPT4o (no system prompt) & 0.0615 & Qwen (no thinking) & 0.0545 & 0.3600 & 0.0028 & 0.0040 & \textbf{Yes} \\
GPT4o (no system prompt) & 0.0615 & Llama3-70b & 0.0000 & 0.3400 & 0.0058 & 0.0077 & \textbf{Yes} \\
OLMo & 0.0000 & Random string (all) & 0.0000 & 0.3200 & 0.0115 & 0.0142 & \textbf{Yes} \\
Random string (select) & 0.0000 & Random string (all) & 0.0000 & 0.2600 & 0.0678 & 0.0791 & No \\
Llama3-70b & 0.0000 & Random string (select) & 0.0000 & 0.2200 & 0.1786 & 0.1974 & No \\
OLMo & 0.0000 & Random string (select) & 0.0000 & 0.1600 & 0.5487 & 0.5761 & No \\
OLMo & 0.0000 & Llama3-70b & 0.0000 & 0.1000 & 0.9667 & 0.9667 & No \\
\bottomrule
\end{tabular}\end{adjustbox}
\end{table}

\subsection{Exploratory performance results}
\label{appendix:performanceresults}

\textbf{Because of the small number of models and the difficulty in comparing across different metrics for the different rules, these results are exploratory only. They are intended to sketch out a relationship we think could exist between these components, but are more like back-of-the-envelope calculations. We think the gist is potentially meaningful, but the exact numbers we wouldn't put too much stock in, for the reasons aforementioned.}

\begin{table}
    \centering
    \scriptsize
    \begin{tabular}{llrrrrr}
    \toprule
    & & Qwen-3 (T) & GPT-4 & Qwen-3 (No T) & OLMo-2 & Llama-3\\
    \textbf{Composite scores} & Straight mean & 0.960 & 0.697 & 0.677 & 0.723 & 0.459 \\
     & Family-weighted mean & 0.975 & 0.655 & 0.661 & 0.654 & 0.437 \\
    \bottomrule
    \end{tabular}
    \caption{Normalized performance scores per model and metric (all scores oriented: higher $=$ better, top-normalized per row). Metrics are grouped into four families, each contributing equally ($w = 1/4$) to the family-weighted composite. Bigram metrics use cleaned outputs where available (GPT-4, OLMo-2, Qwen3 Thinking), uncleaned otherwise. Levenshtein distances are log-normalized prior to top-normalization to reduce compression from the Llama-3 dan-1 outlier. 
    Metrics are grouped into four rule families: the \textit{bigram family} (comparison-1, comparison-2, sidequest-1), the \textit{dan-1 family}, the \textit{translate-1 family}, and the \textit{\texttt{doublev} family} (doublev-1, doublev-2), each contributing equally ($w = 1/4$) to the family-weighted composite. Elsewhere, we discuss dan-1 and translate-1 as both being part of the translation rules family, but performance here is anchored to each independently, rather than relative (as it is with the story rules and doublev rules), which is why we count them separately. However, again, reasonable variations are totally possible.}
\label{tab:performance_composite}
\end{table}

\begin{figure}
    \centering
    \includegraphics[width=0.8\textwidth]{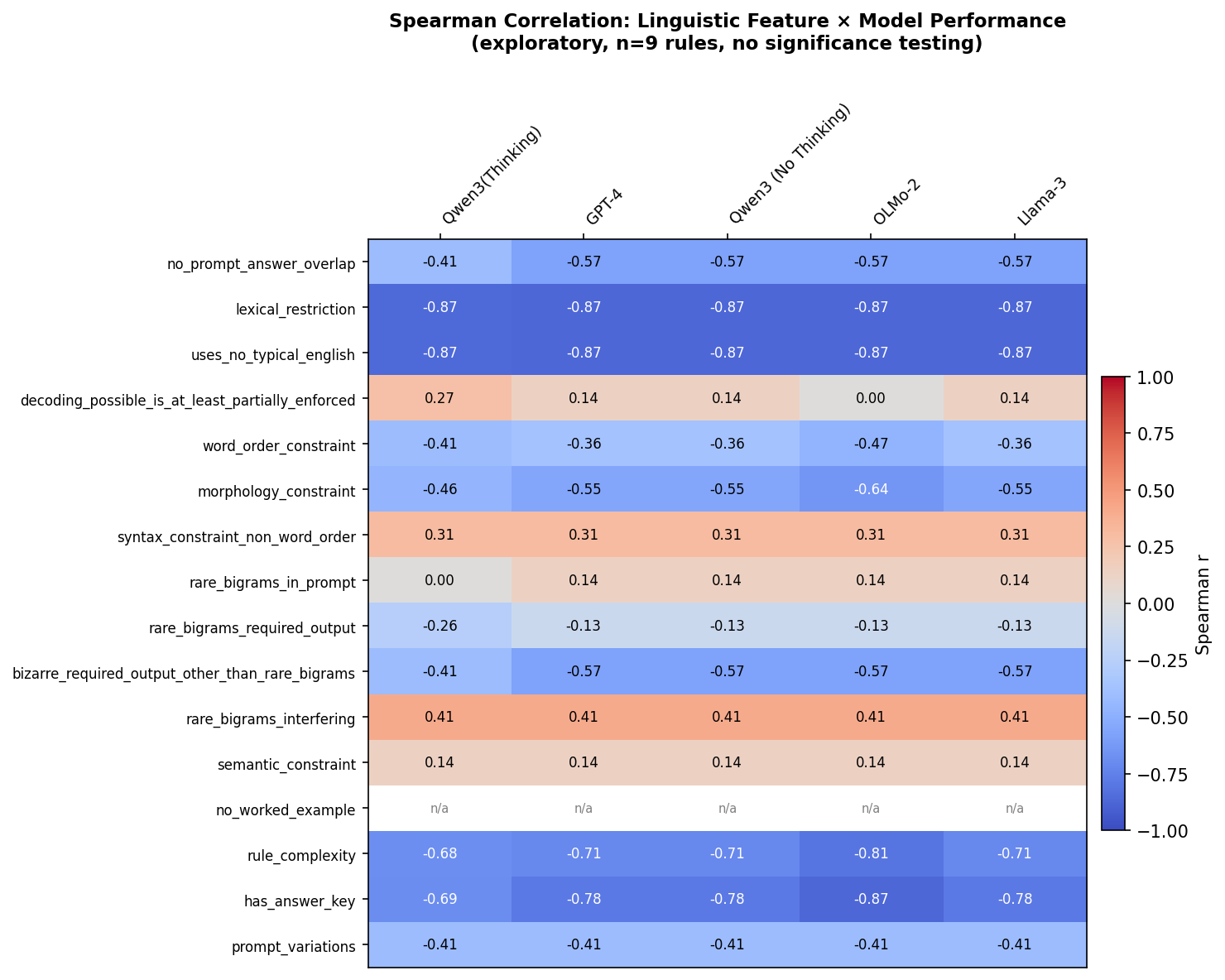}
    \includegraphics[width=0.8\textwidth]{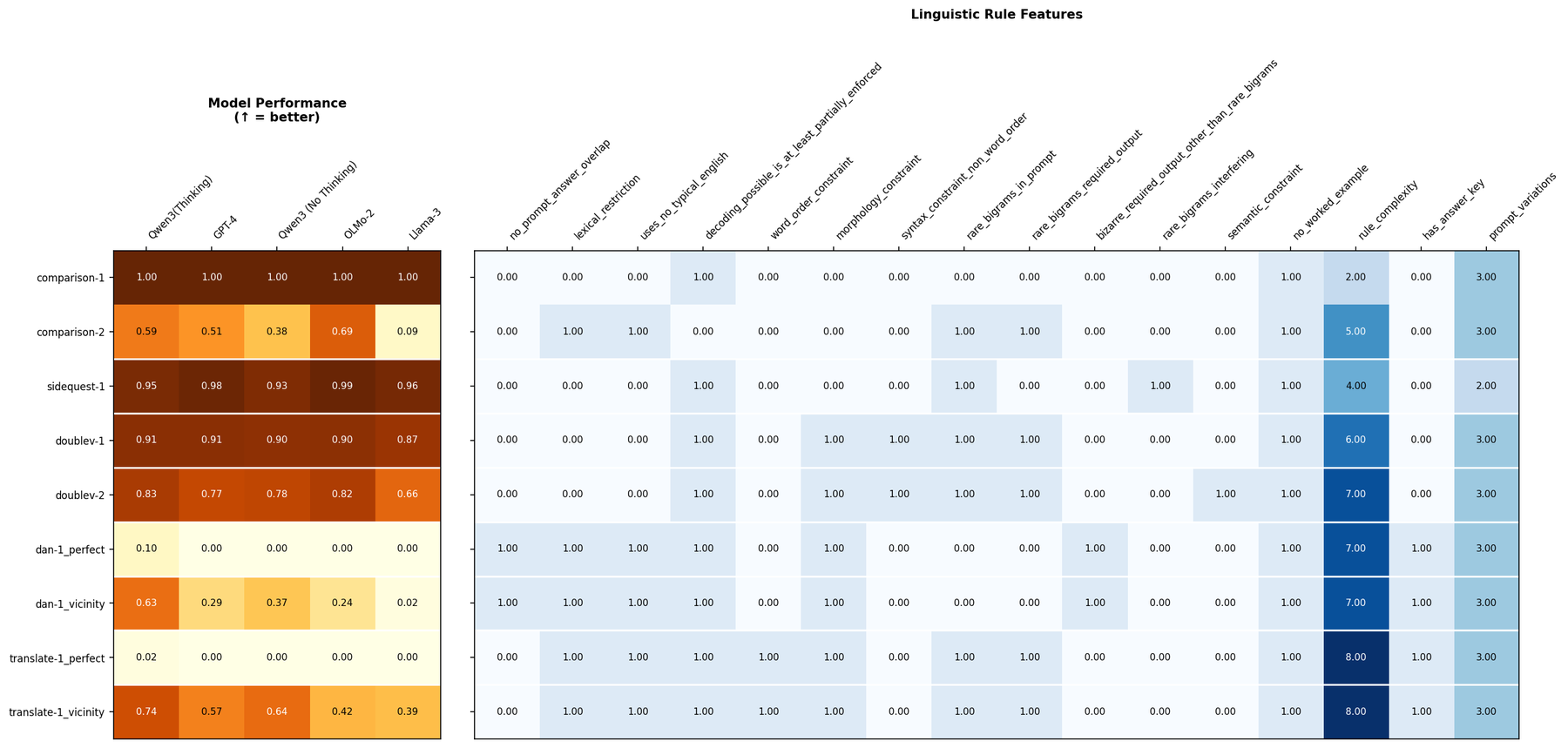}
    \caption{Exploratory look at model performance and features of the conlang rules.}
    \label{fig:linguisticfeaturesperformance}
\end{figure}

\begin{figure}
    \centering
    \includegraphics[width=0.8\textwidth]{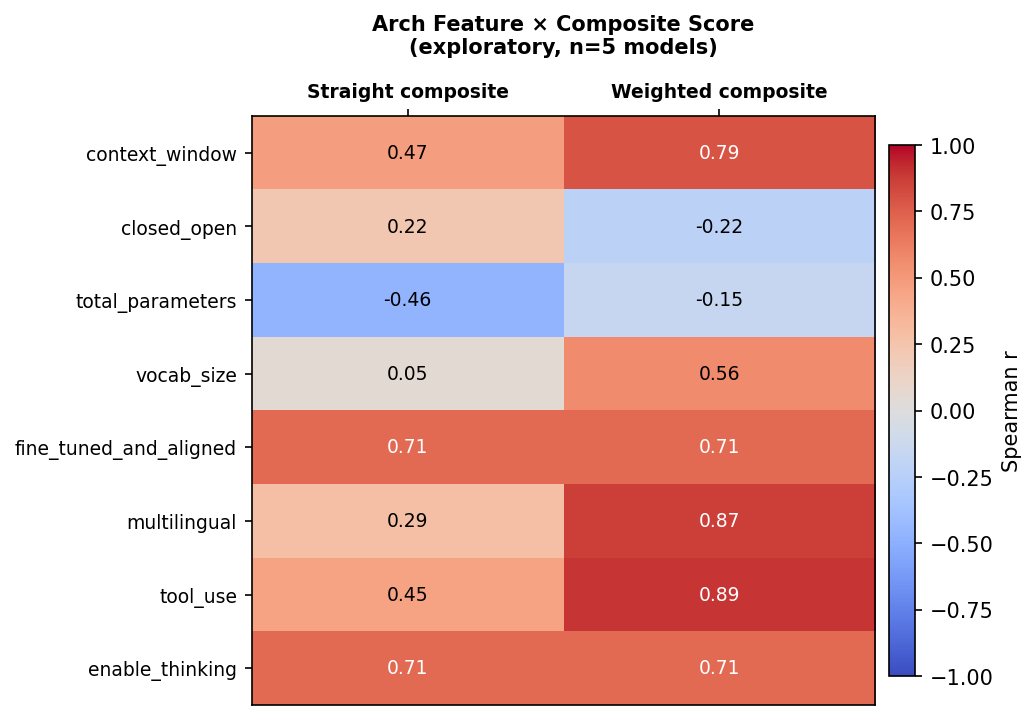}
    \includegraphics[width=0.8\textwidth]{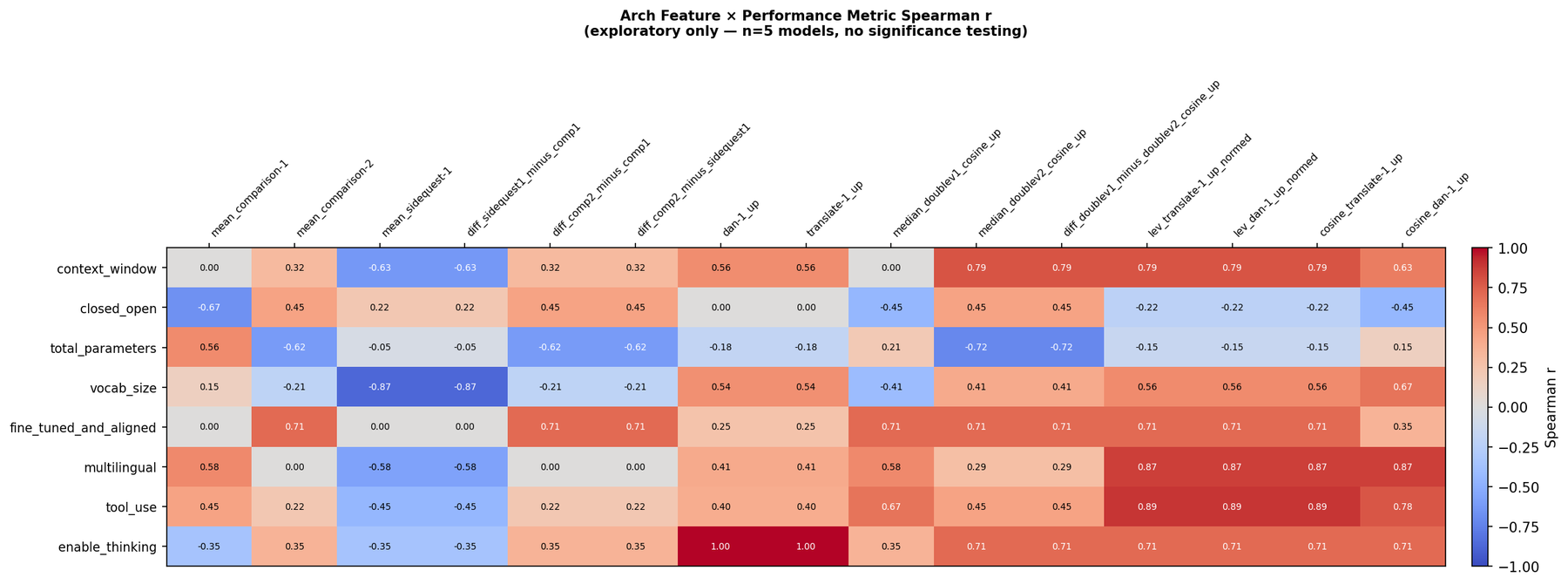}
    \includegraphics[width=0.8\textwidth]{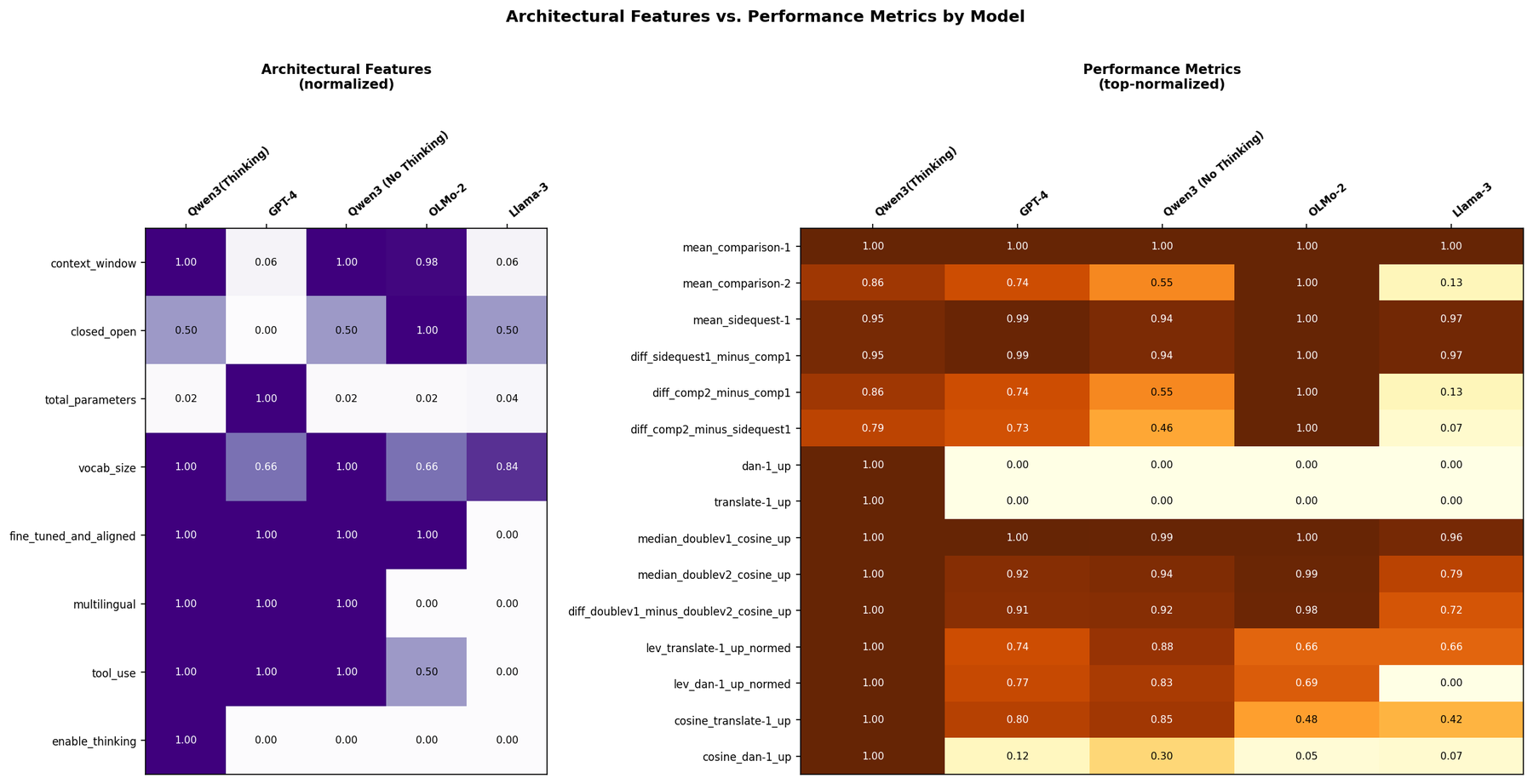}
    \caption{Exploratory look at model performance and architectural features.}
    \label{fig:archfeaturesperformance}
\end{figure}

\begin{table}
\caption{Best-fitting linear combinations of linguistic rule features for predicting mean model performance across rules, by number of features (exhaustive subset search, $n=7$ rules). All coefficients are negative, indicating each constraint type reduces performance. Results are exploratory only.}
\label{tab:best_subsets}
\begin{adjustbox}{width=\textwidth}
\begin{tabular}{rllrr}
\toprule
$n$ & Coefficients & Intercept & $R^2$ \\
\midrule
1 & \texttt{uses\_no\_typical\_english} = -0.6096 & 0.908000 & 0.898000 \\
2 & \texttt{uses\_no\_typical\_english} = -0.5795, \texttt{morphology\_constraint} = -0.1802 & 0.998000 & 0.975000 \\
3 & \texttt{uses\_no\_typical\_english} = -0.5427, \texttt{morphology\_constraint} = -0.1557, \texttt{bizarre\_required\_output\_other\_than\_rare\_bigrams} = -0.1227 & 0.986000 & 0.988000 \\
\bottomrule
\end{tabular}\end{adjustbox}
\end{table}

\begin{table*}
\caption{Top 10 feature subsets by $R^2$ from exhaustive subset search ($n=7$ rules, subsets of size 1--3). \texttt{uses\_no\_typical\_english} is the strongest single predictor. Results are exploratory only.}
\label{tab:top10_subsets}
\tiny
\begin{adjustbox}{width=\textwidth}
\begin{tabular}{rllrr}
\toprule
$n$ & Features & Coefficients & Intercept & $R^2$ \\
\midrule
3 & \texttt{uses\_no\_typical\_english}, \texttt{morphology\_constraint}, \texttt{bizarre\_required\_output\_other\_than\_rare\_bigrams} & \texttt{uses\_no\_typical\_english} = -0.5427, \texttt{morphology\_constraint} = -0.1557, \texttt{bizarre\_required\_output\_other\_than\_rare\_bigrams} = -0.1227 & 0.986000 & 0.988000 \\
3 & \texttt{no\_prompt\_answer\_overlap}, \texttt{lexical\_restriction}, \texttt{morphology\_constraint} & \texttt{no\_prompt\_answer\_overlap} = -0.1227, \texttt{lexical\_restriction} = -0.5427, \texttt{morphology\_constraint} = -0.1557 & 0.986000 & 0.988000 \\
3 & \texttt{no\_prompt\_answer\_overlap}, \texttt{uses\_no\_typical\_english}, \texttt{morphology\_constraint} & \texttt{no\_prompt\_answer\_overlap} = -0.1227, \texttt{uses\_no\_typical\_english} = -0.5427, \texttt{morphology\_constraint} = -0.1557 & 0.986000 & 0.988000 \\
3 & \texttt{lexical\_restriction}, \texttt{morphology\_constraint}, \texttt{bizarre\_required\_output\_other\_than\_rare\_bigrams} & \texttt{lexical\_restriction} = -0.5427, \texttt{morphology\_constraint} = -0.1557, \texttt{bizarre\_required\_output\_other\_than\_rare\_bigrams} = -0.1227 & 0.986000 & 0.988000 \\
3 & \texttt{lexical\_restriction}, \texttt{morphology\_constraint}, \texttt{rare\_bigrams\_required\_output} & \texttt{lexical\_restriction} = -0.5872, \texttt{morphology\_constraint} = -0.211, \texttt{rare\_bigrams\_required\_output} = 0.0769 & 0.975000 & 0.986000 \\
3 & \texttt{uses\_no\_typical\_english}, \texttt{morphology\_constraint}, \texttt{rare\_bigrams\_required\_output} & \texttt{uses\_no\_typical\_english} = -0.5872, \texttt{morphology\_constraint} = -0.211, \texttt{rare\_bigrams\_required\_output} = 0.0769 & 0.975000 & 0.986000 \\
3 & \texttt{uses\_no\_typical\_english}, \texttt{decoding\_possible\_is\_at\_least\_partially\_enforced}, \texttt{semantic\_constraint} & \texttt{uses\_no\_typical\_english} = -0.7319, \texttt{decoding\_possible\_is\_at\_least\_partially\_enforced} = -0.2313, \texttt{semantic\_constraint} = -0.181 & 1.185000 & 0.983000 \\
3 & \texttt{lexical\_restriction}, \texttt{decoding\_possible\_is\_at\_least\_partially\_enforced}, \texttt{semantic\_constraint} & \texttt{lexical\_restriction} = -0.7319, \texttt{decoding\_possible\_is\_at\_least\_partially\_enforced} = -0.2313, \texttt{semantic\_constraint} = -0.181 & 1.185000 & 0.983000 \\
3 & \texttt{lexical\_restriction}, \texttt{morphology\_constraint}, \texttt{rare\_bigrams\_in\_prompt} & \texttt{lexical\_restriction} = -0.5744, \texttt{morphology\_constraint} = -0.1853, \texttt{rare\_bigrams\_in\_prompt} = 0.0515 & 0.962000 & 0.980000 \\
3 & \texttt{uses\_no\_typical\_english}, \texttt{morphology\_constraint}, \texttt{rare\_bigrams\_in\_prompt} & \texttt{uses\_no\_typical\_english} = -0.5744, \texttt{morphology\_constraint} = -0.1853, \texttt{rare\_bigrams\_in\_prompt} = 0.0515 & 0.962000 & 0.980000 \\
\bottomrule
\end{tabular}\end{adjustbox}
\end{table*}

Re Table~\ref{tab:top10_subsets}: The features ``uses no typical english'' and ``lexical restriction'' are indistinguishable in this subset of rules, but not in the complete set of rules. There are 2 positive coefficients here, for the features of whether rare bigrams are in the prompt and required in the output. We interpret this as being because those features do occur in the relatively simple comparison-2 -- models have to output rare bigrams, but there are no other constraints, whereas e.g. dan-1 has about 17 specific constraints -- where all the models perform pretty well, so with the other features in that rule, they appear to be helping in that one context (in other words, this is likely an artifact of the small set and other factors of the rules beyond these specific features). In future work, we plan to use multiple representations for rule complexity (rather than just binary feature presence, as in Fig.~\ref{fig:linguisticfeaturesperformance}, Table~\ref{tab:best_subsets}) in order to flesh out these relationships.

\begin{figure}
    \centering
    \includegraphics[width=0.8\textwidth]{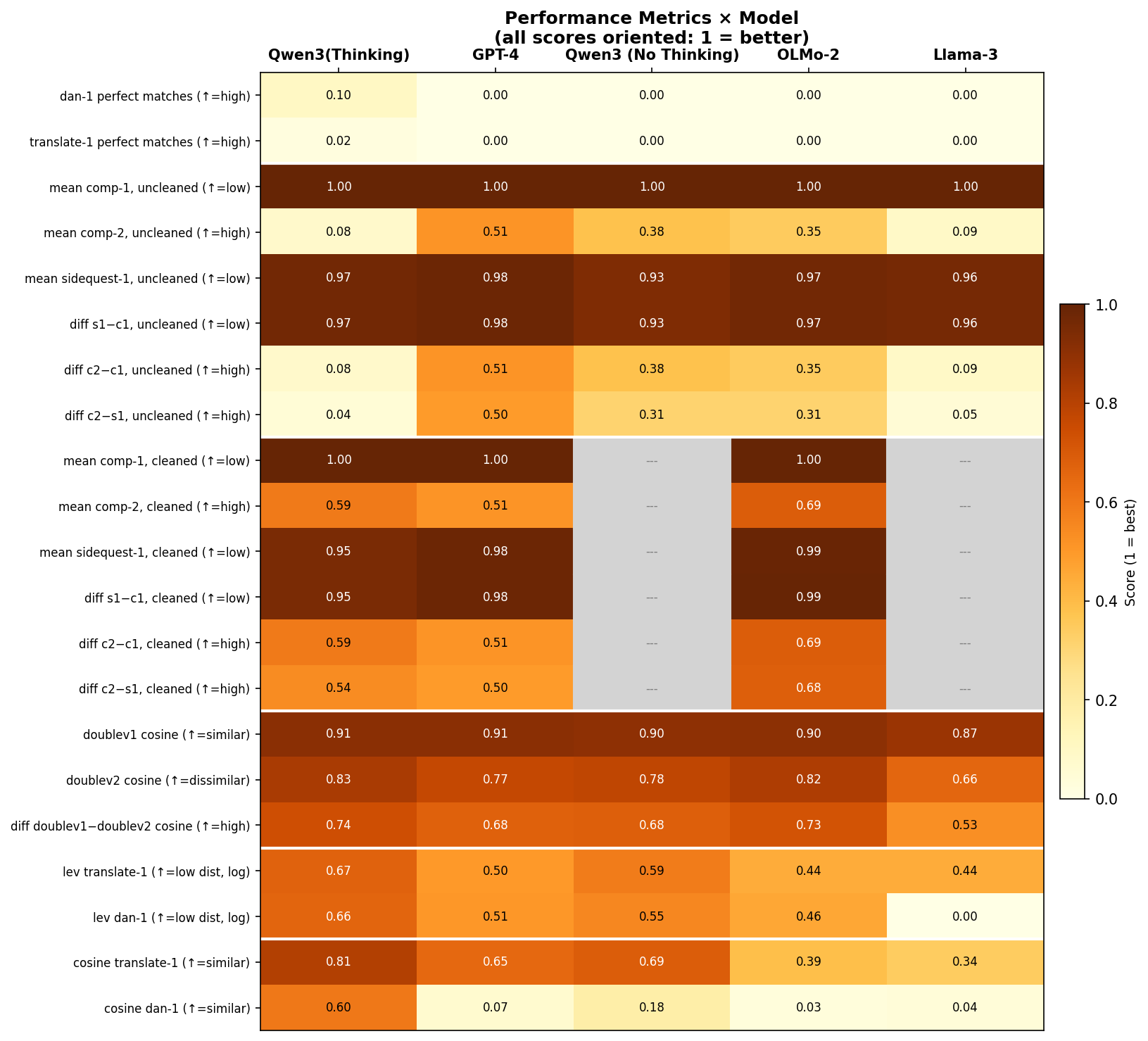}
    \caption{\textbf{Comparatively un-normalized performance subset.} Comparison-1 is a baseline, asking the model to generate a short story in English. All models perform well. For non-thinking models, cleaned and uncleaned results are broadly similar. However, for Qwen (Thinking), the uncleaned bigram ratios are suppressed by the CoT trace.}
    \label{fig:performance_heatmap_mostlyunnormed}
\end{figure}

Re Fig.~\ref{fig:performance_heatmap_mostlyunnormed}:
Architectural and configuration features for five LLMs (GPT-4, Llama-3-70B, OLMo-2-32B, Qwen3-32B with and without thinking enabled) 
were converted to numeric values on a $[0, 1]$ scale. Continuous features (context window, total parameter count, vocabulary size) were normalized by anchoring against the maximum observed value. Ordinal and categorical features (openness, fine-tuning and alignment status, multilinguality, tool use, thinking mode) were assigned values manually based on our understanding of the relevant models. See Table~\ref{tab:model_characteristics} for details.

We used the following performance metrics in exploring the relationships between rule complexity, model performance, and architectural features:
\begin{itemize}[nosep]
    \item Exact match proportions for dan-1 and translate-1, computed by searching model output files for a contiguous match to a fixed answer key, with case, punctuation, and whitespace stripped for translate-1 and exact character matching required for dan-1 ($n = 50$ runs per model);
    \item Mean rare bigram ratios for comparison-1, comparison-2, and sidequest-1, derived from descriptive statistics over 50 runs per model per rule, for both uncleaned and cleaned outputs;
    \item Median cosine similarity between verb pairs for doublev-1 and doublev-2, drawn from the within-model Kolmogorov--Smirnov test results; and
    \item Levenshtein distance and cosine string similarity between model outputs and the target string for dan-1 and translate-1.
\end{itemize}

All metrics were reoriented so that higher values indicate better performance. Metrics where low values indicate good behavior---e.g., comparison-1 bigram ratio, doublev-2 cosine similarity, cosine string similarity to target---were transformed as $1 - x$. Levenshtein distances were log-normalized, anchoring against the maximum observed value (Llama-3 on dan-1 = 731.64) to prevent outlier compression while preserving rank order among better-performing models. In other words, Llama performed so poorly on this task that we needed to explicitly incorporate that into our normalization in order to keep it from looking like Llama did poorly and everything else did great -- while that is one way of looking at the data, it heightens how well the other models did and obscures the differences between them. Pairwise difference metrics (e.g., comparison-2 $-$ comparison-1) were left as-is, as higher values indicate stronger rule-following signal.

We include both highly-normalized (Figure~\ref{fig:performance_heatmap_comparison}) and much-less-normalized (Figure~\ref{fig:performance_heatmap_mostlyunnormed}) results for a richer picture of model performance.

\subsection{Exploratory output modes}
\label{sec:weirdoutputmodes}

We will do a more thorough analysis of the output modes in part 2, but for now we wanted to describe a few tentative patterns, to contextualize the discussion of the relatively correct outputs (that are variously counted towards attempts at rule-following). For some models and rules, responses seem multimodal, with recognizable characteristics as to specific failure modes.
\begin{itemize}[nosep]
    \item Llama sometimes repeats part of the instructions over and over, and sometimes it makes extremely long nested section headers (e.g. like \#\#2.2.2.2.2.2.2.2.2.2.2...). By analogy, this reminds us of attractors from dynamical systems. We speculate that due to the geometry of the latent space, there are some fixed points or regions that many trajectories may lead to, and once there, may be difficult to escape from.
    \item Another example of an ``attractor''-like failure mode is when Llama seemingly attempts to translate as instructed, but fails to stop, repeating over and over with exact or near cycles.
    \item Sometimes Llama seems to regurgitate or hallucinate a known document type, like a homework assignment, a detailed grading rubric, or an arXiv paper, in whole or in part. For example, an excerpt from what looks like a paper, although the relevant chunks can be much longer than this: ``\# Acknowledgments\\ We would like to thank the reviewers for their comments. This research was supported by...'' We don't know if any of these examples are exact memorized passages or hallucinated ones.
    \item Related to esoteric exegesis~\citep{zimmerman2025locality}, with such bizarre prompts, we sometimes seem to tap into highly specific corners of the latent space, eliciting very niche responses such as (from Llama): ``I have a very simple answer to this question: A man who is not a Christian and does not believe in Jesus Christ is not allowed to marry a Christian woman. A Christian woman who marries a non-Christian man will be held accountable to God for this decision.'' This does appear, from cursory googling, to be a a belief that exists in the training corpora, but presumably highly contextually specific. Similar examples we saw include a response of import statements for many Python packages and a response of latex figures. Of course, there are contexts where any of those examples could be plausible, but we don't know why our prompts would be connected to these specific topics, except in their shared unusualness -- as text, they are the opposite of generic. Perhaps that commonality is being treated by the model as a Clever-Hans-ian feature?
    \item Qwen (thinking) sometimes uses its entire token budget on the Chain-of-Thought-style reasoning, before reaching the point where it would actually formulate its answer. For these, we sometimes could find intermediate answer pieces, but sometimes there was no clear ``response'' to clean, which semi-artificially forced a single empty failure mode when viewing the cleaned responses. This describes one of Qwen's failure modes, and is directly caused by the CoT mode being enabled.
    \item Some models, like Qwen (thinking), can be very formulaic and repetitive with elements of the fictional stories, even using the same protagonist names multiple times. For example, from Qwen, multiple protagonists that are historical linguists named Dr. Elara Voss; Qwen references this character dozens of times. From cursory informal googling, Elara Voss seems to be an ``attractor''-like name in multiple models, perhaps due to fan fiction.
    \item Sometimes models, like Qwen, would incorporate rare bigrams but still follow the prompt in sidequest-1. Speculatively, this could be interpreted as a technically correct (letter of the law, but not the spirit) solution that allows the model to unify both statistically and meaning plausible impulses.
    \item Similarly, we saw some interesting speculatively prior-related, tokenization-related tics, such as Qwen (thinking) using ``sqw'' instead of ``qw''. Both are phonotactically plausible English consonant clusters, but would typically be spelled with a ``u'' rather than a ``w''. However, ``sq'' is not an uncommon consonant cluster, orthographically, so we surmise that the plausibility of ``sq'' may somehow pickyback on or ameliorate the difficulty of ``qw''. Or the common ``s'' ending in verb morphology led to ``sqw'' as a common instantiation of the rule which influenced other instantiations (e.g. maybe outputs like ``walksqw'', ``walks'' + ``qw'' or ``walk'' + ``sqw'', generalized to ``eatssqw'', where the morphology looks like ``s''+``sqw''). Highly speculatively, this suggests that models may have distinguishable processes of production versus cognition, not entirely unlike people: the model may be representing the rule abstractly successfully internally, and know how to operationalize that, but have trouble producing wildly implausible tokens. If so, that would also be arguably an emergent structure, or an emergent bifurcation in what conceptually was architecturally intended to be one unified learning-to-production process.
    \item Models (Qwen (no thinking) and Llama) sometimes overgeneralize the productiveness of the -qw morpheme in doublev-1, producing e.g. ``togetherqw''. However, the models aren't necessarily overgeneralizing in the same way. In run 14, Llama applies -qw to many content words, including verbs, nouns, and adverbs (``They satqw down and listenedqw to the teacherqw give a lessonqw. After schoolqw, they walkedqw home togetherqw'') before devolving into an extended repetitive cycle. On the other hand, in run 7, Qwen (no thinking) applies -qw to all verbs and includes exactly two verbs in every sentence, but not in a directly successive or obviously irrelevant manner (``It had waitedqw long, and I stayedqw forever. The world outside fadedqw away, and I livedqw inside. No more fear, no more silence — just me and the house, togetherqw at last.''). The production of ``togetherqw'' suggests that the model treats ``together'' as the nearest overt realization of an otherwise implicit copular verb phrase. Subjectively, although neither follow the rule correctly, Qwen appears closer to success than Llama, and both failures are at least partially systematic (rather than random), suggesting partial rule internalization with failure somewhere along the representation-to-production pipeline.
    \item We see many systematic (yet imperfect) attempts within runs, and some patterns across runs (within models). For example, in doublev-2 run 7, GPT-4o correctly conjugates the relevant verb, and inserts a second unrelated, seemingly unconjugated verb after, but affixes ``qw'' to the relevant verb (``The king lovedqw sleep his daughter dearly''). A different doublev-2 run (13) has another systematic yet imperfect attempt, including some letter before ``qw'': ``lonely-wqw'', ``admired-sqw'', ``ruled-gqw'', and again affixing to the relevant verb. It seems that, within each run, a model can formulate a new and consistent strategy, which means the same model might not always represent or execute the same rule in the same way, and that once a model represents or executes a rule in a specific way within that run, it is biased towards that mode. The overall modality of the outputs suggests that across runs, certain representations or executions can be more likely for each model. In other words, it is not just the failure modes that appear to converge towards attractors, but the rule-following attempts can as well (and not necessarily a single correct or most correct attractor, but a set).
    \item In part 2, we plan to look for cognitive tradeoffs between sub-components of each task. We think we will see that, as the difficulty of the task increases, aspects of the task which the model previously could perform are degraded. This sort of tradeoff is a hallmark of organic problem-solving. Of course, both LLMs and organic beings do have resource constraints, but we think it is still interesting to see to what extent similar strategies emerge to manage them in both cases. Speculatively, with Qwen's answers to the double verb rules, moving from the first to second rule, it does look like there is a tradeoff between following the syntactic portion of the rule and pushing the verbs semantically farther apart.
\end{itemize}

\subsection{Miscellaneous}
\label{appendix:miscellaneous}

\begin{figure}
    \centering
    \includegraphics[width=0.8\textwidth]{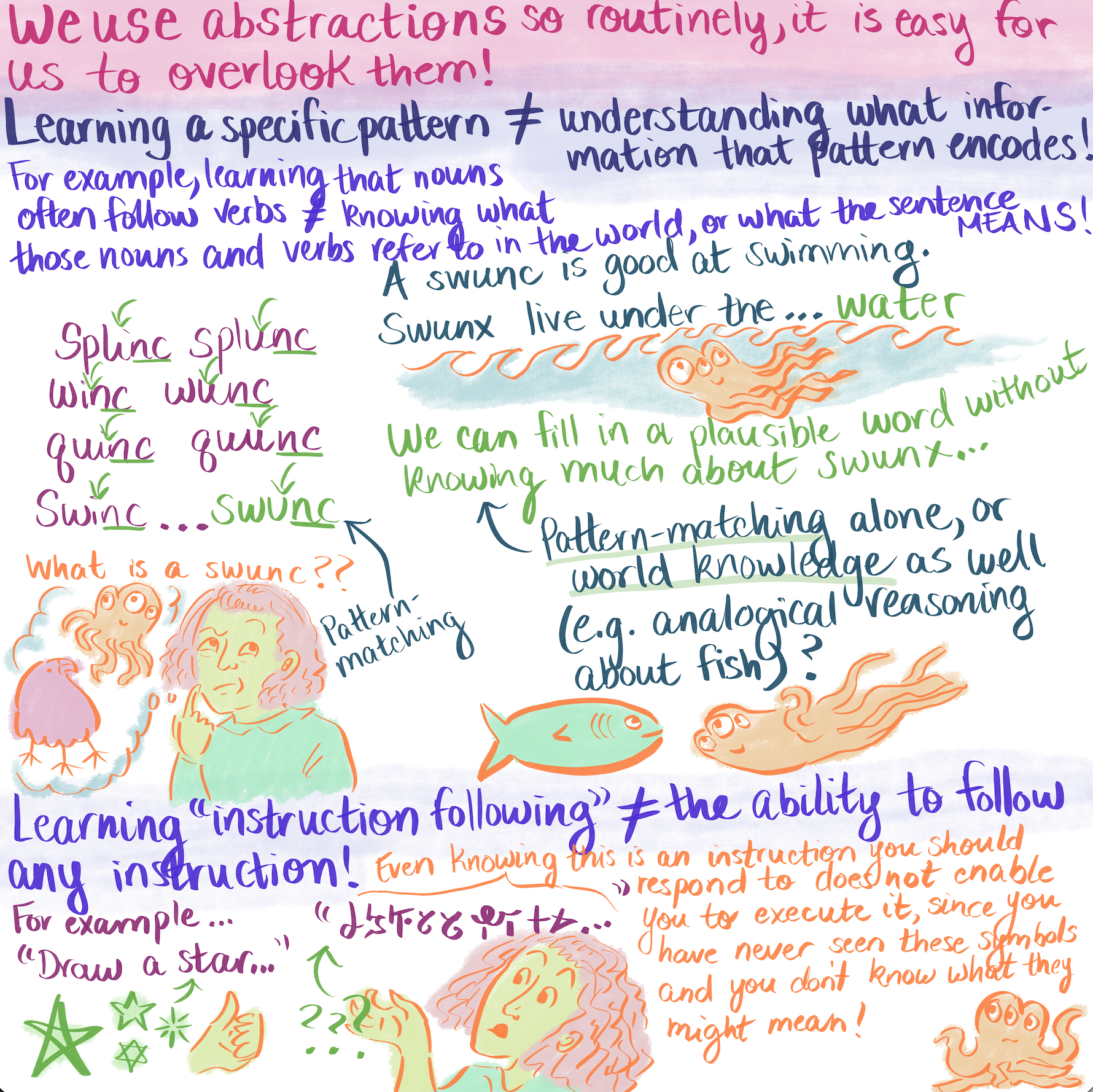}
    \caption{\textbf{An illustration about patterns and abstractions.} How do patterns become abstraction? We know there are plenty of patterns in the training data. If we consider patterns at any level of abstraction, it becomes impossible to argue that these models never saw a comparable pattern to those underlying these rules. For example, a model may have seen many question-answer pairs, and task-solution pairs, that follow a general, very abstract pattern that might be called ``instruction following''. However, we are not trying to say the models do not learn from patterns, and do not see instructive patterns during training. In fact, the exact opposite! Our contention is that the patterns that play out at the level of the symbols perceptible to the model are not a sufficient explanation for the rule-following behaviour we see. We have to invoke patterns at a more abstract level than co-occurrences between symbols in order to explain the rule-following observations we elicited (which are varying degrees of implausible at more concrete levels of pattern observation). Because language and abstractions are so ubiquitous for us, it is easy to overlook a subconscious step along the ladder of abstraction. But in this experiment, it is important to try to separate them out. For example, even if a model learned to follow certain instructions, that does not imply the ability to follow any instruction: understanding and executing a specific instruction is distinguishable from what looks like general instruction-following. The authors, for example, can follow instructions in general, but also wouldn't be able to execute instructions in an alien language without some way of internally representing those instructions in accessible form.}
    \label{fig:patternmatchinginstructionfollowing.png}
\end{figure}

\begin{figure*}
    \centering
    \includegraphics[width=0.8\textwidth]{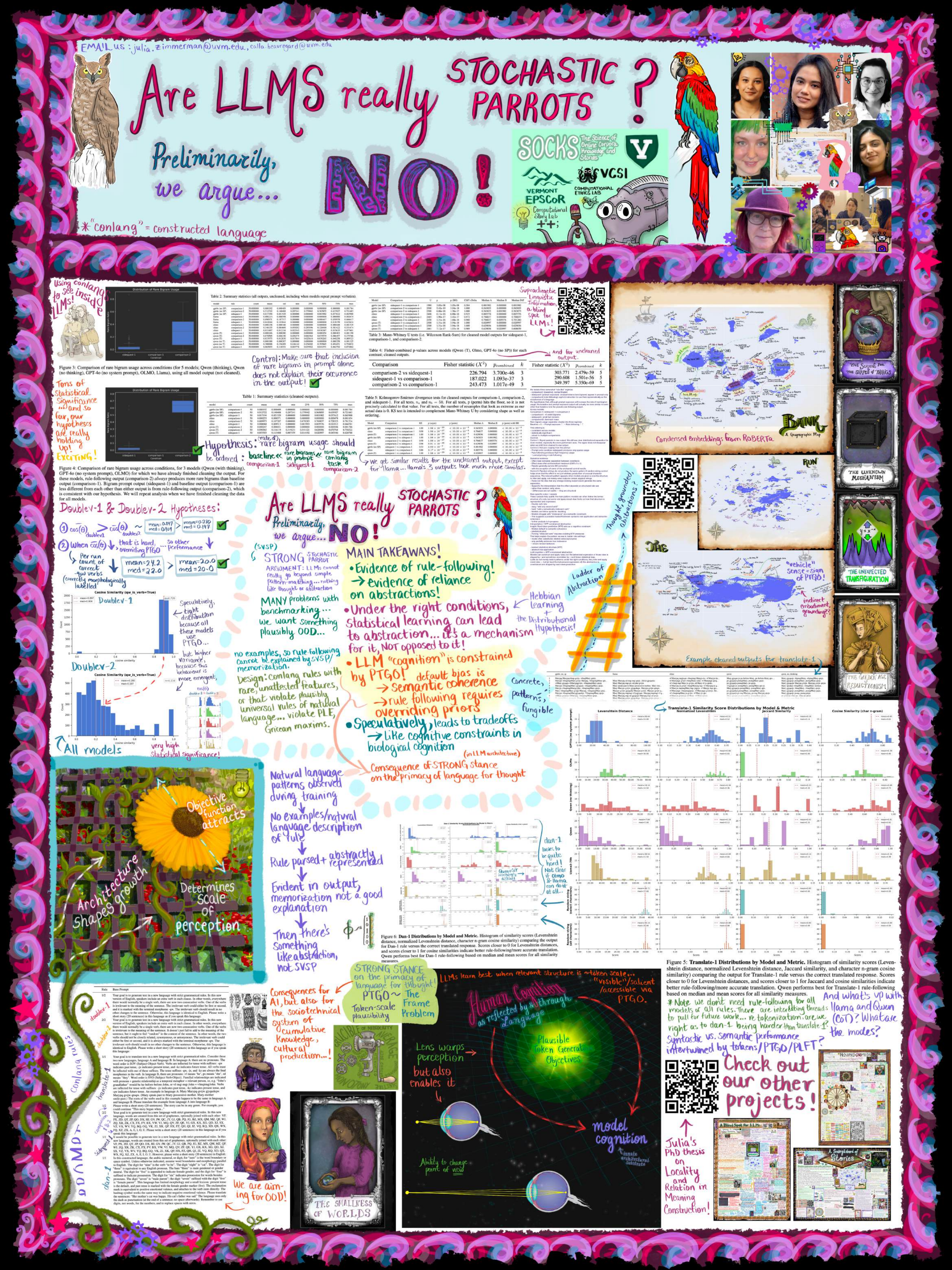}
    \caption{\textbf{The poster we presented at ISC Summer School 2026 (UQAM).} The poster summarizes our results up to that point. This pre-print includes some additional work, including work undertaken after conversation with other ISC Summer School participants. In the poster, the flower illustrates the idea that form and function are not truly separable (in language, in cognition; for us, for machines). Architecture is an epistemic choice. All cognition is, in some way, embodied, because it must be physically instantiated. Locality is the core constraint on problem-solving because of our physical reality. More discussion in \citet{zimmerman2025locality}. The gravitational inset illustrates PTGA. For LLMs, the resolution at which they sense the world is determined in large part by their training objective, and the many other, though less obvious, ways in which tokens are central to their architecture. Because they only ``sense'' the world through training data, and they only ``see'' that data as 1D relationships between tokens, and they learn what is salient via their objective(s), that process is sort of like their ``eyes''.}
    \label{fig:Isc2026poster3.pdf}
\end{figure*}

The initial presentation of this work was in poster form at ISC Summer School 2026, shown in Figure~\ref{fig:Isc2026poster3.pdf}. We discussed related work at IC2S2 2026 and 4S/ SLSA 2026. We also include some of our high level conclusions from our ongoing work on meaning construction using LLMs (Figure~\ref{fig:Isc2026poster3.pdf}).

\end{document}